\documentclass[10pt,twocolumn,letterpaper]{article}

\usepackage[pagenumbers]{cvpr} 

\usepackage[linesnumbered,ruled,vlined]{algorithm2e}

\usepackage{colortbl} 
\usepackage{graphicx}
\usepackage{booktabs}
\usepackage{multirow}
\usepackage{makecell}
\usepackage{arydshln}
\usepackage{rotating}
\usepackage{algpseudocode}
\usepackage{setspace}
\usepackage{amsmath}
\usepackage{bbm}
\usepackage{subcaption}
\usepackage[usestackEOL]{stackengine}

\definecolor{LightGray}{gray}{0.85}
\definecolor{darkerGreen}{rgb}{0,0.5,0}
\definecolor{LightBlue}{RGB}{173, 216, 230}

\usepackage{pifont}
\newcommand{\argmax}{\mathop{\mathrm{argmax}}\limits} 

\usepackage{mathtools}
\DeclarePairedDelimiter\floor{\lfloor}{\rfloor}

\DeclareMathOperator{\Tr}{Tr}

\definecolor{cvprblue}{rgb}{0.21,0.49,0.74}
\usepackage[pagebackref,breaklinks,colorlinks,allcolors=cvprblue]{hyperref}

\def\paperID{12933} 
\def\confName{CVPR}
\def\confYear{2025}

\title{Realistic Test-Time Adaptation of Vision-Language Models

}

\author{Maxime Zanella\thanks{\hspace{0.1cm} Equal contributions and corresponding authors. \texttt{\{maxime.zanella,clement.fuchs\}@uclouvain.be}} $^{\hspace{0.5mm} 1,2}$ \hspace{6mm} Clément Fuchs$^{*}$$^{1}$ \hspace{6mm} Christophe De Vleeschouwer$^{1}$ \hspace{6mm} Ismail Ben Ayed$^{3}$
\\
$^{1}$UCLouvain, Belgium \hspace{6mm} $^{2}$UMons, Belgium \hspace{6mm} $^{3}$\'{E}TS Montreal, Canada
}

\begin{document}
\maketitle
\begin{abstract}
The zero-shot capabilities of Vision-Language Models (VLMs) have been widely leveraged to improve predictive performance. However, previous works on transductive or test-time adaptation (TTA) often make strong assumptions about the data distribution, such as the presence of all classes. Our work challenges these favorable deployment scenarios, and introduces a more realistic evaluation framework, including: (i) a variable number of effective classes for adaptation within a single batch, and (ii) non-i.i.d. batches of test samples in online adaptation settings. We provide comprehensive evaluations, comparisons, and ablation studies that demonstrate how current transductive or TTA methods for VLMs systematically compromise the models’ initial zero-shot robustness across various realistic scenarios, favoring performance gains under advantageous assumptions about the test samples' distributions. Furthermore, we introduce Stat${\cal A}$, a versatile method that could handle a wide range of deployment scenarios, including those with a variable number of effective classes at test time. Our approach incorporates a novel regularization term designed specifically for VLMs, which acts as a statistical anchor preserving the initial text-encoder knowledge, particularly in low-data regimes. Code available at \url{https://github.com/MaxZanella/StatA}.

\end{abstract}

\section{Introduction}
\begin{figure}
    \centering
    \begin{subfigure}[b]{\linewidth}
        \centering
        \includegraphics[width=\linewidth]{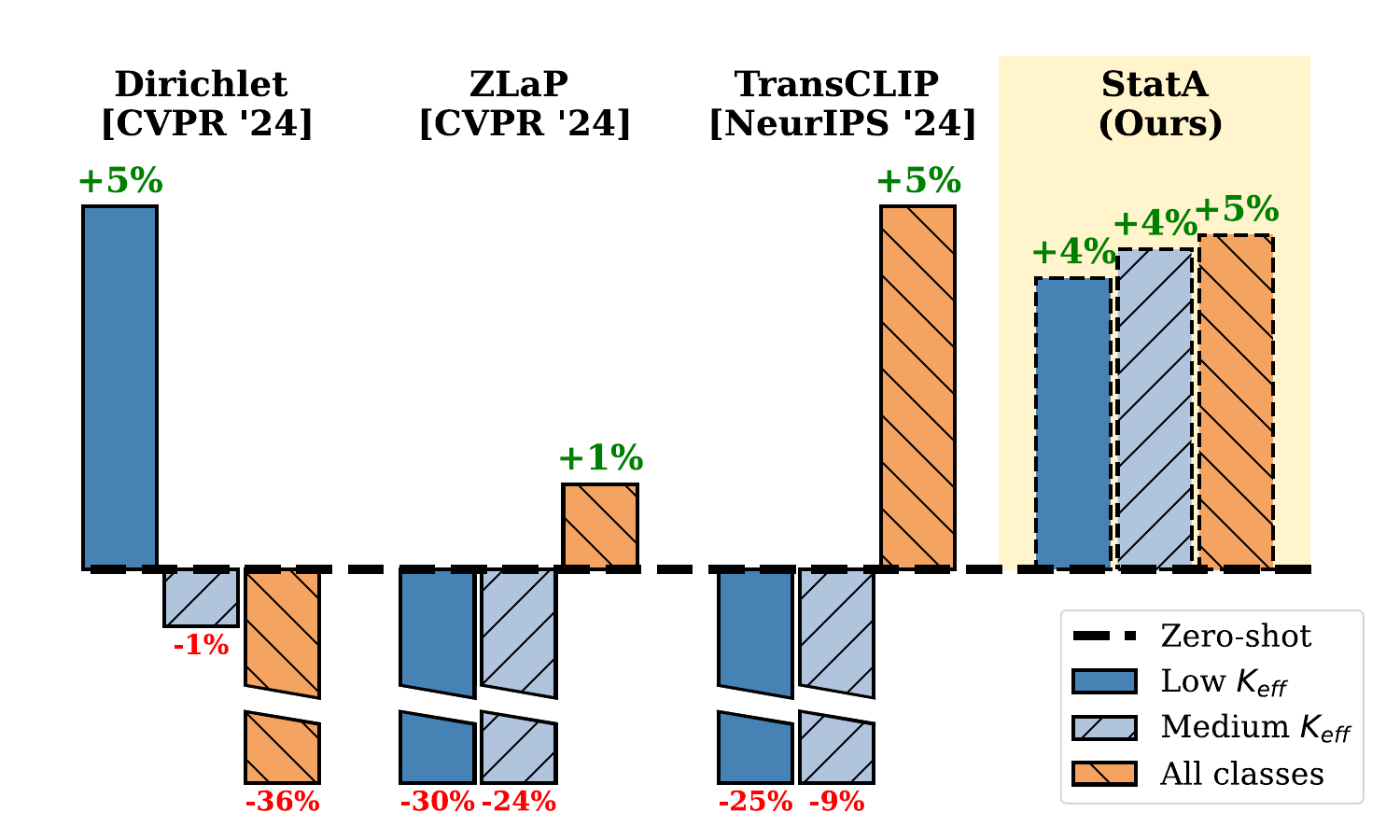}
        \caption{Stat${\cal A}$ brings consistent improvement when facing \texttt{Low} (between 2 and 10), \texttt{Medium} (between 5 and 25) number of effective classes ($K_{\text{eff}}$) in each batch, or \texttt{All} classes. In comparison, other transductive methods engender significant performance drops in at least one scenario.}
        \label{fig:summary_batch}
    \end{subfigure}
    
    \begin{subfigure}[b]{\linewidth}
        \centering
        \includegraphics[width=\linewidth]{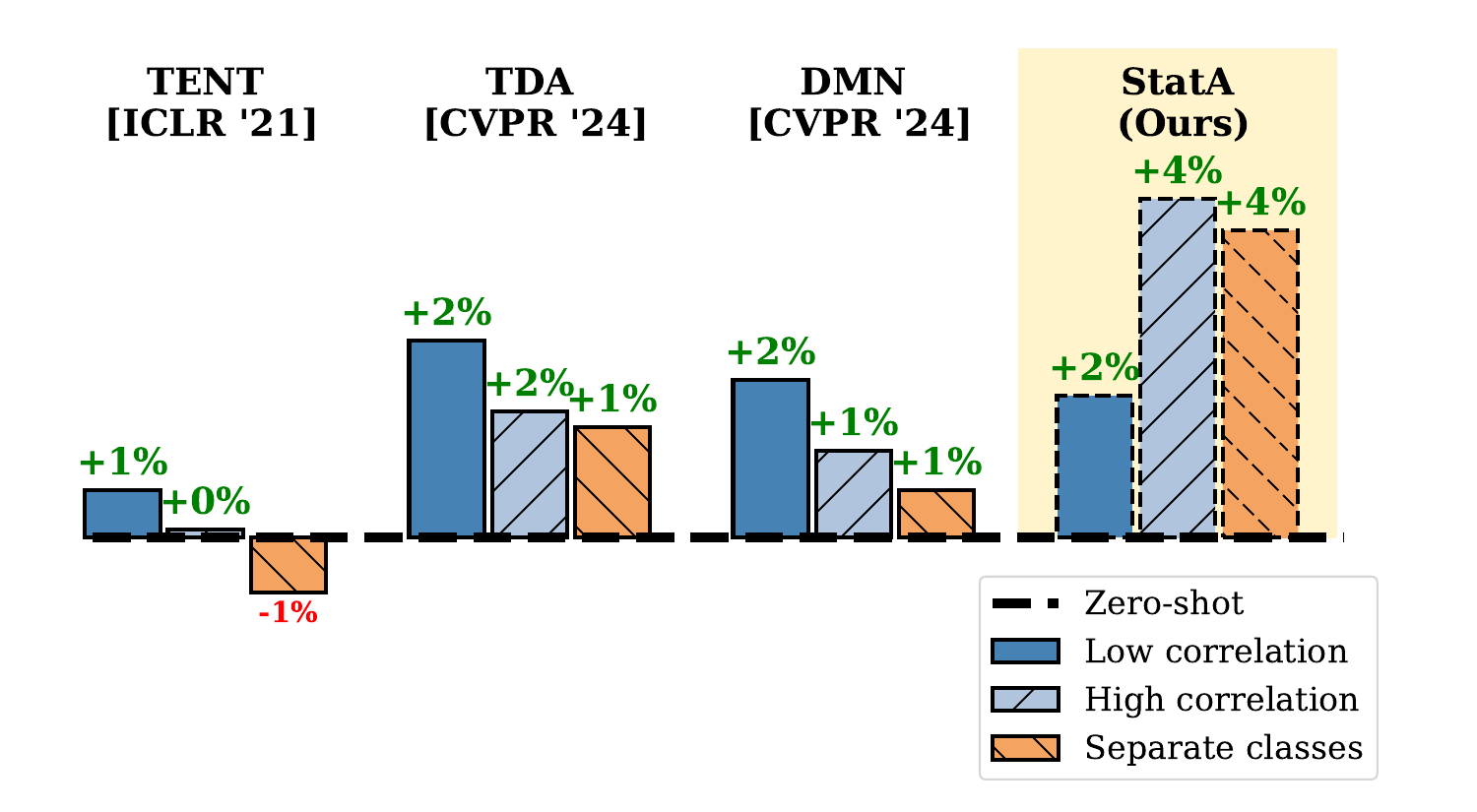}
        \caption{Stat${\cal A}$ shows strong performance when applied on streams of data, with \texttt{Low} or \texttt{High} correlation between batches, and when all the classes are appearing sequentially (\texttt{Separate}).}
        \label{fig:summary_online}
    \end{subfigure}
    \caption{We advocate for evaluating transductive or online TTA methods on more extensive \textit{realistic} scenarios.}
    \label{fig:summary}
\end{figure}
Vision-Language Models (VLMs) have introduced a powerful framework that connects images and texts, enabling models to adapt to new concepts without costly, 
human-labeled training data. This is achieved through a pre-training phase during which an image and its associated text description are aligned through contrastive learning. This enables zero-shot recognition, where novel images could be categorized by matching them to textual class descriptions, supporting tasks beyond traditional supervised learning. VLMs, such as CLIP \cite{radford2021learning}, have triggered wide interest, particularly in the few-shot adaptation setting~\cite{coop, zhang2022tip, khattak2023maple, ouali2023black, zanella2024low}, in which they have shown promising generalization capabilities using limited labeled data for each downstream task.   
Recently, test-time adaptation (TTA) approaches have enhanced the performances 
of these models without any supervision, including test-time augmentation on a single image \cite{shu2022tpt, Feng_2023_ICCV, zanella2024test, farina2024frustratingly}, transductive inference on batches \cite{Martin_2024_CVPR, Stojnic_2024_CVPR, zanella2024boosting}, and adaptation on streams~\cite{Karmanov_2024_CVPR, Zhang_2024_CVPR}.

\begin{figure}
    \centering
    \begin{subfigure}[b]{\linewidth}
        \centering
        \includegraphics[width=\linewidth]{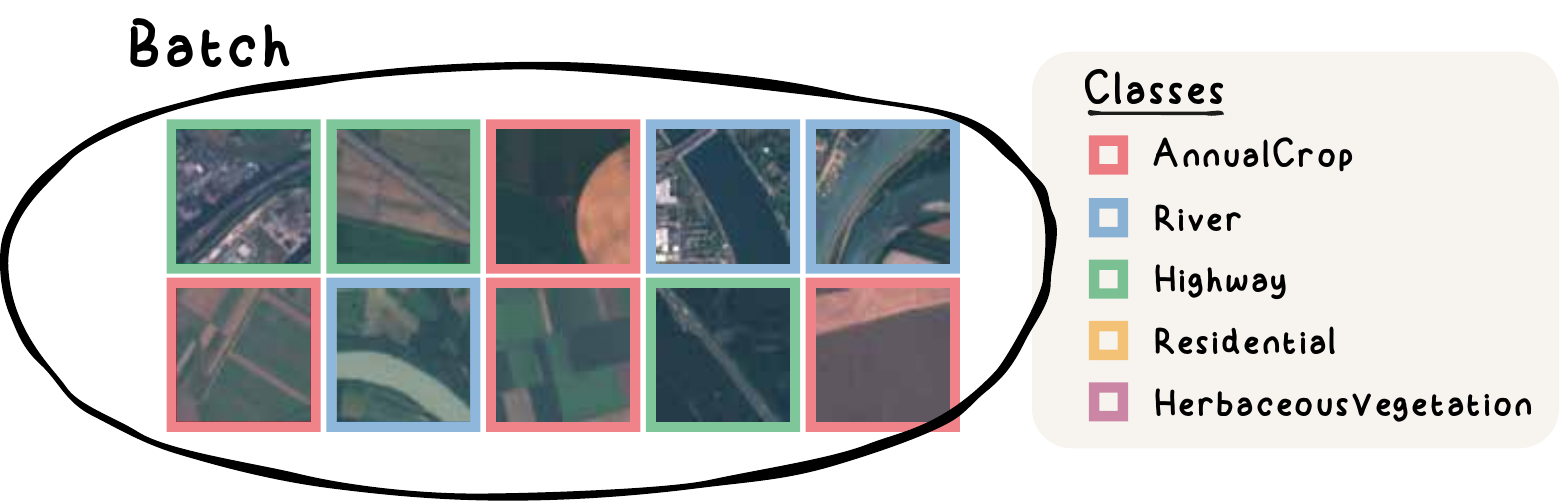}
        \caption{Satellite images of an area may not contain some classes. Images are taken from the EuroSAT~\cite{eurosat} dataset used in this paper.}
        \label{fig:realistic_eurosat}
    \end{subfigure}
    
    \vspace{0.8cm}
    \begin{subfigure}[b]{\linewidth}
        \centering
        \includegraphics[width=\linewidth]{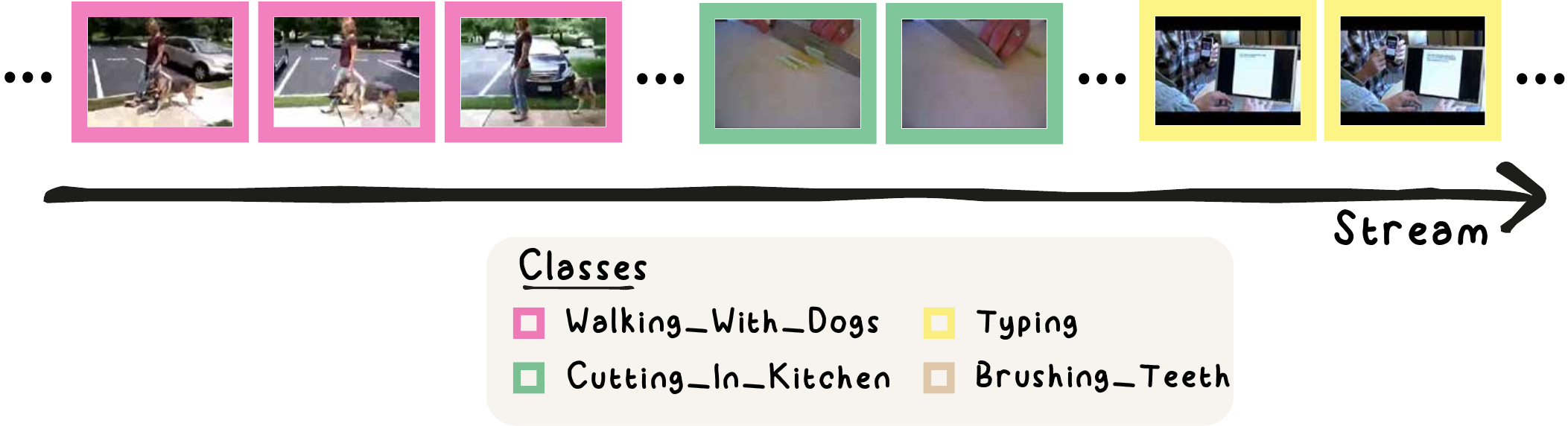}
        \caption{Video recordings may contain multiple frames of the same classes. Images are taken from the UCF101~\cite{ucf101} dataset used in this paper.}
        \label{fig:realistic_ucf101}
    \end{subfigure}
    \caption{Illustration of two \textit{realistic} scenarios: (a) batch adaptation with limited number of effective classes and (b) online test-time adaptation with a correlated, non-i.i.d. data stream.}
    \label{fig:realistic_scenarios}
\end{figure}

TTA is very popular in the vision community \cite{wangtent, chen2022contrastive, gong2022note, boudiaf2022parameter, yuan2023robust}, and this interest is now extending to VLMs~\cite{Karmanov_2024_CVPR, Zhang_2024_CVPR, Martin_2024_CVPR, Stojnic_2024_CVPR, zanella2024boosting}, where specific methods are required to fully exploit their open-vocabulary pre-training and zero-shot capabilities. However, while real-world scenarios frequently involve highly correlated incoming samples—such as patches from a satellite image or video recordings, as shown in Fig. \ref{fig:realistic_scenarios}—previous studies still lack a \textit{realistic} evaluation, i.e., one that envisions test-time class-distribution sampling to mimic what is encountered in real-life deployments. By considering a breadth of realistic scenarios, we found that existing TTA or transductive methods for VLMs are often highly biased towards settings with a small number of effective classes (i.e., the actual number of classes in the batch) \cite{Martin_2024_CVPR}, where the classes are uniformly distributed \cite{Stojnic_2024_CVPR, zanella2024boosting}, or where the streams of samples are i.i.d. \cite{Karmanov_2024_CVPR, Zhang_2024_CVPR}. In contrast, our method makes no assumptions as to the number of effective classes in the batch, and can handle a much wider range of scenarios, as depicted in Fig. \ref{fig:summary}.

More specifically, we develop two \textit{realistic} TTA evaluation settings for VLMs. First, we argue that adaptation methods operating on a single batch must demonstrate robustness when dealing with varying numbers of effective classes. While this aspect has been discussed recently in the context of VLMs \cite{Martin_2024_CVPR}, current methods are effective only within a narrow and specific range of class numbers. In contrast, we perform a comprehensive evaluation of methods over a broader range of effective class numbers. Secondly, online adaptation should be resilient to correlated batches, as already discussed in TTA for vision models \cite{gong2022note,boudiaf2022parameter, yuan2023robust}. Nevertheless, we show that current TTA and transductive methods designed for VLMs frequently compromise the models' initial zero-shot robustness across all these scenarios (detailed in Section \ref{sec:realistic}) in exchange for performance gains in well-specified conditions. In response to these limitations, we introduce Stat${\cal A}$, a more resilient transductive method. Stat${\cal A}$ integrates a novel regularization term specifically designed for VLMs, serving as a statistical anchor to preserve the initial textual class representations, particularly in low-data environments. Stat${\cal A}$ is highly efficient, processing thousands of samples within seconds.
\paragraph{Our contributions.} 
\begin{itemize}
    \item We highlight the current limitations of transductive and TTA techniques for VLMs, which often fail in more \textit{realistic} scenarios, i.e., conditions resembling real-life deployment where test data are unevenly distributed, with imbalanced classes and correlations between samples.
    \item Hence, we introduce two \textit{realistic} TTA evaluation settings for VLMs with (i) a variable number of effective classes for adaptation within a single batch, and (ii) non-i.i.d. batches of test samples in online adaptation settings.
    \item We propose Stat${\cal A}$, a versatile transductive algorithm with a simple yet effective regularization anchor term that allows for handling an arbitrary number of effective classes.
\end{itemize}


\section{Related work}
\paragraph{Transductive learning in VLMs.} Historically, transductive learning has been primarily explored within the few-shot learning literature. In this context, transduction leverages both the few labeled samples and the unlabeled test data, often outperforming inductive approaches \cite{liulearning, boudiaf2020information, ziko2020laplacian}. However, in the new multi-modal paradigm, additional supervision can be derived from the zero-shot predictions based on the joint representation of class textual descriptions and images. As a result, the scope of transductive learning now extends to unsupervised adaptation tasks. EM-Dirichlet \cite{Martin_2024_CVPR} operates within the prediction simplex, and proposes a maximum likelihood estimator of a Dirichlet distribution, while explicitly penalizing the number of the predicted classes. ZLaP \cite{Stojnic_2024_CVPR} constructs a similarity graph based on the cosine similarity of the representations, which is then used for label propagation, as in \cite{NIPS2003_87682805}. TransCLIP \cite{zanella2024boosting, zanella2024boostinghisto, elkhoury2024enhancing} models each class representation with a multivariate Gaussian distribution, while penalizing the deviation of the predicted labels from the zero-shot, text-driven softmax predictions.
While these transductive approaches have demonstrated significant gains, their applicability is limited by strong assumptions regarding the number of classes in each batch, as depicted in Figure \ref{fig:summary_batch}.

\paragraph{Test-time adaptation in VLMs.} There have been different focuses for adapting VLMs at test-time. A first distinction is the parameters they adapt, ranging from input prompts \cite{shu2022tpt, Feng_2023_ICCV, NEURIPS2023_cdd06402}, intermediate layers \cite{osowiechi2024watt}, adapters as memory banks in the embedding space \cite{Karmanov_2024_CVPR, Zhang_2024_CVPR}, to training-free methods \cite{zanella2024test, farina2024frustratingly}. The last two groups are sometimes referred to as black-box methods in the literature, meaning they only require access to the embedding space and not the internal state or parameters of the model \cite{ouali2023black, zanella2024test}. Surprisingly, while very convenient for practical applications (e.g., APIs), these more conservative methods often deliver excellent performance and robustness with minimal additional computational cost, whereas methods relying on prompt tuning typically require significantly more computation \cite{Karmanov_2024_CVPR}. In this work, we adopt the black-box assumption and focus on refining class representations directly within the embedding space, modeling them as a balanced mixture of multivariate Gaussian distributions \cite{zanella2024boosting, zanella2024boostinghisto, wang_hard_to_beat_2024}. A second distinction between adaptation methods lies in how they process the incoming data. One group of methods operates on a single image at a time but requires a large number of augmented views \cite{shu2022tpt, Feng_2023_ICCV, zanella2024test, farina2024frustratingly}. Some others operate in an online setting, iteratively constructing memory banks \cite{Karmanov_2024_CVPR, Zhang_2024_CVPR}. We observe that the latter approaches depend on uniformly distributed samples, as constructing complete memory banks for each class can be disrupted by correlated data streams, as summarized in Figure \ref{fig:summary_online}.

\section{Realistic test-time adaptation}
\label{sec:realistic}
The experimental settings currently envisioned to validate transduction or TTA methods for VLMs rely on unrealistic assumptions about data distribution.
In contrast, our work aims at tackling real-world deployment conditions, referred to as \textit{realistic} scenarios. Figure \ref{fig:realistic_scenarios} illustrates two of those scenarios of interest for TTA deployment. In our experiments, we have divided the study of these \textit{realistic} scenarios into two perspectives:
\paragraph{Batch \textit{realistic} scenarios.} In practical applications, batches often contain a limited number of effective classes, denoted as $K_{\text{eff}}$. For instance, a user will define a large panel of classes of interest meanwhile only a subset are actually present in the data at hand. In this first set-up of interest, each batch is processed independently, for example when processing large satellite images decomposed into patches (Figure \ref{fig:realistic_eurosat}). In this case, we vary the number of effective classes, simulating situations where a batch might not contain all the classes of interest, considering six scenarios: \texttt{Very Low} contains between 1 and 4 classes; \texttt{Low} contains between 2 and 10 classes; \texttt{Medium} contains between 5 and 25 classes; \texttt{High} contains between 25 and 50 classes; \texttt{Very High} contains between 50 and 100 classes; \texttt{All} includes all classes.
\paragraph{Online \textit{realistic} scenarios.} In our second perspective on the notion of realistic scenario, we examine streams of batches, akin to processing a sequence of images. In this case, it is known that real-world data distributions, such as those encountered in autonomous driving and human activity recognition (Figure \ref{fig:realistic_ucf101}), tend to exhibit temporal correlations. Following previous works, we control this correlation using a Dirichlet distribution \cite{gong2022note,boudiaf2022parameter, yuan2023robust}, denoted with $\gamma$; see Supplementary Material \ref{supplemental:implem} for more details. 

\section{Method}
Current TTA or transductive methods based on probabilistic clustering often hold implicit assumptions as to the statistics of the testing batch, such as uniform class distributions and low effective numbers of classes. As a result, they may have difficulty generalizing to diverse scenarios. 
We introduce a versatile method that can handle a breadth of scenarios, with varying batch sizes and numbers of effective classes, without 
any hyper-parameter tuning, while maintaining a strong overall performance. 
We derive our method as Maximum Likelihood Estimation (MLE) tailored to VLMs, with a novel Statistical Anchor (Stat$\cal A$) term that leverages the text-encoder knowledge, acting as a regularizer on the vision features' statistical parameters (i.e., the mean vectors and covariance matrices).   


\subsection{Formulation}

We seek to predict a label $k$ for each image $i$ in a query set $({\mathbf x}_i)_{1 \leq i \leq N}$, given pre-trained vision-encoded features ${\mathbf f}_i = \theta_{v}({\mathbf x}_i) \in {\mathbb R}^d$ and text embeddings ${\mathbf t}_k = \theta_{t}({\mathbf c}_k) \in {\mathbb R}^d$, where $\theta_{v}$ denotes the vision encoder, $\theta_{t}$ the text encoder, and ${\mathbf c}_k$ the text prompt representing class $k$, $k=1, \dots, K$. The total number of all possible classes is $K$, while the effective number of classes occurring in a given query set is $K_{\text{eff}}$, which might be lower but no greater than $K$. 

\paragraph{Regularized Maximum Likelihood Estimation.} The proposed method belongs to the broad family of soft probabilistic clustering approaches, which jointly estimate:
\begin{itemize}
\item Assignment vectors ${\mathbf z} = ({\mathbf z}_{i})_{1 \leq i \leq N}$ within the probability simplex $(\Delta_K)^N$. The $k$-th component $z_{i, k}$ of vector ${\mathbf z}_{i}$ yields the probability that the $i$-th sample is in class $k$. 
\item Statistical models ${\mathbf M} =({\mathbf M}_k)_{1 \leq k \leq K}$. Each ${\mathbf M}_k$ contains the set of parameters of a distribution modeling the features within class $k$, e.g., a mean vector and a covariance matrix in the case of multivariate Gaussian distributions.
\end{itemize}
In the context of transductive few-shot methods and TTA, several recent methods could be included in this family of clustering-based approaches, e.g., PADDLE \cite{martin2022towards}, Dirichlet \cite{Martin_2024_CVPR}, LaplacianShot \cite{ziko2020laplacian} and TransCLIP \cite{zanella2024boosting}, among others. These methods minimize Maximum Likelihood Estimation (MLE) objective functions of the following general form, alternating optimization updates w.r.t both the assignment variables in ${\mathbf z}$ and the model parameters in ${\mathbf M}$: 
\begin{equation}
\label{eq:rem_objective}
    {\cal L}_{\text{MLE}}({\mathbf z}; {\mathbf M}) = - \sum_{i=1}^N  {\mathbf z}_{i}^\top \log ({\mathbf p}_{i}) + {\cal R}({\mathbf z})
\end{equation}
where  ${\mathbf p}_{i} = (p_{i,k})_{1 \leq k \leq K} = (\text{Pr}({\mathbf f}_i|{\mathbf M}_k))_{1 \leq k \leq K}$ contains likelihood probability scores evaluating how likely is feature vector ${\mathbf f}_i$, given the $k$-th class parametric density ${\mathbf M}_k$, and ${\cal R}$ is a regularization term, characteristic of the method. These methods may also differ in how they model likelihoods $p_{i,k}$. For instance, TransCLIP \cite{zanella2024boosting} is based on multivariate Gaussian models, whereas the authors of \cite{Martin_2024_CVPR} used Dirichlet distributions. In fact, the first term in \eqref{eq:rem_objective} is the general log-likelihood model fitting objective, well established in the clustering literature as a probabilistic generalization of $K$-means\footnote{The $K$-means objective corresponds to multivariate Gaussian models, but with the simplifying assumption fixing all the covariance matrices to the identity matrix.} \cite{Kearns-UAI-97,martin2022towards}. This general term has an inherent bias to 
class-balanced clusters, a well-known fact in the clustering literature \cite{Kearns-UAI-97,Boykov-ICCV-05}. As for regularization term ${\cal R}({\mathbf z})$, there are several choices in the recent transductive learning literature, some of which also embed priors on class statistics, inducing biases towards specific scenarios. For instance, PADDLE \cite{martin2022towards} and Dirichlet \cite{Martin_2024_CVPR} explicitly penalize the number of non-empty clusters via a minimum description length (MDL) regularizer. While this regularizer could counter the class-balance 
bias in the log-likelihood term in \eqref{eq:rem_objective}, we found that these methods are strongly biased towards small numbers of effective classes; 
see Table~\ref{tab:big_batch_results} and \ref{tab:dataset_results}. LaplacianShot \cite{ziko2020laplacian} uses a Laplacian regularizer, encouraging samples with nearby 
representations (in the feature space) to have similar predictions. TransCLIP \cite{zanella2024boosting} adopts a combination of this Laplacian term 
and a Kullback-Leibler divergence regularizer, which penalizes the deviation of each assignment variable ${\mathbf p}_i$ from the zero-shot text-driven 
softmax prediction of the $i$-th sample. 

\paragraph{Proposed Statistical Anchor (StatA) term.} 

It is notable that all the state-of-the-art transductive learning methods mentioned in the previous section regularize the assignment variables in ${\mathbf z}$, via regularizer 
${\cal R}({\mathbf z})$,  but not statistical model parameters ${\mathbf M}$ that appear in the log-likelihood term in \eqref{eq:rem_objective}. However, in the context of VLMs, and as these learn from aligning vision and language representations, we argue that useful priors about those statistical parameters could be leveraged from the text-encoder knowledge. For instance, one could derive prototypes from the textual prompts. In the following, we introduce a Statistical Anchor (Stat${\cal A}$) term, which acts as regularizer on the model parameters using text-driven priors. As will be shown in our experiments, Stat${\cal A}$ brings important gains in robustness, yielding strong performances across all settings.

In the following, we develop Stat${\cal A}$ under the multivariate Gaussian assumption for models $({\mathbf M}_k)_{1 \leq k \leq K}$, but the concept could be extended to other density functions in the exponential family.     
Assume that the feature vectors ${\mathbf f}_i$ within a class $k$ are random variables following a multivariate Gaussian distribution $\mathcal{N}_k = \mathcal{N}(\boldsymbol{\mu}_k, \boldsymbol{\Sigma}_k)$, with a mean vector $\boldsymbol{\mu}_k$ and 
a diagonal covariance matrix $\boldsymbol{\Sigma}_k$, i.e., ${\mathbf M}_k = (\boldsymbol{\mu}_k, \boldsymbol{\Sigma}_k)$. Thus, the likelihood probabilities appearing in the general problem in \eqref{eq:rem_objective} read as follows:      
\begin{align}
p_{i,k} 
\propto \dfrac{1}{\sqrt{|\boldsymbol{\Sigma}_k}|} \exp \left(-{\frac {1}{2}}({\mathbf f}_i - \boldsymbol{\mu}_k)^\top \boldsymbol{\Sigma}_k^{-1}({\mathbf f}_i - \boldsymbol{\mu}_k) \right)
\end{align}
We introduce an additional penalty, which regularizes the statistical parameters of multivariate Gaussian $\mathcal{N}_k$. Specifically, we exploit the textual prompts to build an initial estimation of the mean vectors and covariance matrices, i.e., a fixed multivariate Gaussian distribution $\mathcal{N'}_k = \mathcal{N}(\boldsymbol{\mu'}_k,\boldsymbol{\Sigma'}_k)$, which serves as a statistical anchor (Stat${\cal A}$). Then, we add a Kullback-Leibler term that penalizes the overall objective when the multivariate Gaussian of class $k$ deviates from these initial mean vectors $\boldsymbol{\mu'}_k$ and covariance matrices $\boldsymbol{\Sigma'}_k$, which are kept fixed and not optimized upon:
\begin{align}
\label{kl-anchor}
    \mbox{KL} \left (\mathcal{N}'_k||\mathcal{N}_k \right ) &= \frac{1}{2} (  (\boldsymbol{\mu'}_k - \boldsymbol{\mu}_k)^\top \boldsymbol{\Sigma}_k^{-1}(\boldsymbol{\mu'}_k - \boldsymbol{\mu}_k) \notag \\ & + \Tr (\boldsymbol{\Sigma}_k^{-1}\boldsymbol{\Sigma'}_k) +\log \frac{|\boldsymbol{\Sigma}_k|}{| \boldsymbol{\Sigma'}_k|} - d).
\end{align} 
Our statistical anchor term could be added to the generic MLE clustering objective defined in Eq. \eqref{eq:rem_objective} and, hence, used in conjunction with any method befitting this general setting, provided that the likelihood probabilities are assumed to be multivariate Gaussian: 
\begin{equation}
\label{final-zero-shot-objective}
\begin{aligned}
{\cal L}_{{\cal A}}({\mathbf z}; \boldsymbol{\mu}, \boldsymbol{\Sigma}) =  {\cal L}_{\text{MLE}}({\mathbf z};  \boldsymbol{\mu}, \boldsymbol{\Sigma})
+ \alpha {\cal A}(\boldsymbol{\mu}, \boldsymbol{\Sigma}) 
\end{aligned}
\end{equation} 
where $\boldsymbol{\mu} = (\boldsymbol{\mu}_k)_{1 \leq k \leq K}$, $\boldsymbol{\Sigma}=(\boldsymbol{\Sigma}_k)_{1 \leq k \leq K}$, $\alpha$ is a non-negative constant, and our statistical anchor is given by: 
\begin{equation}
    {\cal A} (\boldsymbol{\mu}, \boldsymbol{\Sigma}) = \sum_{k=1}^{K} \mbox{KL} \left (\mathcal{N}'_k||\mathcal{N}_k \right )
\end{equation}
We compute mean vectors and a shared diagonal covariance matrix of the anchor (fixed) distribution $\mathcal{N}'_k$ from the text-encoder knowledge and zero-shot predictions as follows:
\begin{align}
\label{eq:parameter-initialization}
\boldsymbol{\mu}'_k = {\mathbf t}_k; \,  
\boldsymbol{\Sigma}' = \text{Diag} \left ( \frac{\sum_{i,k} \hat{y}_{i,k} ({\mathbf f}_i - \boldsymbol{\mu}_k)({\mathbf f}_i - \boldsymbol{\mu}_k)^\top}{\sum_{i,k} \hat{y}_{i,k} } \right )
\end{align}
where $\hat{y}_{i,k}$ denotes the zero-shot, text-driven prediction: 
\begin{equation}
\label{zero-shot-prediction}
\hat{y}_{i,k} = \frac{\exp ( \tau {\mathbf f}_i^\top {\mathbf t}_k)}{\sum_j \exp ( \tau {\mathbf f}_i^\top {\mathbf t}_j)}
\end{equation}



Our overall objective in \eqref{final-zero-shot-objective} depends on two types of variables, statistical parameters $(\boldsymbol{\mu}, \boldsymbol{\Sigma})$ and the assignments in ${\mathbf z}$. Therefore, we proceed with a block-coordinate descent scheme, which alternates two update steps, one fixing ${\mathbf z}$ and updating $(\boldsymbol{\mu}, \boldsymbol{\Sigma})$ in closed-form, and the other fixing the parameters and optimizing over ${\mathbf z}$ (see Algorithm \ref{alg:transclip++}).   
\begin{algorithm}[!t]
\caption{Stat${\cal A}$ procedure}
\label{alg:transclip++}
\SetAlgoLined
\SetKwComment{Comment}{$\triangleright$\ }{}
\KwIn{$\mathbf{f}_i$, $i=1, \dots, N$, $\mathbf{t}_k$, $k=1, \dots, K$, $\mathbf{\tau}$}

$\mathbf{z}_i \gets \hat{\mathbf{y}}_i = (y_{i,k})_{1 \leq k \leq K} \quad \forall i$ \Comment*{See Eq. \eqref{zero-shot-prediction}}

 $\boldsymbol{\mu}_k = \boldsymbol{\mu'}_k; \,  \boldsymbol{\Sigma}_k = \boldsymbol{\Sigma'} \quad \forall k $\Comment*{See Eq. \eqref{eq:parameter-initialization}}

\While{not converged}{
    \tcp{\textcolor{Green}{Iterative decoupled updates}}
    \For{$l = 1:\dots$}{
        Update $\mathbf{z}^{(l+1)}_i \quad \forall i$ \Comment*{See Eq. \eqref{eq:z_update}}
    }
    \tcp{\textcolor{Green}{Closed-form updates}}
    Update $\beta_k \quad \forall k$\Comment*{See Eq. \eqref{eq:hard_beta_k}}
    
    Update $\boldsymbol{\mu}_k$ and $\boldsymbol{\Sigma}_k$  $\quad \forall k$ \Comment*{See Eq. \eqref{regularized-updates}}
}
\textbf{return $\mathbf{z}$}\

\end{algorithm}
\subsection{Regularized updates of the parameters}
\label{subsec:regularized}
When the variables in ${\mathbf z}$ are fixed, notice that the proposed objective function in \eqref{final-zero-shot-objective} is strictly convex w.r.t each $\boldsymbol{\mu}_k$ and each 
$\boldsymbol{\Sigma}_k$, $k=1, \dots, K$, for any non-negative $\alpha$. Therefore, by setting the gradient of \eqref{final-zero-shot-objective} with respect to each of these statistical variables equal to zero (detailed derivation in the Appendix), we obtain the following closed-form updates:
\begin{align}
\label{regularized-updates}
\boldsymbol{\mu}_k &= \beta_k \boldsymbol{v}_k + (1 - \beta_k) \boldsymbol{\mu'}_k \nonumber \\
\boldsymbol{\Sigma}_k &= \beta_k \boldsymbol{T}_k + (1-\beta_k)(\boldsymbol{\Sigma'} + \text{Diag}((\boldsymbol{\mu'}_k - \boldsymbol{\mu}_k)^2 )
\end{align}
where scalar $\beta_k \in [0, 1]$ is given by:
    $\beta_k = \frac{\sum_{i=1}^N z_{i,k}}{\sum_{i=1}^N z_{i,k} + \alpha}$,
and $\boldsymbol{v}_k$ and $\boldsymbol{T}_k$ are, respectively, the sample mean vector and diagonal covariance matrix of class $k$:
\begin{equation}
\label{eq:closed-form-gmm}
    \boldsymbol{v}_k = \frac{\sum_{i=1}^N z_{i,k}  {\mathbf f}_i}{ \sum_{i=1}^N z_{i,k}}; \quad \boldsymbol{T}_k = \frac{\sum_{i=1}^N  z_{i,k} \text{Diag}(({\mathbf f}_i - \boldsymbol{\mu}_k)^2)}{\sum_{i=1}^N  z_{i,k}}
\end{equation}




\paragraph{Interpretation.} Updates in \eqref{regularized-updates} enables a clear interpretation. Indeed,
the sample mean and covariance updates in Eq.~\eqref{eq:closed-form-gmm} are the standard MLE estimates optimizing \eqref{eq:rem_objective}, which corresponds to the case $\alpha=0$ in our objective in \eqref{final-zero-shot-objective} (i.e., no anchor term). The updates in \eqref{regularized-updates}, $\boldsymbol{\mu}_k$ and $\boldsymbol{\Sigma_k}$, are a convex combination of these MLE estimates and a term dependent on the statistical anchor, with scalar $\beta_k \in [0, 1]$ controlling the combination. Notice that the larger the number of samples predicted to belong to class $k$, the larger the value of this scalar. Therefore, the convex combination in \eqref{regularized-updates} gets closer to the statistical anchor when few or no samples are predicted as belonging to class $K$, and closer to the standard MLE estimates in ~\eqref{eq:closed-form-gmm} otherwise. This makes sense as a small number of samples may not be sufficient to estimate reliably the mean and covariance statistics. Weighting factor $\beta_k$ depends on $\alpha$, which is a hyper-parameter. While we have found that the straightforward choice $\alpha = 1$ (used in all experiments) works well across various settings, one could further tune it depending, for example, on the quality of the chosen anchor (see Figure \ref{fig:ablation_anchor} of the ablation study). 
\paragraph{Implementation.} In our numerical implementation of the algorithm, we modified the weighting factor $\beta_k$ by replacing the soft assignment predictions with hard ones on the vertices of the simplex, for a better estimation of the predicted cardinality (i.e., number of samples) of each class $k$: 
\begin{equation}
\beta_k = \frac{\sum_{i=1}^N \mathbbm{1}\left[k = \argmax_r z_{i,r} \right]}{\sum_{i=1}^N \mathbbm{1}\left[k = \argmax_r z_{i,r} \right] + \alpha}
\label{eq:hard_beta_k}
\end{equation}
 We found \eqref{eq:hard_beta_k} to be more robust to the noise induced by residuals in components other than the one with the maximum probability in ${\mathbf z}_i$ (see Table \ref{tab:ablation_anchor_strategy} of the ablation study).

\definecolor{AverageColor}{RGB}{255, 237, 206}
\definecolor{AverageDarkerColor}{RGB}{236, 227, 211}

\newcolumntype{n}{>{\columncolor{AverageColor}} p{1.7cm}}

\aboverulesep = 0.2mm 
\belowrulesep = 0.2mm 

\begin{table*}
\centering
\caption{Comparison of different batch sizes and ranges of effective class numbers. The best average performance for each configuration is highlighted in \textbf{bold}, while the second-best is indicated with \underline{underline}. Each reported performance is averaged over 1,000 tasks.}
\label{tab:batch_results}
\begin{subtable}{\linewidth}
\centering
\caption{Three scenarios with a batch size of 64: \texttt{Very Low} (1–4 effective classes per task), \texttt{Low} (2–10 classes), and \texttt{Medium} (5–25 classes).}
\label{tab:small_batch_results}
\resizebox{\textwidth}{!}{%
\setlength\dashlinedash{0.2pt}
\setlength\dashlinegap{1.5pt}
\setlength\arrayrulewidth{0.3pt}
\renewcommand{\arraystretch}{1.}
    \begin{tabular}{p{1.7cm}l|nccccccccccc}
      \toprule
$K_{\text{eff}}$ & Method & \textsc{\fontsize{11}{10}\selectfont\textcolor{black}{Average}}& \rotatebox[origin=c]{45}{ImageNet} & \rotatebox[origin=c]{45}{SUN397} & \rotatebox[origin=c]{45}{Aircraft} & \rotatebox[origin=c]{45}{EuroSAT} & \rotatebox[origin=c]{45}{StanfordCars} & \rotatebox[origin=c]{45}{Food101} & \rotatebox[origin=c]{45}{Pets} & \rotatebox[origin=c]{45}{Flower102} & \rotatebox[origin=c]{45}{Caltech101} & \rotatebox[origin=c]{45}{DTD} & \rotatebox[origin=c]{45}{UCF101}  \\
\midrule
 & CLIP & $\textcolor{black}{65.2}$ & $\text{66.6}$ & $\text{62.5}$ & $\text{24.7}$ & $\text{48.3}$ & $\text{65.6}$ & $\text{85.9}$ & $\text{89.1}$ & $\text{70.7}$ & $\text{93.2}$ & $\text{43.5}$ & $\text{67.5}$  \\
 & MTA &  $\textcolor{black}{66.6}_{\textcolor{darkerGreen}{\raisebox{2pt}{\normalsize+\textbf{1.3}}}}$ & $\text{69.3}_{\textcolor{darkerGreen}{\raisebox{2pt}{\normalsize+2.7}}}$ & $\text{64.8}_{\textcolor{darkerGreen}{\raisebox{2pt}{\normalsize+2.3}}}$ & $\text{27.4}_{\textcolor{darkerGreen}{\raisebox{2pt}{\normalsize+2.7}}}$ & $\text{46.9}_{\textcolor{red}{\raisebox{2pt}{\normalsize-1.4}}}$ & $\text{68.0}_{\textcolor{darkerGreen}{\raisebox{2pt}{\normalsize+2.4}}}$ & $\text{87.2}_{\textcolor{darkerGreen}{\raisebox{2pt}{\normalsize+1.3}}}$ & $\text{89.4}_{\textcolor{darkerGreen}{\raisebox{2pt}{\normalsize+0.3}}}$ & $\text{71.7}_{\textcolor{darkerGreen}{\raisebox{2pt}{\normalsize+1.0}}}$ & $\text{94.0}_{\textcolor{darkerGreen}{\raisebox{2pt}{\normalsize+0.8}}}$ & $\text{44.4}_{\textcolor{darkerGreen}{\raisebox{2pt}{\normalsize+0.9}}}$ & $\text{69.0}_{\textcolor{darkerGreen}{\raisebox{2pt}{\normalsize+1.5}}}$  \\
\midrule
\multirow{4}{*}{\makecell{\texttt{Very Low} \\ (1--4)}} & Dirichlet & $  \textcolor{black}{\underline{68.5}}_{\textcolor{darkerGreen}{\raisebox{2pt}{\normalsize+\textbf{3.3}}}}$ & $\text{79.2}_{\textcolor{darkerGreen}{\raisebox{2pt}{\normalsize+12.6}}}$ & $\text{75.7}_{\textcolor{darkerGreen}{\raisebox{2pt}{\normalsize+13.2}}}$ & $\text{28.2}_{\textcolor{darkerGreen}{\raisebox{2pt}{\normalsize+3.5}}}$ & $\text{47.2}_{\textcolor{red}{\raisebox{2pt}{\normalsize-1.1}}}$ & $\text{68.2}_{\textcolor{darkerGreen}{\raisebox{2pt}{\normalsize+2.6}}}$ & $\text{88.1}_{\textcolor{darkerGreen}{\raisebox{2pt}{\normalsize+2.2}}}$ & $\text{87.5}_{\textcolor{red}{\raisebox{2pt}{\normalsize-1.6}}}$ & $\text{71.2}_{\textcolor{darkerGreen}{\raisebox{2pt}{\normalsize+0.5}}}$ & $\text{88.8}_{\textcolor{red}{\raisebox{2pt}{\normalsize-4.4}}}$ & $\text{50.3}_{\textcolor{darkerGreen}{\raisebox{2pt}{\normalsize+6.8}}}$ & $\text{69.0}_{\textcolor{darkerGreen}{\raisebox{2pt}{\normalsize+1.5}}}$  \\
 & ZLaP & $  \textcolor{black}{27.5}_{\textcolor{red}{\raisebox{2pt}{\normalsize-\textbf{37.8}}}}$ & $\text{14.5}_{\textcolor{red}{\raisebox{2pt}{\normalsize-52.1}}}$ & $\text{13.0}_{\textcolor{red}{\raisebox{2pt}{\normalsize-49.5}}}$ & $\text{8.4}_{\textcolor{red}{\raisebox{2pt}{\normalsize-16.3}}}$ & $\text{36.6}_{\textcolor{red}{\raisebox{2pt}{\normalsize-11.7}}}$ & $\text{23.7}_{\textcolor{red}{\raisebox{2pt}{\normalsize-41.9}}}$ & $\text{31.9}_{\textcolor{red}{\raisebox{2pt}{\normalsize-54.0}}}$ & $\text{57.0}_{\textcolor{red}{\raisebox{2pt}{\normalsize-32.1}}}$ & $\text{22.4}_{\textcolor{red}{\raisebox{2pt}{\normalsize-48.3}}}$ & $\text{52.4}_{\textcolor{red}{\raisebox{2pt}{\normalsize-40.8}}}$ & $\text{13.0}_{\textcolor{red}{\raisebox{2pt}{\normalsize-30.5}}}$ & $\text{29.2}_{\textcolor{red}{\raisebox{2pt}{\normalsize-38.3}}}$  \\
 & TransCLIP & $  \textcolor{black}{38.9}_{\textcolor{red}{\raisebox{2pt}{\normalsize-\textbf{26.3}}}}$ & $\text{21.6}_{\textcolor{red}{\raisebox{2pt}{\normalsize-45.0}}}$ & $\text{21.1}_{\textcolor{red}{\raisebox{2pt}{\normalsize-41.4}}}$ & $\text{11.6}_{\textcolor{red}{\raisebox{2pt}{\normalsize-13.1}}}$ & $\text{45.1}_{\textcolor{red}{\raisebox{2pt}{\normalsize-3.2}}}$ & $\text{34.7}_{\textcolor{red}{\raisebox{2pt}{\normalsize-30.9}}}$ & $\text{59.2}_{\textcolor{red}{\raisebox{2pt}{\normalsize-26.7}}}$ & $\text{72.4}_{\textcolor{red}{\raisebox{2pt}{\normalsize-16.7}}}$ & $\text{36.4}_{\textcolor{red}{\raisebox{2pt}{\normalsize-34.3}}}$ & $\text{62.3}_{\textcolor{red}{\raisebox{2pt}{\normalsize-30.9}}}$ & $\text{26.1}_{\textcolor{red}{\raisebox{2pt}{\normalsize-17.4}}}$ & $\text{37.7}_{\textcolor{red}{\raisebox{2pt}{\normalsize-29.8}}}$  \\
 & \cellcolor{LightGray} Stat${\cal A}$ & \cellcolor{AverageDarkerColor} $  \textcolor{black}{\textbf{70.4}}_{\textcolor{darkerGreen}{\raisebox{2pt}{\normalsize+\textbf{5.1}}}}$ & \cellcolor{LightGray} $\text{72.9}_{\textcolor{darkerGreen}{\raisebox{2pt}{\normalsize+6.3}}}$ & \cellcolor{LightGray} $\text{66.0}_{\textcolor{darkerGreen}{\raisebox{2pt}{\normalsize+3.5}}}$ & \cellcolor{LightGray} $\text{29.3}_{\textcolor{darkerGreen}{\raisebox{2pt}{\normalsize+4.6}}}$ & \cellcolor{LightGray} $\text{56.8}_{\textcolor{darkerGreen}{\raisebox{2pt}{\normalsize+8.5}}}$ & \cellcolor{LightGray} $\text{76.2}_{\textcolor{darkerGreen}{\raisebox{2pt}{\normalsize+10.6}}}$ & \cellcolor{LightGray} $\text{90.3}_{\textcolor{darkerGreen}{\raisebox{2pt}{\normalsize+4.4}}}$ & \cellcolor{LightGray} $\text{95.5}_{\textcolor{darkerGreen}{\raisebox{2pt}{\normalsize+6.4}}}$ & \cellcolor{LightGray} $\text{77.6}_{\textcolor{darkerGreen}{\raisebox{2pt}{\normalsize+6.9}}}$ & \cellcolor{LightGray} $\text{93.0}_{\textcolor{red}{\raisebox{2pt}{\normalsize-0.2}}}$ & \cellcolor{LightGray} $\text{46.1}_{\textcolor{darkerGreen}{\raisebox{2pt}{\normalsize+2.6}}}$ & \cellcolor{LightGray} $\text{70.2}_{\textcolor{darkerGreen}{\raisebox{2pt}{\normalsize+2.7}}}$  \\
\midrule
\multirow{4}{*}{\makecell{\texttt{Low} \\ (2--10)}} & Dirichlet & $  \textcolor{black}{\textbf{70.3}}_{\textcolor{darkerGreen}{\raisebox{2pt}{\normalsize+\textbf{5.1}}}}$ & $\text{80.1}_{\textcolor{darkerGreen}{\raisebox{2pt}{\normalsize+13.5}}}$ & $\text{78.0}_{\textcolor{darkerGreen}{\raisebox{2pt}{\normalsize+15.5}}}$ & $\text{28.1}_{\textcolor{darkerGreen}{\raisebox{2pt}{\normalsize+3.4}}}$ & $\text{43.5}_{\textcolor{red}{\raisebox{2pt}{\normalsize-4.8}}}$ & $\text{71.5}_{\textcolor{darkerGreen}{\raisebox{2pt}{\normalsize+5.9}}}$ & $\text{92.3}_{\textcolor{darkerGreen}{\raisebox{2pt}{\normalsize+6.4}}}$ & $\text{92.7}_{\textcolor{darkerGreen}{\raisebox{2pt}{\normalsize+3.6}}}$ & $\text{74.7}_{\textcolor{darkerGreen}{\raisebox{2pt}{\normalsize+4.0}}}$ & $\text{93.0}_{\textcolor{red}{\raisebox{2pt}{\normalsize-0.2}}}$ & $\text{48.9}_{\textcolor{darkerGreen}{\raisebox{2pt}{\normalsize+5.4}}}$ & $\text{70.9}_{\textcolor{darkerGreen}{\raisebox{2pt}{\normalsize+3.4}}}$  \\
 & ZLaP & $  \textcolor{black}{35.2}_{\textcolor{red}{\raisebox{2pt}{\normalsize-\textbf{30.0}}}}$ & $\text{19.1}_{\textcolor{red}{\raisebox{2pt}{\normalsize-47.5}}}$ & $\text{19.0}_{\textcolor{red}{\raisebox{2pt}{\normalsize-43.5}}}$ & $\text{12.0}_{\textcolor{red}{\raisebox{2pt}{\normalsize-12.7}}}$ & $\text{46.4}_{\textcolor{red}{\raisebox{2pt}{\normalsize-1.9}}}$ & $\text{27.9}_{\textcolor{red}{\raisebox{2pt}{\normalsize-37.7}}}$ & $\text{43.5}_{\textcolor{red}{\raisebox{2pt}{\normalsize-42.4}}}$ & $\text{66.6}_{\textcolor{red}{\raisebox{2pt}{\normalsize-22.5}}}$ & $\text{31.3}_{\textcolor{red}{\raisebox{2pt}{\normalsize-39.4}}}$ & $\text{60.8}_{\textcolor{red}{\raisebox{2pt}{\normalsize-32.4}}}$ & $\text{22.4}_{\textcolor{red}{\raisebox{2pt}{\normalsize-21.1}}}$ & $\text{38.7}_{\textcolor{red}{\raisebox{2pt}{\normalsize-28.8}}}$  \\
 & TransCLIP & $  \textcolor{black}{40.4}_{\textcolor{red}{\raisebox{2pt}{\normalsize-\textbf{24.8}}}}$ & $\text{20.3}_{\textcolor{red}{\raisebox{2pt}{\normalsize-46.3}}}$ & $\text{22.4}_{\textcolor{red}{\raisebox{2pt}{\normalsize-40.1}}}$ & $\text{14.3}_{\textcolor{red}{\raisebox{2pt}{\normalsize-10.4}}}$ & $\text{53.9}_{\textcolor{darkerGreen}{\raisebox{2pt}{\normalsize+5.6}}}$ & $\text{30.8}_{\textcolor{red}{\raisebox{2pt}{\normalsize-34.8}}}$ & $\text{55.6}_{\textcolor{red}{\raisebox{2pt}{\normalsize-30.3}}}$ & $\text{69.4}_{\textcolor{red}{\raisebox{2pt}{\normalsize-19.7}}}$ & $\text{40.9}_{\textcolor{red}{\raisebox{2pt}{\normalsize-29.8}}}$ & $\text{64.6}_{\textcolor{red}{\raisebox{2pt}{\normalsize-28.6}}}$ & $\text{31.6}_{\textcolor{red}{\raisebox{2pt}{\normalsize-11.9}}}$ & $\text{40.9}_{\textcolor{red}{\raisebox{2pt}{\normalsize-26.6}}}$  \\
 & \cellcolor{LightGray} Stat${\cal A}$ & \cellcolor{AverageDarkerColor} $  \textcolor{black}{\underline{69.3}}_{\textcolor{darkerGreen}{\raisebox{2pt}{\normalsize+\textbf{4.1}}}}$ & \cellcolor{LightGray} $\text{72.8}_{\textcolor{darkerGreen}{\raisebox{2pt}{\normalsize+6.2}}}$ & \cellcolor{LightGray} $\text{66.9}_{\textcolor{darkerGreen}{\raisebox{2pt}{\normalsize+4.4}}}$ & \cellcolor{LightGray} $\text{27.7}_{\textcolor{darkerGreen}{\raisebox{2pt}{\normalsize+3.0}}}$ & \cellcolor{LightGray} $\text{51.3}_{\textcolor{darkerGreen}{\raisebox{2pt}{\normalsize+3.0}}}$ & \cellcolor{LightGray} $\text{73.5}_{\textcolor{darkerGreen}{\raisebox{2pt}{\normalsize+7.9}}}$ & \cellcolor{LightGray} $\text{89.5}_{\textcolor{darkerGreen}{\raisebox{2pt}{\normalsize+3.6}}}$ & \cellcolor{LightGray} $\text{93.7}_{\textcolor{darkerGreen}{\raisebox{2pt}{\normalsize+4.6}}}$ & \cellcolor{LightGray} $\text{76.6}_{\textcolor{darkerGreen}{\raisebox{2pt}{\normalsize+5.9}}}$ & \cellcolor{LightGray} $\text{93.6}_{\textcolor{darkerGreen}{\raisebox{2pt}{\normalsize+0.4}}}$ & \cellcolor{LightGray} $\text{46.9}_{\textcolor{darkerGreen}{\raisebox{2pt}{\normalsize+3.4}}}$ & \cellcolor{LightGray} $\text{69.6}_{\textcolor{darkerGreen}{\raisebox{2pt}{\normalsize+2.1}}}$  \\
\midrule
\multirow{4}{*}{\makecell{\texttt{Medium} \\ (5--25)}} & Dirichlet & $  \textcolor{black}{\textbf{67.5}}_{\textcolor{darkerGreen}{\raisebox{2pt}{\normalsize+\textbf{2.2}}}}$ & $\text{77.7}_{\textcolor{darkerGreen}{\raisebox{2pt}{\normalsize+11.1}}}$ & $\text{72.9}_{\textcolor{darkerGreen}{\raisebox{2pt}{\normalsize+10.4}}}$ & $\text{26.1}_{\textcolor{darkerGreen}{\raisebox{2pt}{\normalsize+1.4}}}$ & $\text{38.6}_{\textcolor{red}{\raisebox{2pt}{\normalsize-9.7}}}$ & $\text{71.6}_{\textcolor{darkerGreen}{\raisebox{2pt}{\normalsize+6.0}}}$ & $\text{90.8}_{\textcolor{darkerGreen}{\raisebox{2pt}{\normalsize+4.9}}}$ & $\text{88.4}_{\textcolor{red}{\raisebox{2pt}{\normalsize-0.7}}}$ & $\text{71.5}_{\textcolor{darkerGreen}{\raisebox{2pt}{\normalsize+0.8}}}$ & $\text{93.7}_{\textcolor{darkerGreen}{\raisebox{2pt}{\normalsize+0.5}}}$ & $\text{42.9}_{\textcolor{red}{\raisebox{2pt}{\normalsize-0.6}}}$ & $\text{67.8}_{\textcolor{darkerGreen}{\raisebox{2pt}{\normalsize+0.3}}}$  \\
 & ZLaP & $  \textcolor{black}{44.7}_{\textcolor{red}{\raisebox{2pt}{\normalsize-\textbf{20.6}}}}$ & $\text{29.0}_{\textcolor{red}{\raisebox{2pt}{\normalsize-37.6}}}$ & $\text{27.9}_{\textcolor{red}{\raisebox{2pt}{\normalsize-34.6}}}$ & $\text{16.5}_{\textcolor{red}{\raisebox{2pt}{\normalsize-8.2}}}$ & $\text{49.0}_{\textcolor{darkerGreen}{\raisebox{2pt}{\normalsize+0.7}}}$ & $\text{36.0}_{\textcolor{red}{\raisebox{2pt}{\normalsize-29.6}}}$ & $\text{59.1}_{\textcolor{red}{\raisebox{2pt}{\normalsize-26.8}}}$ & $\text{76.4}_{\textcolor{red}{\raisebox{2pt}{\normalsize-12.7}}}$ & $\text{42.9}_{\textcolor{red}{\raisebox{2pt}{\normalsize-27.8}}}$ & $\text{72.0}_{\textcolor{red}{\raisebox{2pt}{\normalsize-21.2}}}$ & $\text{32.0}_{\textcolor{red}{\raisebox{2pt}{\normalsize-11.5}}}$ & $\text{50.3}_{\textcolor{red}{\raisebox{2pt}{\normalsize-17.2}}}$  \\
 & TransCLIP & $  \textcolor{black}{42.7}_{\textcolor{red}{\raisebox{2pt}{\normalsize-\textbf{22.5}}}}$ & $\text{15.5}_{\textcolor{red}{\raisebox{2pt}{\normalsize-51.1}}}$ & $\text{22.8}_{\textcolor{red}{\raisebox{2pt}{\normalsize-39.7}}}$ & $\text{17.0}_{\textcolor{red}{\raisebox{2pt}{\normalsize-7.7}}}$ & $\text{58.2}_{\textcolor{darkerGreen}{\raisebox{2pt}{\normalsize+9.9}}}$ & $\text{32.9}_{\textcolor{red}{\raisebox{2pt}{\normalsize-32.7}}}$ & $\text{56.3}_{\textcolor{red}{\raisebox{2pt}{\normalsize-29.6}}}$ & $\text{72.6}_{\textcolor{red}{\raisebox{2pt}{\normalsize-16.5}}}$ & $\text{45.0}_{\textcolor{red}{\raisebox{2pt}{\normalsize-25.7}}}$ & $\text{65.6}_{\textcolor{red}{\raisebox{2pt}{\normalsize-27.6}}}$ & $\text{37.5}_{\textcolor{red}{\raisebox{2pt}{\normalsize-6.0}}}$ & $\text{46.5}_{\textcolor{red}{\raisebox{2pt}{\normalsize-21.0}}}$  \\
 & \cellcolor{LightGray} Stat${\cal A}$ & \cellcolor{AverageDarkerColor} $  \textcolor{black}{\underline{67.4}}_{\textcolor{darkerGreen}{\raisebox{2pt}{\normalsize+\textbf{2.2}}}}$ & \cellcolor{LightGray} $\text{70.7}_{\textcolor{darkerGreen}{\raisebox{2pt}{\normalsize+4.1}}}$ & \cellcolor{LightGray} $\text{65.3}_{\textcolor{darkerGreen}{\raisebox{2pt}{\normalsize+2.8}}}$ & \cellcolor{LightGray} $\text{26.0}_{\textcolor{darkerGreen}{\raisebox{2pt}{\normalsize+1.3}}}$ & \cellcolor{LightGray} $\text{45.0}_{\textcolor{red}{\raisebox{2pt}{\normalsize-3.3}}}$ & \cellcolor{LightGray} $\text{71.1}_{\textcolor{darkerGreen}{\raisebox{2pt}{\normalsize+5.5}}}$ & \cellcolor{LightGray} $\text{88.2}_{\textcolor{darkerGreen}{\raisebox{2pt}{\normalsize+2.3}}}$ & \cellcolor{LightGray} $\text{90.8}_{\textcolor{darkerGreen}{\raisebox{2pt}{\normalsize+1.7}}}$ & \cellcolor{LightGray} $\text{73.7}_{\textcolor{darkerGreen}{\raisebox{2pt}{\normalsize+3.0}}}$ & \cellcolor{LightGray} $\text{93.9}_{\textcolor{darkerGreen}{\raisebox{2pt}{\normalsize+0.7}}}$ & \cellcolor{LightGray} $\text{47.5}_{\textcolor{darkerGreen}{\raisebox{2pt}{\normalsize+4.0}}}$ & \cellcolor{LightGray} $\text{69.1}_{\textcolor{darkerGreen}{\raisebox{2pt}{\normalsize+1.6}}}$  \\
\midrule
\end{tabular}}
\end{subtable}
\\
\vspace{0.5cm}
\begin{subtable}{\linewidth}
\centering
\caption{Three scenarios with a batch size of 1000: \texttt{Medium} (5–25 effective classes per task), \texttt{High} (25–50 classes), and \texttt{Very High} (50-100 classes).}
\label{tab:big_batch_results}
\resizebox{\textwidth}{!}{%
\setlength\dashlinedash{0.2pt}
\setlength\dashlinegap{1.5pt}
\setlength\arrayrulewidth{0.3pt}
\renewcommand{\arraystretch}{1.}
    \begin{tabular}{p{1.7cm}l|nccccccccccc}
      \toprule
$K_{\text{eff}}$ & Method & \textsc{\fontsize{11}{10}\selectfont\textcolor{black}{Average}}& \rotatebox[origin=c]{45}{ImageNet} & \rotatebox[origin=c]{45}{SUN397} & \rotatebox[origin=c]{45}{Aircraft} & \rotatebox[origin=c]{45}{EuroSAT} & \rotatebox[origin=c]{45}{StanfordCars} & \rotatebox[origin=c]{45}{Food101} & \rotatebox[origin=c]{45}{Pets} & \rotatebox[origin=c]{45}{Flower102} & \rotatebox[origin=c]{45}{Caltech101} & \rotatebox[origin=c]{45}{DTD} & \rotatebox[origin=c]{45}{UCF101}  \\
\midrule
 & CLIP & $\textcolor{black}{65.2}$ & $\text{66.6}$ & $\text{62.5}$ & $\text{24.7}$ & $\text{48.3}$ & $\text{65.6}$ & $\text{85.9}$ & $\text{89.1}$ & $\text{70.7}$ & $\text{93.2}$ & $\text{43.5}$ & $\text{67.5}$  \\
 & MTA &  $\textcolor{black}{66.6}_{\textcolor{darkerGreen}{\raisebox{2pt}{\normalsize+\textbf{1.3}}}}$ & $\text{69.3}_{\textcolor{darkerGreen}{\raisebox{2pt}{\normalsize+2.7}}}$ & $\text{64.8}_{\textcolor{darkerGreen}{\raisebox{2pt}{\normalsize+2.3}}}$ & $\text{27.4}_{\textcolor{darkerGreen}{\raisebox{2pt}{\normalsize+2.7}}}$ & $\text{46.9}_{\textcolor{red}{\raisebox{2pt}{\normalsize-1.4}}}$ & $\text{68.0}_{\textcolor{darkerGreen}{\raisebox{2pt}{\normalsize+2.4}}}$ & $\text{87.2}_{\textcolor{darkerGreen}{\raisebox{2pt}{\normalsize+1.3}}}$ & $\text{89.4}_{\textcolor{darkerGreen}{\raisebox{2pt}{\normalsize+0.3}}}$ & $\text{71.7}_{\textcolor{darkerGreen}{\raisebox{2pt}{\normalsize+1.0}}}$ & $\text{94.0}_{\textcolor{darkerGreen}{\raisebox{2pt}{\normalsize+0.8}}}$ & $\text{44.4}_{\textcolor{darkerGreen}{\raisebox{2pt}{\normalsize+0.9}}}$ & $\text{69.0}_{\textcolor{darkerGreen}{\raisebox{2pt}{\normalsize+1.5}}}$  \\
\midrule
\multirow{4}{*}{\makecell{\texttt{Medium} \\ (5--25)}} & Dirichlet & $  \textcolor{black}{\underline{64.4}}_{\textcolor{red}{\raisebox{2pt}{\normalsize-\textbf{0.8}}}}$ & $\text{60.9}_{\textcolor{red}{\raisebox{2pt}{\normalsize-5.7}}}$ & $\text{75.4}_{\textcolor{darkerGreen}{\raisebox{2pt}{\normalsize+12.9}}}$ & $\text{26.7}_{\textcolor{darkerGreen}{\raisebox{2pt}{\normalsize+2.0}}}$ & $\text{38.8}_{\textcolor{red}{\raisebox{2pt}{\normalsize-9.5}}}$ & $\text{74.1}_{\textcolor{darkerGreen}{\raisebox{2pt}{\normalsize+8.5}}}$ & $\text{76.2}_{\textcolor{red}{\raisebox{2pt}{\normalsize-9.7}}}$ & $\text{91.0}_{\textcolor{darkerGreen}{\raisebox{2pt}{\normalsize+1.9}}}$ & $\text{71.6}_{\textcolor{darkerGreen}{\raisebox{2pt}{\normalsize+0.9}}}$ & $\text{92.4}_{\textcolor{red}{\raisebox{2pt}{\normalsize-0.8}}}$ & $\text{36.2}_{\textcolor{red}{\raisebox{2pt}{\normalsize-7.3}}}$ & $\text{65.4}_{\textcolor{red}{\raisebox{2pt}{\normalsize-2.1}}}$  \\
 & ZLaP & $  \textcolor{black}{41.5}_{\textcolor{red}{\raisebox{2pt}{\normalsize-\textbf{23.7}}}}$ & $\text{16.6}_{\textcolor{red}{\raisebox{2pt}{\normalsize-50.0}}}$ & $\text{20.1}_{\textcolor{red}{\raisebox{2pt}{\normalsize-42.4}}}$ & $\text{16.4}_{\textcolor{red}{\raisebox{2pt}{\normalsize-8.3}}}$ & $\text{49.0}_{\textcolor{darkerGreen}{\raisebox{2pt}{\normalsize+0.7}}}$ & $\text{32.2}_{\textcolor{red}{\raisebox{2pt}{\normalsize-33.4}}}$ & $\text{55.5}_{\textcolor{red}{\raisebox{2pt}{\normalsize-30.4}}}$ & $\text{76.4}_{\textcolor{red}{\raisebox{2pt}{\normalsize-12.7}}}$ & $\text{40.6}_{\textcolor{red}{\raisebox{2pt}{\normalsize-30.1}}}$ & $\text{67.7}_{\textcolor{red}{\raisebox{2pt}{\normalsize-25.5}}}$ & $\text{34.2}_{\textcolor{red}{\raisebox{2pt}{\normalsize-9.3}}}$ & $\text{48.1}_{\textcolor{red}{\raisebox{2pt}{\normalsize-19.4}}}$  \\
 & TransCLIP & $  \textcolor{black}{56.5}_{\textcolor{red}{\raisebox{2pt}{\normalsize-\textbf{8.7}}}}$ & $\text{39.9}_{\textcolor{red}{\raisebox{2pt}{\normalsize-26.7}}}$ & $\text{42.7}_{\textcolor{red}{\raisebox{2pt}{\normalsize-19.8}}}$ & $\text{22.0}_{\textcolor{red}{\raisebox{2pt}{\normalsize-2.7}}}$ & $\text{63.1}_{\textcolor{darkerGreen}{\raisebox{2pt}{\normalsize+14.8}}}$ & $\text{49.9}_{\textcolor{red}{\raisebox{2pt}{\normalsize-15.7}}}$ & $\text{80.6}_{\textcolor{red}{\raisebox{2pt}{\normalsize-5.3}}}$ & $\text{87.9}_{\textcolor{red}{\raisebox{2pt}{\normalsize-1.2}}}$ & $\text{58.7}_{\textcolor{red}{\raisebox{2pt}{\normalsize-12.0}}}$ & $\text{79.1}_{\textcolor{red}{\raisebox{2pt}{\normalsize-14.1}}}$ & $\text{42.9}_{\textcolor{red}{\raisebox{2pt}{\normalsize-0.6}}}$ & $\text{55.0}_{\textcolor{red}{\raisebox{2pt}{\normalsize-12.5}}}$  \\
 & \cellcolor{LightGray} Stat${\cal A}$ & \cellcolor{AverageDarkerColor} $  \textcolor{black}{\textbf{69.7}}_{\textcolor{darkerGreen}{\raisebox{2pt}{\normalsize+\textbf{4.4}}}}$ & \cellcolor{LightGray} $\text{70.8}_{\textcolor{darkerGreen}{\raisebox{2pt}{\normalsize+4.2}}}$ & \cellcolor{LightGray} $\text{64.5}_{\textcolor{darkerGreen}{\raisebox{2pt}{\normalsize+2.0}}}$ & \cellcolor{LightGray} $\text{28.4}_{\textcolor{darkerGreen}{\raisebox{2pt}{\normalsize+3.7}}}$ & \cellcolor{LightGray} $\text{60.4}_{\textcolor{darkerGreen}{\raisebox{2pt}{\normalsize+12.1}}}$ & \cellcolor{LightGray} $\text{74.0}_{\textcolor{darkerGreen}{\raisebox{2pt}{\normalsize+8.4}}}$ & \cellcolor{LightGray} $\text{87.5}_{\textcolor{darkerGreen}{\raisebox{2pt}{\normalsize+1.6}}}$ & \cellcolor{LightGray} $\text{93.1}_{\textcolor{darkerGreen}{\raisebox{2pt}{\normalsize+4.0}}}$ & \cellcolor{LightGray} $\text{77.5}_{\textcolor{darkerGreen}{\raisebox{2pt}{\normalsize+6.8}}}$ & \cellcolor{LightGray} $\text{92.8}_{\textcolor{red}{\raisebox{2pt}{\normalsize-0.4}}}$ & \cellcolor{LightGray} $\text{47.1}_{\textcolor{darkerGreen}{\raisebox{2pt}{\normalsize+3.6}}}$ & \cellcolor{LightGray} $\text{70.2}_{\textcolor{darkerGreen}{\raisebox{2pt}{\normalsize+2.7}}}$  \\
\midrule
\multirow{4}{*}{\makecell{\texttt{High} \\ (25--50)}} & Dirichlet & $  \textcolor{black}{45.3}_{\textcolor{red}{\raisebox{2pt}{\normalsize-\textbf{20.0}}}}$ & $\text{17.3}_{\textcolor{red}{\raisebox{2pt}{\normalsize-49.3}}}$ & $\text{37.3}_{\textcolor{red}{\raisebox{2pt}{\normalsize-25.2}}}$ & $\text{21.0}_{\textcolor{red}{\raisebox{2pt}{\normalsize-3.7}}}$ & $\text{37.9}_{\textcolor{red}{\raisebox{2pt}{\normalsize-10.4}}}$ & $\text{65.4}_{\textcolor{red}{\raisebox{2pt}{\normalsize-0.2}}}$ & $\text{46.3}_{\textcolor{red}{\raisebox{2pt}{\normalsize-39.6}}}$ & $\text{81.3}_{\textcolor{red}{\raisebox{2pt}{\normalsize-7.8}}}$ & $\text{46.3}_{\textcolor{red}{\raisebox{2pt}{\normalsize-24.4}}}$ & $\text{80.5}_{\textcolor{red}{\raisebox{2pt}{\normalsize-12.7}}}$ & $\text{21.1}_{\textcolor{red}{\raisebox{2pt}{\normalsize-22.4}}}$ & $\text{43.6}_{\textcolor{red}{\raisebox{2pt}{\normalsize-23.9}}}$  \\
 & ZLaP & $  \textcolor{black}{52.2}_{\textcolor{red}{\raisebox{2pt}{\normalsize-\textbf{13.0}}}}$ & $\text{23.8}_{\textcolor{red}{\raisebox{2pt}{\normalsize-42.8}}}$ & $\text{32.2}_{\textcolor{red}{\raisebox{2pt}{\normalsize-30.3}}}$ & $\text{22.2}_{\textcolor{red}{\raisebox{2pt}{\normalsize-2.5}}}$ & $\text{49.3}_{\textcolor{darkerGreen}{\raisebox{2pt}{\normalsize+1.0}}}$ & $\text{45.4}_{\textcolor{red}{\raisebox{2pt}{\normalsize-20.2}}}$ & $\text{74.9}_{\textcolor{red}{\raisebox{2pt}{\normalsize-11.0}}}$ & $\text{86.5}_{\textcolor{red}{\raisebox{2pt}{\normalsize-2.6}}}$ & $\text{56.2}_{\textcolor{red}{\raisebox{2pt}{\normalsize-14.5}}}$ & $\text{79.7}_{\textcolor{red}{\raisebox{2pt}{\normalsize-13.5}}}$ & $\text{43.6}_{\textcolor{darkerGreen}{\raisebox{2pt}{\normalsize+0.1}}}$ & $\text{60.8}_{\textcolor{red}{\raisebox{2pt}{\normalsize-6.7}}}$  \\
 & TransCLIP & $  \textcolor{black}{\underline{62.0}}_{\textcolor{red}{\raisebox{2pt}{\normalsize-\textbf{3.3}}}}$ & $\text{43.9}_{\textcolor{red}{\raisebox{2pt}{\normalsize-22.7}}}$ & $\text{49.6}_{\textcolor{red}{\raisebox{2pt}{\normalsize-12.9}}}$ & $\text{24.8}_{\textcolor{darkerGreen}{\raisebox{2pt}{\normalsize+0.1}}}$ & $\text{64.0}_{\textcolor{darkerGreen}{\raisebox{2pt}{\normalsize+15.7}}}$ & $\text{57.3}_{\textcolor{red}{\raisebox{2pt}{\normalsize-8.3}}}$ & $\text{83.0}_{\textcolor{red}{\raisebox{2pt}{\normalsize-2.9}}}$ & $\text{91.4}_{\textcolor{darkerGreen}{\raisebox{2pt}{\normalsize+2.3}}}$ & $\text{69.1}_{\textcolor{red}{\raisebox{2pt}{\normalsize-1.6}}}$ & $\text{85.5}_{\textcolor{red}{\raisebox{2pt}{\normalsize-7.7}}}$ & $\text{47.5}_{\textcolor{darkerGreen}{\raisebox{2pt}{\normalsize+4.0}}}$ & $\text{65.4}_{\textcolor{red}{\raisebox{2pt}{\normalsize-2.1}}}$  \\
 & \cellcolor{LightGray} Stat${\cal A}$ & \cellcolor{AverageDarkerColor} $  \textcolor{black}{\textbf{69.8}}_{\textcolor{darkerGreen}{\raisebox{2pt}{\normalsize+\textbf{4.5}}}}$ & \cellcolor{LightGray} $\text{71.9}_{\textcolor{darkerGreen}{\raisebox{2pt}{\normalsize+5.3}}}$ & \cellcolor{LightGray} $\text{66.4}_{\textcolor{darkerGreen}{\raisebox{2pt}{\normalsize+3.9}}}$ & \cellcolor{LightGray} $\text{25.9}_{\textcolor{darkerGreen}{\raisebox{2pt}{\normalsize+1.2}}}$ & \cellcolor{LightGray} $\text{60.7}_{\textcolor{darkerGreen}{\raisebox{2pt}{\normalsize+12.4}}}$ & \cellcolor{LightGray} $\text{73.6}_{\textcolor{darkerGreen}{\raisebox{2pt}{\normalsize+8.0}}}$ & \cellcolor{LightGray} $\text{88.0}_{\textcolor{darkerGreen}{\raisebox{2pt}{\normalsize+2.1}}}$ & \cellcolor{LightGray} $\text{91.4}_{\textcolor{darkerGreen}{\raisebox{2pt}{\normalsize+2.3}}}$ & \cellcolor{LightGray} $\text{76.7}_{\textcolor{darkerGreen}{\raisebox{2pt}{\normalsize+6.0}}}$ & \cellcolor{LightGray} $\text{93.2}_{\textcolor{gray}{\raisebox{2pt}{\normalsize0.0}}}$ & \cellcolor{LightGray} $\text{47.9}_{\textcolor{darkerGreen}{\raisebox{2pt}{\normalsize+4.4}}}$ & \cellcolor{LightGray} $\text{71.5}_{\textcolor{darkerGreen}{\raisebox{2pt}{\normalsize+4.0}}}$  \\
\midrule
\multirow{4}{*}{\makecell{\texttt{Very High} \\ (50--100)}} & Dirichlet & $  \textcolor{black}{33.6}_{\textcolor{red}{\raisebox{2pt}{\normalsize-\textbf{31.6}}}}$ & $\text{10.8}_{\textcolor{red}{\raisebox{2pt}{\normalsize-55.8}}}$ & $\text{15.7}_{\textcolor{red}{\raisebox{2pt}{\normalsize-46.8}}}$ & $\text{17.5}_{\textcolor{red}{\raisebox{2pt}{\normalsize-7.2}}}$ & $\text{37.8}_{\textcolor{red}{\raisebox{2pt}{\normalsize-10.5}}}$ & $\text{51.2}_{\textcolor{red}{\raisebox{2pt}{\normalsize-14.4}}}$ & $\text{29.1}_{\textcolor{red}{\raisebox{2pt}{\normalsize-56.8}}}$ & $\text{79.3}_{\textcolor{red}{\raisebox{2pt}{\normalsize-9.8}}}$ & $\text{24.3}_{\textcolor{red}{\raisebox{2pt}{\normalsize-46.4}}}$ & $\text{59.1}_{\textcolor{red}{\raisebox{2pt}{\normalsize-34.1}}}$ & $\text{19.0}_{\textcolor{red}{\raisebox{2pt}{\normalsize-24.5}}}$ & $\text{26.1}_{\textcolor{red}{\raisebox{2pt}{\normalsize-41.4}}}$  \\
 & ZLaP & $  \textcolor{black}{58.4}_{\textcolor{red}{\raisebox{2pt}{\normalsize-\textbf{6.8}}}}$ & $\text{32.7}_{\textcolor{red}{\raisebox{2pt}{\normalsize-33.9}}}$ & $\text{44.0}_{\textcolor{red}{\raisebox{2pt}{\normalsize-18.5}}}$ & $\text{25.4}_{\textcolor{darkerGreen}{\raisebox{2pt}{\normalsize+0.7}}}$ & $\text{49.3}_{\textcolor{darkerGreen}{\raisebox{2pt}{\normalsize+1.0}}}$ & $\text{55.2}_{\textcolor{red}{\raisebox{2pt}{\normalsize-10.4}}}$ & $\text{83.3}_{\textcolor{red}{\raisebox{2pt}{\normalsize-2.6}}}$ & $\text{87.3}_{\textcolor{red}{\raisebox{2pt}{\normalsize-1.8}}}$ & $\text{64.8}_{\textcolor{red}{\raisebox{2pt}{\normalsize-5.9}}}$ & $\text{87.9}_{\textcolor{red}{\raisebox{2pt}{\normalsize-5.3}}}$ & $\text{45.2}_{\textcolor{darkerGreen}{\raisebox{2pt}{\normalsize+1.7}}}$ & $\text{67.8}_{\textcolor{darkerGreen}{\raisebox{2pt}{\normalsize+0.3}}}$  \\
 & TransCLIP & $  \textcolor{black}{\underline{64.4}}_{\textcolor{red}{\raisebox{2pt}{\normalsize-\textbf{0.8}}}}$ & $\text{44.5}_{\textcolor{red}{\raisebox{2pt}{\normalsize-22.1}}}$ & $\text{53.0}_{\textcolor{red}{\raisebox{2pt}{\normalsize-9.5}}}$ & $\text{25.6}_{\textcolor{darkerGreen}{\raisebox{2pt}{\normalsize+0.9}}}$ & $\text{64.1}_{\textcolor{darkerGreen}{\raisebox{2pt}{\normalsize+15.8}}}$ & $\text{60.9}_{\textcolor{red}{\raisebox{2pt}{\normalsize-4.7}}}$ & $\text{85.2}_{\textcolor{red}{\raisebox{2pt}{\normalsize-0.7}}}$ & $\text{91.9}_{\textcolor{darkerGreen}{\raisebox{2pt}{\normalsize+2.8}}}$ & $\text{74.3}_{\textcolor{darkerGreen}{\raisebox{2pt}{\normalsize+3.6}}}$ & $\text{90.5}_{\textcolor{red}{\raisebox{2pt}{\normalsize-2.7}}}$ & $\text{48.1}_{\textcolor{darkerGreen}{\raisebox{2pt}{\normalsize+4.6}}}$ & $\text{70.7}_{\textcolor{darkerGreen}{\raisebox{2pt}{\normalsize+3.2}}}$  \\
 & \cellcolor{LightGray} Stat${\cal A}$ & \cellcolor{AverageDarkerColor} $  \textcolor{black}{\textbf{69.0}}_{\textcolor{darkerGreen}{\raisebox{2pt}{\normalsize+\textbf{3.7}}}}$ & \cellcolor{LightGray} $\text{71.8}_{\textcolor{darkerGreen}{\raisebox{2pt}{\normalsize+5.2}}}$ & \cellcolor{LightGray} $\text{67.1}_{\textcolor{darkerGreen}{\raisebox{2pt}{\normalsize+4.6}}}$ & \cellcolor{LightGray} $\text{23.9}_{\textcolor{red}{\raisebox{2pt}{\normalsize-0.8}}}$ & \cellcolor{LightGray} $\text{60.7}_{\textcolor{darkerGreen}{\raisebox{2pt}{\normalsize+12.4}}}$ & \cellcolor{LightGray} $\text{70.2}_{\textcolor{darkerGreen}{\raisebox{2pt}{\normalsize+4.6}}}$ & \cellcolor{LightGray} $\text{87.1}_{\textcolor{darkerGreen}{\raisebox{2pt}{\normalsize+1.2}}}$ & \cellcolor{LightGray} $\text{91.1}_{\textcolor{darkerGreen}{\raisebox{2pt}{\normalsize+2.0}}}$ & \cellcolor{LightGray} $\text{74.3}_{\textcolor{darkerGreen}{\raisebox{2pt}{\normalsize+3.6}}}$ & \cellcolor{LightGray} $\text{93.7}_{\textcolor{darkerGreen}{\raisebox{2pt}{\normalsize+0.5}}}$ & \cellcolor{LightGray} $\text{48.0}_{\textcolor{darkerGreen}{\raisebox{2pt}{\normalsize+4.5}}}$ & \cellcolor{LightGray} $\text{70.7}_{\textcolor{darkerGreen}{\raisebox{2pt}{\normalsize+3.2}}}$  \\
\midrule
\end{tabular}}
\end{subtable}
\\
\vspace{0.5cm}
\begin{subtable}{\linewidth}
\centering
\caption{Scenario with full dataset and all classes.}
\label{tab:dataset_results}
\resizebox{\textwidth}{!}{%
\setlength\dashlinedash{0.2pt}
\setlength\dashlinegap{1.5pt}
\setlength\arrayrulewidth{0.3pt}
\renewcommand{\arraystretch}{1.}
    \begin{tabular}{p{1.7cm}l|nccccccccccc}
      \toprule
$K_{\text{eff}}$ & Method & \textsc{\fontsize{11}{10}\selectfont\textcolor{black}{Average}}& \rotatebox[origin=c]{45}{ImageNet} & \rotatebox[origin=c]{45}{SUN397} & \rotatebox[origin=c]{45}{Aircraft} & \rotatebox[origin=c]{45}{EuroSAT} & \rotatebox[origin=c]{45}{StanfordCars} & \rotatebox[origin=c]{45}{Food101} & \rotatebox[origin=c]{45}{Pets} & \rotatebox[origin=c]{45}{Flower102} & \rotatebox[origin=c]{45}{Caltech101} & \rotatebox[origin=c]{45}{DTD} & \rotatebox[origin=c]{45}{UCF101}  \\
\midrule
 & CLIP & $\textcolor{black}{65.2}$ & $\text{66.6}$ & $\text{62.5}$ & $\text{24.7}$ & $\text{48.3}$ & $\text{65.6}$ & $\text{85.9}$ & $\text{89.1}$ & $\text{70.7}$ & $\text{93.2}$ & $\text{43.5}$ & $\text{67.5}$  \\
 & MTA &  $\textcolor{black}{66.6}_{\textcolor{darkerGreen}{\raisebox{2pt}{\normalsize+\textbf{1.3}}}}$ & $\text{69.3}_{\textcolor{darkerGreen}{\raisebox{2pt}{\normalsize+2.7}}}$ & $\text{64.8}_{\textcolor{darkerGreen}{\raisebox{2pt}{\normalsize+2.3}}}$ & $\text{27.4}_{\textcolor{darkerGreen}{\raisebox{2pt}{\normalsize+2.7}}}$ & $\text{46.9}_{\textcolor{red}{\raisebox{2pt}{\normalsize-1.4}}}$ & $\text{68.0}_{\textcolor{darkerGreen}{\raisebox{2pt}{\normalsize+2.4}}}$ & $\text{87.2}_{\textcolor{darkerGreen}{\raisebox{2pt}{\normalsize+1.3}}}$ & $\text{89.4}_{\textcolor{darkerGreen}{\raisebox{2pt}{\normalsize+0.3}}}$ & $\text{71.7}_{\textcolor{darkerGreen}{\raisebox{2pt}{\normalsize+1.0}}}$ & $\text{94.0}_{\textcolor{darkerGreen}{\raisebox{2pt}{\normalsize+0.8}}}$ & $\text{44.4}_{\textcolor{darkerGreen}{\raisebox{2pt}{\normalsize+0.9}}}$ & $\text{69.0}_{\textcolor{darkerGreen}{\raisebox{2pt}{\normalsize+1.5}}}$  \\
\midrule
 \multirow{4}{*}{\makecell{\texttt{All}}} & Dirichlet &  $\textcolor{black}{29.5}_{\textcolor{red}{\raisebox{2pt}{\normalsize-\textbf{35.7}}}}$ & $\text{66.6}_{\textcolor{gray}{\raisebox{2pt}{\normalsize0.0}}}$ & $\text{0.7}_{\textcolor{red}{\raisebox{2pt}{\normalsize-61.8}}}$ & $\text{15.5}_{\textcolor{red}{\raisebox{2pt}{\normalsize-9.2}}}$ & $\text{39.9}_{\textcolor{red}{\raisebox{2pt}{\normalsize-8.4}}}$ & $\text{21.2}_{\textcolor{red}{\raisebox{2pt}{\normalsize-44.4}}}$ & $\text{8.5}_{\textcolor{red}{\raisebox{2pt}{\normalsize-77.4}}}$ & $\text{76.3}_{\textcolor{red}{\raisebox{2pt}{\normalsize-12.8}}}$ & $\text{14.3}_{\textcolor{red}{\raisebox{2pt}{\normalsize-56.4}}}$ & $\text{44.0}_{\textcolor{red}{\raisebox{2pt}{\normalsize-49.2}}}$ & $\text{19.4}_{\textcolor{red}{\raisebox{2pt}{\normalsize-24.1}}}$ & $\text{18.4}_{\textcolor{red}{\raisebox{2pt}{\normalsize-49.1}}}$  \\
 & ZLaP &  $\textcolor{black}{66.4}_{\textcolor{darkerGreen}{\raisebox{2pt}{\normalsize+\textbf{1.1}}}}$ & $\text{68.8}_{\textcolor{darkerGreen}{\raisebox{2pt}{\normalsize+2.2}}}$ & $\text{67.2}_{\textcolor{darkerGreen}{\raisebox{2pt}{\normalsize+4.7}}}$ & $\text{26.9}_{\textcolor{darkerGreen}{\raisebox{2pt}{\normalsize+2.2}}}$ & $\text{49.1}_{\textcolor{darkerGreen}{\raisebox{2pt}{\normalsize+0.8}}}$ & $\text{67.1}_{\textcolor{darkerGreen}{\raisebox{2pt}{\normalsize+1.5}}}$ & $\text{86.9}_{\textcolor{darkerGreen}{\raisebox{2pt}{\normalsize+1.0}}}$ & $\text{87.8}_{\textcolor{red}{\raisebox{2pt}{\normalsize-1.3}}}$ & $\text{68.7}_{\textcolor{red}{\raisebox{2pt}{\normalsize-2.0}}}$ & $\text{90.8}_{\textcolor{red}{\raisebox{2pt}{\normalsize-2.4}}}$ & $\text{45.5}_{\textcolor{darkerGreen}{\raisebox{2pt}{\normalsize+2.0}}}$ & $\text{71.4}_{\textcolor{darkerGreen}{\raisebox{2pt}{\normalsize+3.9}}}$  \\
 & TransCLIP &  $\textcolor{black}{\textbf{70.3}}_{\textcolor{darkerGreen}{\raisebox{2pt}{\normalsize+\textbf{5.1}}}}$ & $\text{70.4}_{\textcolor{darkerGreen}{\raisebox{2pt}{\normalsize+3.8}}}$ & $\text{68.9}_{\textcolor{darkerGreen}{\raisebox{2pt}{\normalsize+6.4}}}$ & $\text{26.9}_{\textcolor{darkerGreen}{\raisebox{2pt}{\normalsize+2.2}}}$ & $\text{66.1}_{\textcolor{darkerGreen}{\raisebox{2pt}{\normalsize+17.8}}}$ & $\text{69.5}_{\textcolor{darkerGreen}{\raisebox{2pt}{\normalsize+3.9}}}$ & $\text{87.1}_{\textcolor{darkerGreen}{\raisebox{2pt}{\normalsize+1.2}}}$ & $\text{92.5}_{\textcolor{darkerGreen}{\raisebox{2pt}{\normalsize+3.4}}}$ & $\text{76.5}_{\textcolor{darkerGreen}{\raisebox{2pt}{\normalsize+5.8}}}$ & $\text{92.7}_{\textcolor{red}{\raisebox{2pt}{\normalsize-0.5}}}$ & $\text{48.6}_{\textcolor{darkerGreen}{\raisebox{2pt}{\normalsize+5.1}}}$ & $\text{74.1}_{\textcolor{darkerGreen}{\raisebox{2pt}{\normalsize+6.6}}}$  \\
 & \cellcolor{LightGray} Stat${\cal A}$ & \cellcolor{AverageDarkerColor} $\textcolor{black}{\underline{69.9}}_{\textcolor{darkerGreen}{\raisebox{2pt}{\normalsize+\textbf{4.7}}}}$ & \cellcolor{LightGray} $\text{69.9}_{\textcolor{darkerGreen}{\raisebox{2pt}{\normalsize+3.3}}}$ & \cellcolor{LightGray} $\text{68.7}_{\textcolor{darkerGreen}{\raisebox{2pt}{\normalsize+6.2}}}$ & \cellcolor{LightGray} $\text{24.7}_{\textcolor{gray}{\raisebox{2pt}{\normalsize0.0}}}$ & \cellcolor{LightGray} $\text{67.3}_{\textcolor{darkerGreen}{\raisebox{2pt}{\normalsize+19.0}}}$ & \cellcolor{LightGray} $\text{68.0}_{\textcolor{darkerGreen}{\raisebox{2pt}{\normalsize+2.4}}}$ & \cellcolor{LightGray} $\text{87.1}_{\textcolor{darkerGreen}{\raisebox{2pt}{\normalsize+1.2}}}$ & \cellcolor{LightGray} $\text{92.4}_{\textcolor{darkerGreen}{\raisebox{2pt}{\normalsize+3.3}}}$ & \cellcolor{LightGray} $\text{75.2}_{\textcolor{darkerGreen}{\raisebox{2pt}{\normalsize+4.5}}}$ & \cellcolor{LightGray} $\text{94.2}_{\textcolor{darkerGreen}{\raisebox{2pt}{\normalsize+1.0}}}$ & \cellcolor{LightGray} $\text{48.4}_{\textcolor{darkerGreen}{\raisebox{2pt}{\normalsize+4.9}}}$ & \cellcolor{LightGray} $\text{73.5}_{\textcolor{darkerGreen}{\raisebox{2pt}{\normalsize+6.0}}}$  \\
\midrule
\end{tabular}}
\end{subtable}
\\
\end{table*}

\definecolor{AverageColor}{RGB}{255, 237, 206}
\definecolor{AverageDarkerColor}{RGB}{236, 227, 211}

\newcolumntype{n}{>{\columncolor{AverageColor}} p{1.7cm}}

\aboverulesep = 0.2mm 
\belowrulesep = 0.2mm 

\begin{table*}
\centering
\caption{Comparison of methods for online TTA. The best average performance for each configuration is highlighted in \textbf{bold}, while the second-best is indicated with \underline{underline}. Each reported performance is averaged over 100 tasks.}
\label{tab:online_tta}
\begin{subtable}{\linewidth}
\centering
\caption{Four scenarios with a batch size of 128: \texttt{Low}, \texttt{Medium}, and \texttt{High} correlation. For \texttt{Separate}, classes are seen sequentially.}
\resizebox{\textwidth}{!}{%
\setlength\dashlinedash{0.2pt}
\setlength\dashlinegap{1.5pt}
\setlength\arrayrulewidth{0.3pt}
\renewcommand{\arraystretch}{1.}
    \begin{tabular}{ll|ncccccccccccc}
      \toprule
Scenario & Method & \textsc{\fontsize{11}{10}\selectfont\textcolor{black}{Average}} & \rotatebox[origin=c]{45}{ImageNet} & \rotatebox[origin=c]{45}{SUN397} & \rotatebox[origin=c]{45}{Aircraft} & \rotatebox[origin=c]{45}{EuroSAT} & \rotatebox[origin=c]{45}{StanfordCars} & \rotatebox[origin=c]{45}{Food101} & \rotatebox[origin=c]{45}{Pets} & \rotatebox[origin=c]{45}{Flower102} & \rotatebox[origin=c]{45}{Caltech101} & \rotatebox[origin=c]{45}{DTD} & \rotatebox[origin=c]{45}{UCF101} \\
\midrule
  & CLIP& $\text{65.2}$ & $\text{66.6}$ & $\text{62.5}$ & $\text{24.7}$ & $\text{48.3}$ & $\text{65.6}$ & $\text{85.9}$ & $\text{89.1}$ & $\text{70.7}$ & $\text{93.2}$ & $\text{43.5}$ & $\text{67.5}$ \\
 & MTA & $\text{66.6}_{\textcolor{darkerGreen}{\raisebox{2pt}{\normalsize+\textbf{1.3}}}}$ & $\text{69.3}_{\textcolor{darkerGreen}{\raisebox{2pt}{\normalsize+\textbf{2.7}}}}$ & $\text{64.8}_{\textcolor{darkerGreen}{\raisebox{2pt}{\normalsize+\textbf{2.3}}}}$ & $\text{27.4}_{\textcolor{darkerGreen}{\raisebox{2pt}{\normalsize+\textbf{2.7}}}}$ & $\text{46.9}_{\textcolor{red}{\raisebox{2pt}{\normalsize-\textbf{1.4}}}}$ & $\text{68.0}_{\textcolor{darkerGreen}{\raisebox{2pt}{\normalsize+\textbf{2.4}}}}$ & $\text{87.2}_{\textcolor{darkerGreen}{\raisebox{2pt}{\normalsize+\textbf{1.3}}}}$ & $\text{89.4}_{\textcolor{darkerGreen}{\raisebox{2pt}{\normalsize+\textbf{0.3}}}}$ & $\text{71.7}_{\textcolor{darkerGreen}{\raisebox{2pt}{\normalsize+\textbf{1.0}}}}$ & $\text{94.0}_{\textcolor{darkerGreen}{\raisebox{2pt}{\normalsize+\textbf{0.8}}}}$ & $\text{44.4}_{\textcolor{darkerGreen}{\raisebox{2pt}{\normalsize+\textbf{0.9}}}}$ & $\text{69.0}_{\textcolor{darkerGreen}{\raisebox{2pt}{\normalsize+\textbf{1.5}}}}$ \\
\midrule
\multirow{5}{*}{\makecell{\texttt{Low} \\ ($\gamma = 0.1$)}} & TENT & $\text{65.8}_{\textcolor{darkerGreen}{\raisebox{2pt}{\normalsize+\textbf{0.6}}}}$ & $\text{66.6}_{\textcolor{gray}{\raisebox{2pt}{\normalsize\textbf{0.0}}}}$ & $\text{64.5}_{\textcolor{darkerGreen}{\raisebox{2pt}{\normalsize+\textbf{2.0}}}}$ & $\text{24.6}_{\textcolor{red}{\raisebox{2pt}{\normalsize-\textbf{0.1}}}}$ & $\text{51.8}_{\textcolor{darkerGreen}{\raisebox{2pt}{\normalsize+\textbf{3.5}}}}$ & $\text{65.7}_{\textcolor{darkerGreen}{\raisebox{2pt}{\normalsize+\textbf{0.1}}}}$ & $\text{85.9}_{\textcolor{gray}{\raisebox{2pt}{\normalsize\textbf{0.0}}}}$ & $\text{89.3}_{\textcolor{darkerGreen}{\raisebox{2pt}{\normalsize+\textbf{0.2}}}}$ & $\text{70.6}_{\textcolor{red}{\raisebox{2pt}{\normalsize-\textbf{0.1}}}}$ & $\text{93.4}_{\textcolor{darkerGreen}{\raisebox{2pt}{\normalsize+\textbf{0.2}}}}$ & $\text{44.0}_{\textcolor{darkerGreen}{\raisebox{2pt}{\normalsize+\textbf{0.5}}}}$ & $\text{67.8}_{\textcolor{darkerGreen}{\raisebox{2pt}{\normalsize+\textbf{0.3}}}}$ \\
 & TDA & $\textbf{67.7}_{\textcolor{darkerGreen}{\raisebox{2pt}{\normalsize+\textbf{2.5}}}}$ & $\text{68.3}_{\textcolor{darkerGreen}{\raisebox{2pt}{\normalsize+\textbf{1.7}}}}$ & $\text{66.0}_{\textcolor{darkerGreen}{\raisebox{2pt}{\normalsize+\textbf{3.5}}}}$ & $\text{25.4}_{\textcolor{darkerGreen}{\raisebox{2pt}{\normalsize+\textbf{0.7}}}}$ & $\text{60.6}_{\textcolor{darkerGreen}{\raisebox{2pt}{\normalsize+\textbf{12.3}}}}$ & $\text{66.9}_{\textcolor{darkerGreen}{\raisebox{2pt}{\normalsize+\textbf{1.3}}}}$ & $\text{86.1}_{\textcolor{darkerGreen}{\raisebox{2pt}{\normalsize+\textbf{0.2}}}}$ & $\text{89.6}_{\textcolor{darkerGreen}{\raisebox{2pt}{\normalsize+\textbf{0.5}}}}$ & $\text{72.5}_{\textcolor{darkerGreen}{\raisebox{2pt}{\normalsize+\textbf{1.8}}}}$ & $\text{93.4}_{\textcolor{darkerGreen}{\raisebox{2pt}{\normalsize+\textbf{0.2}}}}$ & $\text{45.5}_{\textcolor{darkerGreen}{\raisebox{2pt}{\normalsize+\textbf{2.0}}}}$ & $\text{71.0}_{\textcolor{darkerGreen}{\raisebox{2pt}{\normalsize+\textbf{3.5}}}}$ \\
 & DMN & $\underline{\text{67.2}}_{\textcolor{darkerGreen}{\raisebox{2pt}{\normalsize+\textbf{2.0}}}}$ & $\text{68.0}_{\textcolor{darkerGreen}{\raisebox{2pt}{\normalsize+\textbf{1.4}}}}$ & $\text{64.8}_{\textcolor{darkerGreen}{\raisebox{2pt}{\normalsize+\textbf{2.3}}}}$ & $\text{24.9}_{\textcolor{darkerGreen}{\raisebox{2pt}{\normalsize+\textbf{0.2}}}}$ & $\text{59.8}_{\textcolor{darkerGreen}{\raisebox{2pt}{\normalsize+\textbf{11.5}}}}$ & $\text{67.0}_{\textcolor{darkerGreen}{\raisebox{2pt}{\normalsize+\textbf{1.4}}}}$ & $\text{84.2}_{\textcolor{red}{\raisebox{2pt}{\normalsize-\textbf{1.7}}}}$ & $\text{89.9}_{\textcolor{darkerGreen}{\raisebox{2pt}{\normalsize+\textbf{0.8}}}}$ & $\text{73.3}_{\textcolor{darkerGreen}{\raisebox{2pt}{\normalsize+\textbf{2.6}}}}$ & $\text{92.2}_{\textcolor{red}{\raisebox{2pt}{\normalsize-\textbf{1.0}}}}$ & $\text{44.8}_{\textcolor{darkerGreen}{\raisebox{2pt}{\normalsize+\textbf{1.3}}}}$ & $\text{70.3}_{\textcolor{darkerGreen}{\raisebox{2pt}{\normalsize+\textbf{2.8}}}}$ \\
 & \cellcolor{LightGray} Stat${\cal A}$ & \cellcolor{LightGray} $\text{67.0}_{\textcolor{darkerGreen}{\raisebox{2pt}{\normalsize+\textbf{1.7}}}}$ & \cellcolor{LightGray} $\text{66.2}_{\textcolor{red}{\raisebox{2pt}{\normalsize-\textbf{0.4}}}}$ & \cellcolor{LightGray} $\text{63.6}_{\textcolor{darkerGreen}{\raisebox{2pt}{\normalsize+\textbf{1.1}}}}$ & \cellcolor{LightGray} $\text{24.3}_{\textcolor{red}{\raisebox{2pt}{\normalsize-\textbf{0.4}}}}$ & \cellcolor{LightGray} $\text{52.3}_{\textcolor{darkerGreen}{\raisebox{2pt}{\normalsize+\textbf{4.0}}}}$ & \cellcolor{LightGray} $\text{67.4}_{\textcolor{darkerGreen}{\raisebox{2pt}{\normalsize+\textbf{1.8}}}}$ & \cellcolor{LightGray} $\text{88.0}_{\textcolor{darkerGreen}{\raisebox{2pt}{\normalsize+\textbf{2.1}}}}$ & \cellcolor{LightGray} $\text{92.5}_{\textcolor{darkerGreen}{\raisebox{2pt}{\normalsize+\textbf{3.4}}}}$ & \cellcolor{LightGray} $\text{72.7}_{\textcolor{darkerGreen}{\raisebox{2pt}{\normalsize+\textbf{2.0}}}}$ & \cellcolor{LightGray} $\text{94.2}_{\textcolor{darkerGreen}{\raisebox{2pt}{\normalsize+\textbf{1.0}}}}$ & \cellcolor{LightGray} $\text{46.8}_{\textcolor{darkerGreen}{\raisebox{2pt}{\normalsize+\textbf{3.3}}}}$ & \cellcolor{LightGray} $\text{68.8}_{\textcolor{darkerGreen}{\raisebox{2pt}{\normalsize+\textbf{1.3}}}}$ \\
\midrule
\multirow{5}{*}{\makecell{\texttt{Medium} \\ ($\gamma = 0.01$)}} & TENT & $\text{65.5}_{\textcolor{darkerGreen}{\raisebox{2pt}{\normalsize+\textbf{0.2}}}}$ & $\text{66.7}_{\textcolor{darkerGreen}{\raisebox{2pt}{\normalsize+\textbf{0.1}}}}$ & $\text{64.3}_{\textcolor{darkerGreen}{\raisebox{2pt}{\normalsize+\textbf{1.8}}}}$ & $\text{24.6}_{\textcolor{red}{\raisebox{2pt}{\normalsize-\textbf{0.1}}}}$ & $\text{47.9}_{\textcolor{red}{\raisebox{2pt}{\normalsize-\textbf{0.4}}}}$ & $\text{65.6}_{\textcolor{gray}{\raisebox{2pt}{\normalsize\textbf{0.0}}}}$ & $\text{85.9}_{\textcolor{gray}{\raisebox{2pt}{\normalsize\textbf{0.0}}}}$ & $\text{89.4}_{\textcolor{darkerGreen}{\raisebox{2pt}{\normalsize+\textbf{0.3}}}}$ & $\text{70.6}_{\textcolor{red}{\raisebox{2pt}{\normalsize-\textbf{0.1}}}}$ & $\text{93.3}_{\textcolor{darkerGreen}{\raisebox{2pt}{\normalsize+\textbf{0.1}}}}$ & $\text{44.0}_{\textcolor{darkerGreen}{\raisebox{2pt}{\normalsize+\textbf{0.5}}}}$ & $\text{67.8}_{\textcolor{darkerGreen}{\raisebox{2pt}{\normalsize+\textbf{0.3}}}}$ \\
 & TDA & $\underline{\text{67.1}}_{\textcolor{darkerGreen}{\raisebox{2pt}{\normalsize+\textbf{1.9}}}}$ & $\text{68.2}_{\textcolor{darkerGreen}{\raisebox{2pt}{\normalsize+\textbf{1.6}}}}$ & $\text{65.6}_{\textcolor{darkerGreen}{\raisebox{2pt}{\normalsize+\textbf{3.1}}}}$ & $\text{25.2}_{\textcolor{darkerGreen}{\raisebox{2pt}{\normalsize+\textbf{0.5}}}}$ & $\text{56.5}_{\textcolor{darkerGreen}{\raisebox{2pt}{\normalsize+\textbf{8.2}}}}$ & $\text{66.5}_{\textcolor{darkerGreen}{\raisebox{2pt}{\normalsize+\textbf{0.9}}}}$ & $\text{85.8}_{\textcolor{red}{\raisebox{2pt}{\normalsize-\textbf{0.1}}}}$ & $\text{89.3}_{\textcolor{darkerGreen}{\raisebox{2pt}{\normalsize+\textbf{0.2}}}}$ & $\text{72.6}_{\textcolor{darkerGreen}{\raisebox{2pt}{\normalsize+\textbf{1.9}}}}$ & $\text{93.5}_{\textcolor{darkerGreen}{\raisebox{2pt}{\normalsize+\textbf{0.3}}}}$ & $\text{45.2}_{\textcolor{darkerGreen}{\raisebox{2pt}{\normalsize+\textbf{1.7}}}}$ & $\text{70.1}_{\textcolor{darkerGreen}{\raisebox{2pt}{\normalsize+\textbf{2.6}}}}$ \\
 & DMN & $\text{66.5}_{\textcolor{darkerGreen}{\raisebox{2pt}{\normalsize+\textbf{1.2}}}}$ & $\text{68.0}_{\textcolor{darkerGreen}{\raisebox{2pt}{\normalsize+\textbf{1.4}}}}$ & $\text{64.8}_{\textcolor{darkerGreen}{\raisebox{2pt}{\normalsize+\textbf{2.3}}}}$ & $\text{24.9}_{\textcolor{darkerGreen}{\raisebox{2pt}{\normalsize+\textbf{0.2}}}}$ & $\text{56.2}_{\textcolor{darkerGreen}{\raisebox{2pt}{\normalsize+\textbf{7.9}}}}$ & $\text{66.8}_{\textcolor{darkerGreen}{\raisebox{2pt}{\normalsize+\textbf{1.2}}}}$ & $\text{81.9}_{\textcolor{red}{\raisebox{2pt}{\normalsize-\textbf{4.0}}}}$ & $\text{89.0}_{\textcolor{red}{\raisebox{2pt}{\normalsize-\textbf{0.1}}}}$ & $\text{73.0}_{\textcolor{darkerGreen}{\raisebox{2pt}{\normalsize+\textbf{2.3}}}}$ & $\text{92.1}_{\textcolor{red}{\raisebox{2pt}{\normalsize-\textbf{1.1}}}}$ & $\text{44.9}_{\textcolor{darkerGreen}{\raisebox{2pt}{\normalsize+\textbf{1.4}}}}$ & $\text{69.6}_{\textcolor{darkerGreen}{\raisebox{2pt}{\normalsize+\textbf{2.1}}}}$ \\
 & \cellcolor{LightGray} Stat${\cal A}$ & \cellcolor{LightGray} $\textbf{68.9}_{\textcolor{darkerGreen}{\raisebox{2pt}{\normalsize+\textbf{3.7}}}}$ & \cellcolor{LightGray} $\text{69.6}_{\textcolor{darkerGreen}{\raisebox{2pt}{\normalsize+\textbf{3.0}}}}$ & \cellcolor{LightGray} $\text{65.9}_{\textcolor{darkerGreen}{\raisebox{2pt}{\normalsize+\textbf{3.4}}}}$ & \cellcolor{LightGray} $\text{27.3}_{\textcolor{darkerGreen}{\raisebox{2pt}{\normalsize+\textbf{2.6}}}}$ & \cellcolor{LightGray} $\text{52.3}_{\textcolor{darkerGreen}{\raisebox{2pt}{\normalsize+\textbf{4.0}}}}$ & \cellcolor{LightGray} $\text{73.2}_{\textcolor{darkerGreen}{\raisebox{2pt}{\normalsize+\textbf{7.6}}}}$ & \cellcolor{LightGray} $\text{89.1}_{\textcolor{darkerGreen}{\raisebox{2pt}{\normalsize+\textbf{3.2}}}}$ & \cellcolor{LightGray} $\text{94.6}_{\textcolor{darkerGreen}{\raisebox{2pt}{\normalsize+\textbf{5.5}}}}$ & \cellcolor{LightGray} $\text{75.6}_{\textcolor{darkerGreen}{\raisebox{2pt}{\normalsize+\textbf{4.9}}}}$ & \cellcolor{LightGray} $\text{94.3}_{\textcolor{darkerGreen}{\raisebox{2pt}{\normalsize+\textbf{1.1}}}}$ & \cellcolor{LightGray} $\text{46.8}_{\textcolor{darkerGreen}{\raisebox{2pt}{\normalsize+\textbf{3.3}}}}$ & \cellcolor{LightGray} $\text{69.7}_{\textcolor{darkerGreen}{\raisebox{2pt}{\normalsize+\textbf{2.2}}}}$ \\
\midrule
\multirow{5}{*}{\makecell{\texttt{High} \\ ($\gamma = 0.001$)}} & TENT & $\text{65.3}_{\textcolor{darkerGreen}{\raisebox{2pt}{\normalsize+\textbf{0.1}}}}$ & $\text{66.8}_{\textcolor{darkerGreen}{\raisebox{2pt}{\normalsize+\textbf{0.2}}}}$ & $\text{64.3}_{\textcolor{darkerGreen}{\raisebox{2pt}{\normalsize+\textbf{1.8}}}}$ & $\text{24.8}_{\textcolor{darkerGreen}{\raisebox{2pt}{\normalsize+\textbf{0.1}}}}$ & $\text{45.6}_{\textcolor{red}{\raisebox{2pt}{\normalsize-\textbf{2.7}}}}$ & $\text{65.6}_{\textcolor{gray}{\raisebox{2pt}{\normalsize\textbf{0.0}}}}$ & $\text{86.1}_{\textcolor{darkerGreen}{\raisebox{2pt}{\normalsize+\textbf{0.2}}}}$ & $\text{89.4}_{\textcolor{darkerGreen}{\raisebox{2pt}{\normalsize+\textbf{0.3}}}}$ & $\text{70.5}_{\textcolor{red}{\raisebox{2pt}{\normalsize-\textbf{0.2}}}}$ & $\text{93.4}_{\textcolor{darkerGreen}{\raisebox{2pt}{\normalsize+\textbf{0.2}}}}$ & $\text{44.0}_{\textcolor{darkerGreen}{\raisebox{2pt}{\normalsize+\textbf{0.5}}}}$ & $\text{67.9}_{\textcolor{darkerGreen}{\raisebox{2pt}{\normalsize+\textbf{0.4}}}}$ \\
 & TDA & $\underline{\text{66.8}}_{\textcolor{darkerGreen}{\raisebox{2pt}{\normalsize+\textbf{1.6}}}}$ & $\text{67.9}_{\textcolor{darkerGreen}{\raisebox{2pt}{\normalsize+\textbf{1.3}}}}$ & $\text{65.1}_{\textcolor{darkerGreen}{\raisebox{2pt}{\normalsize+\textbf{2.6}}}}$ & $\text{25.1}_{\textcolor{darkerGreen}{\raisebox{2pt}{\normalsize+\textbf{0.4}}}}$ & $\text{55.3}_{\textcolor{darkerGreen}{\raisebox{2pt}{\normalsize+\textbf{7.0}}}}$ & $\text{66.3}_{\textcolor{darkerGreen}{\raisebox{2pt}{\normalsize+\textbf{0.7}}}}$ & $\text{85.5}_{\textcolor{red}{\raisebox{2pt}{\normalsize-\textbf{0.4}}}}$ & $\text{89.0}_{\textcolor{red}{\raisebox{2pt}{\normalsize-\textbf{0.1}}}}$ & $\text{72.5}_{\textcolor{darkerGreen}{\raisebox{2pt}{\normalsize+\textbf{1.8}}}}$ & $\text{93.6}_{\textcolor{darkerGreen}{\raisebox{2pt}{\normalsize+\textbf{0.4}}}}$ & $\text{45.1}_{\textcolor{darkerGreen}{\raisebox{2pt}{\normalsize+\textbf{1.6}}}}$ & $\text{69.7}_{\textcolor{darkerGreen}{\raisebox{2pt}{\normalsize+\textbf{2.2}}}}$ \\
 & DMN & $\text{66.3}_{\textcolor{darkerGreen}{\raisebox{2pt}{\normalsize+\textbf{1.0}}}}$ & $\text{67.9}_{\textcolor{darkerGreen}{\raisebox{2pt}{\normalsize+\textbf{1.3}}}}$ & $\text{64.8}_{\textcolor{darkerGreen}{\raisebox{2pt}{\normalsize+\textbf{2.3}}}}$ & $\text{24.9}_{\textcolor{darkerGreen}{\raisebox{2pt}{\normalsize+\textbf{0.2}}}}$ & $\text{56.3}_{\textcolor{darkerGreen}{\raisebox{2pt}{\normalsize+\textbf{8.0}}}}$ & $\text{66.8}_{\textcolor{darkerGreen}{\raisebox{2pt}{\normalsize+\textbf{1.2}}}}$ & $\text{79.9}_{\textcolor{red}{\raisebox{2pt}{\normalsize-\textbf{6.0}}}}$ & $\text{88.9}_{\textcolor{red}{\raisebox{2pt}{\normalsize-\textbf{0.2}}}}$ & $\text{72.9}_{\textcolor{darkerGreen}{\raisebox{2pt}{\normalsize+\textbf{2.2}}}}$ & $\text{92.1}_{\textcolor{red}{\raisebox{2pt}{\normalsize-\textbf{1.1}}}}$ & $\text{44.8}_{\textcolor{darkerGreen}{\raisebox{2pt}{\normalsize+\textbf{1.3}}}}$ & $\text{69.4}_{\textcolor{darkerGreen}{\raisebox{2pt}{\normalsize+\textbf{1.9}}}}$ \\
 & \cellcolor{LightGray} Stat${\cal A}$ & \cellcolor{LightGray} $\textbf{69.5}_{\textcolor{darkerGreen}{\raisebox{2pt}{\normalsize+\textbf{4.2}}}}$ & \cellcolor{LightGray} $\text{71.9}_{\textcolor{darkerGreen}{\raisebox{2pt}{\normalsize+\textbf{5.3}}}}$ & \cellcolor{LightGray} $\text{66.0}_{\textcolor{darkerGreen}{\raisebox{2pt}{\normalsize+\textbf{3.5}}}}$ & \cellcolor{LightGray} $\text{27.9}_{\textcolor{darkerGreen}{\raisebox{2pt}{\normalsize+\textbf{3.2}}}}$ & \cellcolor{LightGray} $\text{51.8}_{\textcolor{darkerGreen}{\raisebox{2pt}{\normalsize+\textbf{3.5}}}}$ & \cellcolor{LightGray} $\text{74.7}_{\textcolor{darkerGreen}{\raisebox{2pt}{\normalsize+\textbf{9.1}}}}$ & \cellcolor{LightGray} $\text{89.3}_{\textcolor{darkerGreen}{\raisebox{2pt}{\normalsize+\textbf{3.4}}}}$ & \cellcolor{LightGray} $\text{94.8}_{\textcolor{darkerGreen}{\raisebox{2pt}{\normalsize+\textbf{5.7}}}}$ & \cellcolor{LightGray} $\text{76.4}_{\textcolor{darkerGreen}{\raisebox{2pt}{\normalsize+\textbf{5.7}}}}$ & \cellcolor{LightGray} $\text{94.4}_{\textcolor{darkerGreen}{\raisebox{2pt}{\normalsize+\textbf{1.2}}}}$ & \cellcolor{LightGray} $\text{47.0}_{\textcolor{darkerGreen}{\raisebox{2pt}{\normalsize+\textbf{3.5}}}}$ & \cellcolor{LightGray} $\text{69.8}_{\textcolor{darkerGreen}{\raisebox{2pt}{\normalsize+\textbf{2.3}}}}$ \\
\midrule
\multirow{5}{*}{\makecell{\texttt{Separate}}} & TENT & $\text{64.5}_{\textcolor{red}{\raisebox{2pt}{\normalsize-\textbf{0.7}}}}$ & $\text{66.7}_{\textcolor{darkerGreen}{\raisebox{2pt}{\normalsize+\textbf{0.1}}}}$ & $\text{64.2}_{\textcolor{darkerGreen}{\raisebox{2pt}{\normalsize+\textbf{1.7}}}}$ & $\text{24.7}_{\textcolor{gray}{\raisebox{2pt}{\normalsize\textbf{0.0}}}}$ & $\text{37.0}_{\textcolor{red}{\raisebox{2pt}{\normalsize-\textbf{11.3}}}}$ & $\text{65.6}_{\textcolor{gray}{\raisebox{2pt}{\normalsize\textbf{0.0}}}}$ & $\text{86.1}_{\textcolor{darkerGreen}{\raisebox{2pt}{\normalsize+\textbf{0.2}}}}$ & $\text{89.3}_{\textcolor{darkerGreen}{\raisebox{2pt}{\normalsize+\textbf{0.2}}}}$ & $\text{70.8}_{\textcolor{darkerGreen}{\raisebox{2pt}{\normalsize+\textbf{0.1}}}}$ & $\text{93.4}_{\textcolor{darkerGreen}{\raisebox{2pt}{\normalsize+\textbf{0.2}}}}$ & $\text{43.9}_{\textcolor{darkerGreen}{\raisebox{2pt}{\normalsize+\textbf{0.4}}}}$ & $\text{67.9}_{\textcolor{darkerGreen}{\raisebox{2pt}{\normalsize+\textbf{0.4}}}}$ \\
 & TDA & $\underline{\text{66.6}}_{\textcolor{darkerGreen}{\raisebox{2pt}{\normalsize+\textbf{1.4}}}}$ & $\text{67.4}_{\textcolor{darkerGreen}{\raisebox{2pt}{\normalsize+\textbf{0.8}}}}$ & $\text{64.6}_{\textcolor{darkerGreen}{\raisebox{2pt}{\normalsize+\textbf{2.1}}}}$ & $\text{24.9}_{\textcolor{darkerGreen}{\raisebox{2pt}{\normalsize+\textbf{0.2}}}}$ & $\text{55.3}_{\textcolor{darkerGreen}{\raisebox{2pt}{\normalsize+\textbf{7.0}}}}$ & $\text{65.9}_{\textcolor{darkerGreen}{\raisebox{2pt}{\normalsize+\textbf{0.3}}}}$ & $\text{85.2}_{\textcolor{red}{\raisebox{2pt}{\normalsize-\textbf{0.7}}}}$ & $\text{88.9}_{\textcolor{red}{\raisebox{2pt}{\normalsize-\textbf{0.2}}}}$ & $\text{72.3}_{\textcolor{darkerGreen}{\raisebox{2pt}{\normalsize+\textbf{1.6}}}}$ & $\text{93.6}_{\textcolor{darkerGreen}{\raisebox{2pt}{\normalsize+\textbf{0.4}}}}$ & $\text{45.0}_{\textcolor{darkerGreen}{\raisebox{2pt}{\normalsize+\textbf{1.5}}}}$ & $\text{69.6}_{\textcolor{darkerGreen}{\raisebox{2pt}{\normalsize+\textbf{2.1}}}}$ \\
 & DMN & $\text{65.8}_{\textcolor{darkerGreen}{\raisebox{2pt}{\normalsize+\textbf{0.6}}}}$ & $\text{67.7}_{\textcolor{darkerGreen}{\raisebox{2pt}{\normalsize+\textbf{1.1}}}}$ & $\text{64.7}_{\textcolor{darkerGreen}{\raisebox{2pt}{\normalsize+\textbf{2.2}}}}$ & $\text{24.9}_{\textcolor{darkerGreen}{\raisebox{2pt}{\normalsize+\textbf{0.2}}}}$ & $\text{55.1}_{\textcolor{darkerGreen}{\raisebox{2pt}{\normalsize+\textbf{6.8}}}}$ & $\text{66.7}_{\textcolor{darkerGreen}{\raisebox{2pt}{\normalsize+\textbf{1.1}}}}$ & $\text{78.5}_{\textcolor{red}{\raisebox{2pt}{\normalsize-\textbf{7.4}}}}$ & $\text{88.0}_{\textcolor{red}{\raisebox{2pt}{\normalsize-\textbf{1.1}}}}$ & $\text{72.8}_{\textcolor{darkerGreen}{\raisebox{2pt}{\normalsize+\textbf{2.1}}}}$ & $\text{91.9}_{\textcolor{red}{\raisebox{2pt}{\normalsize-\textbf{1.3}}}}$ & $\text{44.8}_{\textcolor{darkerGreen}{\raisebox{2pt}{\normalsize+\textbf{1.3}}}}$ & $\text{69.0}_{\textcolor{darkerGreen}{\raisebox{2pt}{\normalsize+\textbf{1.5}}}}$ \\
 & \cellcolor{LightGray} Stat${\cal A}$ & \cellcolor{LightGray} $\textbf{69.1}_{\textcolor{darkerGreen}{\raisebox{2pt}{\normalsize+\textbf{3.8}}}}$ & \cellcolor{LightGray} $\text{71.7}_{\textcolor{darkerGreen}{\raisebox{2pt}{\normalsize+\textbf{5.1}}}}$ & \cellcolor{LightGray} $\text{64.9}_{\textcolor{darkerGreen}{\raisebox{2pt}{\normalsize+\textbf{2.4}}}}$ & \cellcolor{LightGray} $\text{28.9}_{\textcolor{darkerGreen}{\raisebox{2pt}{\normalsize+\textbf{4.2}}}}$ & \cellcolor{LightGray} $\text{48.2}_{\textcolor{red}{\raisebox{2pt}{\normalsize-\textbf{0.1}}}}$ & \cellcolor{LightGray} $\text{75.2}_{\textcolor{darkerGreen}{\raisebox{2pt}{\normalsize+\textbf{9.6}}}}$ & \cellcolor{LightGray} $\text{88.9}_{\textcolor{darkerGreen}{\raisebox{2pt}{\normalsize+\textbf{3.0}}}}$ & \cellcolor{LightGray} $\text{95.2}_{\textcolor{darkerGreen}{\raisebox{2pt}{\normalsize+\textbf{6.1}}}}$ & \cellcolor{LightGray} $\text{77.6}_{\textcolor{darkerGreen}{\raisebox{2pt}{\normalsize+\textbf{6.9}}}}$ & \cellcolor{LightGray} $\text{94.3}_{\textcolor{darkerGreen}{\raisebox{2pt}{\normalsize+\textbf{1.1}}}}$ & \cellcolor{LightGray} $\text{45.8}_{\textcolor{darkerGreen}{\raisebox{2pt}{\normalsize+\textbf{2.3}}}}$ & \cellcolor{LightGray} $\text{69.0}_{\textcolor{darkerGreen}{\raisebox{2pt}{\normalsize+\textbf{1.5}}}}$ \\
\midrule
\end{tabular}}
\end{subtable}
\\
\end{table*}

\subsection{Complete formulation}
\paragraph{Assignments' regularization.}
In addition to the proposed regularization of statistical parameters $(\boldsymbol{\mu}, \boldsymbol{\Sigma})$, we follow recent works, regularizing assignments ${\mathbf z}$. We adopt the ubiquitous Laplacian regularizer \cite{ziko2020laplacian, zanella2024boosting, scholkopf_learning_2007}, as well as the text-supervision term of \cite{zanella2024boosting} specifically designed for VLMs. The former encourages samples with close visual features to have the same predictions, whereas the latter penalizes deviation of assignment variable $\mathbf z_i$ from zero-shot predictions $\hat{\mathbf{y}}_{i} = (y_{i,k})_{1 \leq k \leq K}$. 
More precisely, we set regularizer ${\cal R} ({\mathbf z})$, which appears in Eqs. \eqref{eq:rem_objective} and \eqref{final-zero-shot-objective}, as follows: 
\begin{align}
\label{original-transclip-zero-shot-objective}
{\cal R} ({\mathbf z}) =   - \underbrace{\sum_{i,j} w_{ij} \mathbf{z}_{i}^\top \mathbf{z}_{j}}_{\text{Laplacian reg.}}   + \underbrace{\sum_{i \in \cal Q} \mbox{KL} (\mathbf{z}_{i} || \hat{\mathbf{y}}_{i})}_{\text{text supervision}}
\end{align}
where $w_{ij} = \mathbf{f}_i^\top \mathbf{f}_j$.
With parameters $(\boldsymbol{\mu}, \boldsymbol{\Sigma})$ fixed, and using the regularizer in \eqref{original-transclip-zero-shot-objective}, the overall objective in \eqref{final-zero-shot-objective} becomes the sum of concave and convex functions of $\mathbf z$. Therefore, we could deploy a Concave-Convex Procedure (CCCP) minimization procedure \cite{CCCP-2001}, which, at each update step, 
minimizes a tight convex upper bound on the objective. This guarantees that each update does not increase the objective. CCCP replaces the concave part by its linear first-order 
approximation at the current step, which is a tight upper bound, while keeping the convex part unchanged. In our case, the Laplacian term is concave while the remaining part of the objective is convex. Solving the necessary conditions for minimizing the convex upper bound gives the following decoupled updates of the assignment variables (detailed derivation provided in the Appendix): 
\begin{equation}
    \mathbf{z}_{i}^{(l+1)} = \frac{\hat{\mathbf{y}}_{i} \odot \exp (\log (\mathbf{p}_{i}) + \sum\limits_{j} w_{ij} \mathbf{z}_{j}^{(l)})}{(\hat{\mathbf{y}}_{i} \odot \exp (\log (\mathbf{p}_{i}) + \sum\limits_{j} w_{ij}\mathbf{z}_{j}^{(l)}))^\top\mathbbm{1}_K}
\label{eq:z_update}
\end{equation}



\section{Experiments}
\label{sec:experiments}
We use the same datasets as in previous work~\cite{coop}. These diverse datasets provide a comprehensive visual classification benchmark. Additional information on the statistics of each dataset is provided in Table \ref{tab:appendix_datasets}. Details on datasets, hyperparameters and comparative methods are provided in the Supplementary Material \ref{sec:details_exp}. We use the same handcrafted prompts for all methods, which are listed in Table \ref{tab:app_all_prompts} of the Appendix and the ViT-B/16 CLIP model \cite{radford2021learning}.

\subsection{Batch test-time adaptation}

\paragraph{Experimental details.} Each task corresponds to a batch, which is processed independently. To reduce variability in results, 1,000 tasks are generated for each scenario, and accuracy is averaged across tasks. We report results for batch sizes of 64 (with \texttt{Very Low}, \texttt{Low}, \texttt{Medium} number of effective classes) and 1,000 (with \texttt{Medium}, \texttt{High}, \texttt{Very High} number of effective classes), and the whole dataset (with \texttt{All} classes). See Section \ref{sec:realistic} for more details. 

\paragraph{Results.} Table \ref{tab:small_batch_results} presents results for small batches containing relatively few effective classes, Table \ref{tab:big_batch_results} shows results for larger batches, and 
Table \ref{tab:dataset_results} reports results for large-scale, whole-dataset processing. Stat${\cal A}$ demonstrates robustness across all scenarios, whereas other transductive methods exhibit strong performance only within specific, narrow application ranges. Dirichlet performs well in the \texttt{Low} setting but rapidly declines as conditions deviate, which aligns closely with the setting emphasized in their paper (3 to 10 effective classes per batch of size 75). TransCLIP excels in the \texttt{All} scenario but quickly underperforms when the number of effective classes decreases, even with relatively large batch sizes, consistent with their experimental focus on large-scale transduction with all classes present. Similarly, ZLaP shows small performance improvements when all classes are present, but struggles in more diverse settings. Stat${\cal A}$ emerges as a well-balanced solution, providing stable gains across a wide range of scenarios.
 
\subsection{Online test-time adaptation}
\paragraph{Experimental details.} 
Each task corresponds to a full pass through a generated stream. For a given stream, class correlation between batches is controlled with a Dirichlet distribution denoted by $\gamma$ (lower values indicate higher correlation), as described in Section \ref{sec:realistic}. Visualization for three datasets are shown in Figure \ref{fig:online_data_streams_correlation} of the Supplementary Material. To reduce variability in results, 100 tasks are generated for each configuration and accuracy is averaged. We report results using a batch size of 128.

\paragraph{Results.} Table \ref{tab:online_tta} presents the results for various correlated streams. Stat${\cal A}$ demonstrates robustness across all scenarios. In contrast, while TDA and DMN do not experience as sharp a performance drop as some transductive methods from the previous sections, they still struggle to deliver significant performance gains, demonstrating that advances are still to be made for online test-time adaptation of VLMs. Moreover, TENT does not provide consistent improvements which might be due to the absence of batch normalization layers in the CLIP encoder. In contrast, Stat${\cal A}$ benefits from more correlated streams and still emerges as a good compromise across a wide range of scenarios.
 
\subsection{Ablation studies}

\begin{figure*}[t]
\centering

\begin{subfigure}[t]{0.97\textwidth}
    \centering
    \includegraphics[width=.33\textwidth]{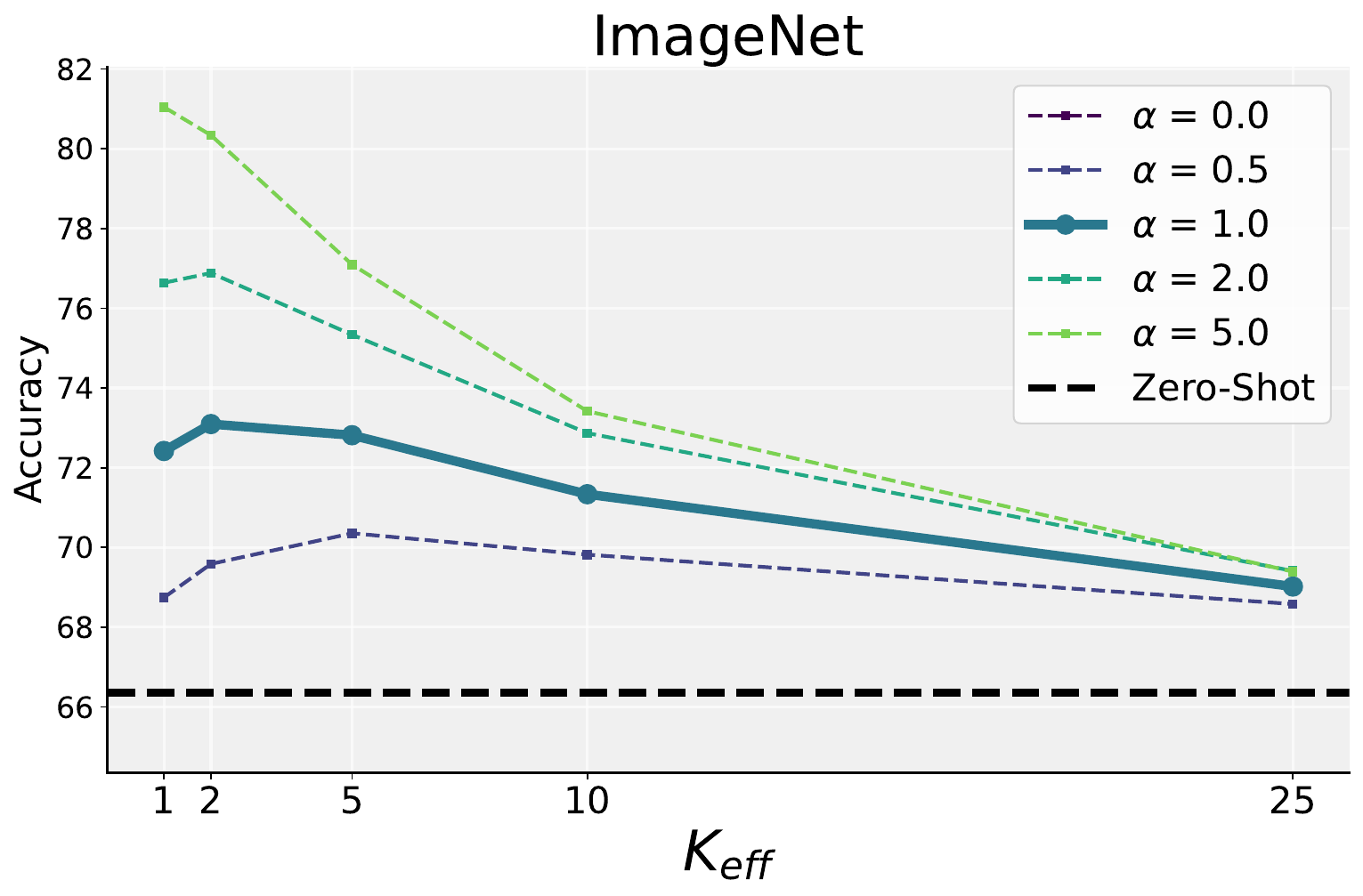}\hfill
    \includegraphics[width=.33\textwidth]{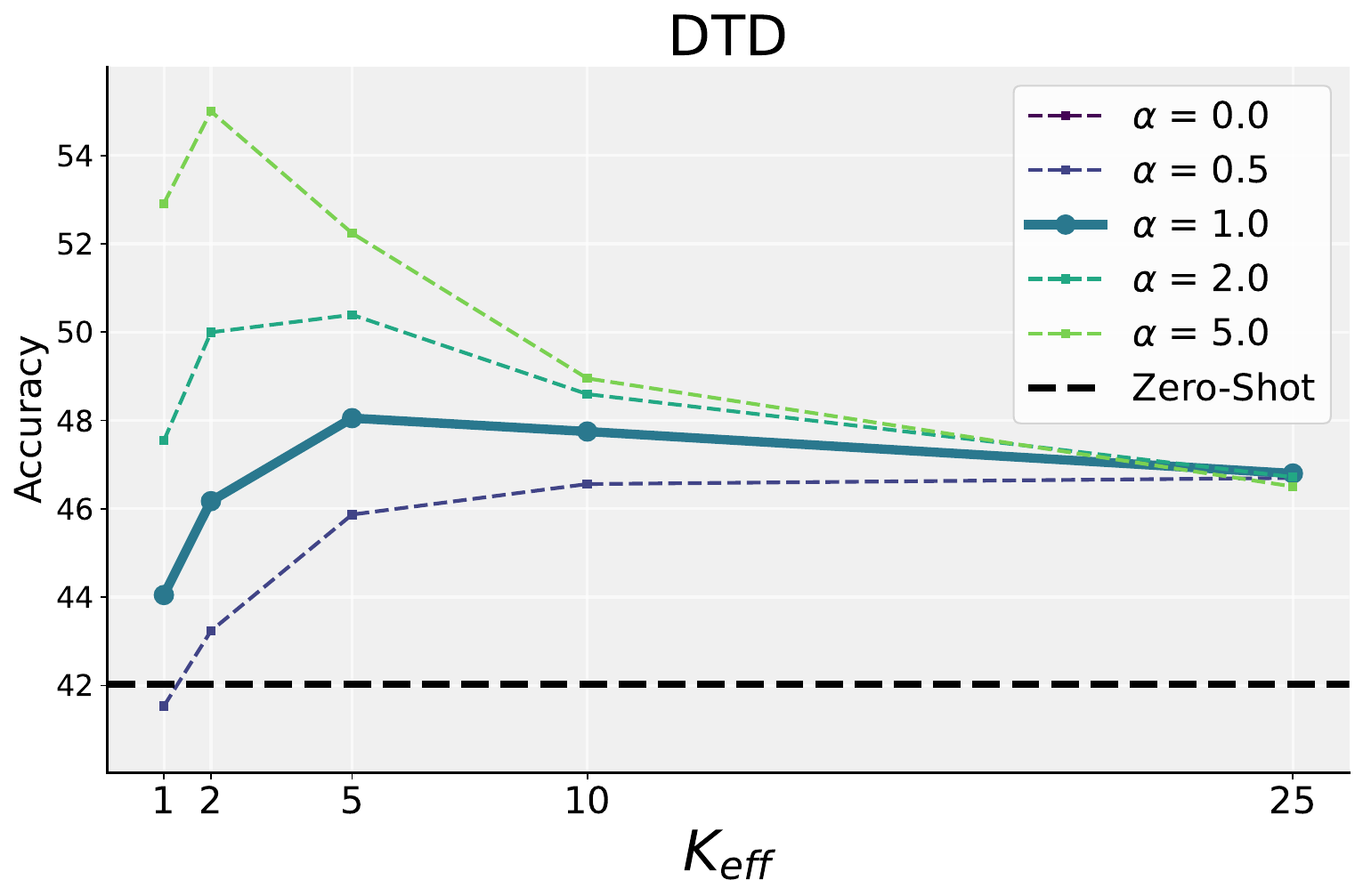}\hfill
    \includegraphics[width=.33\textwidth]{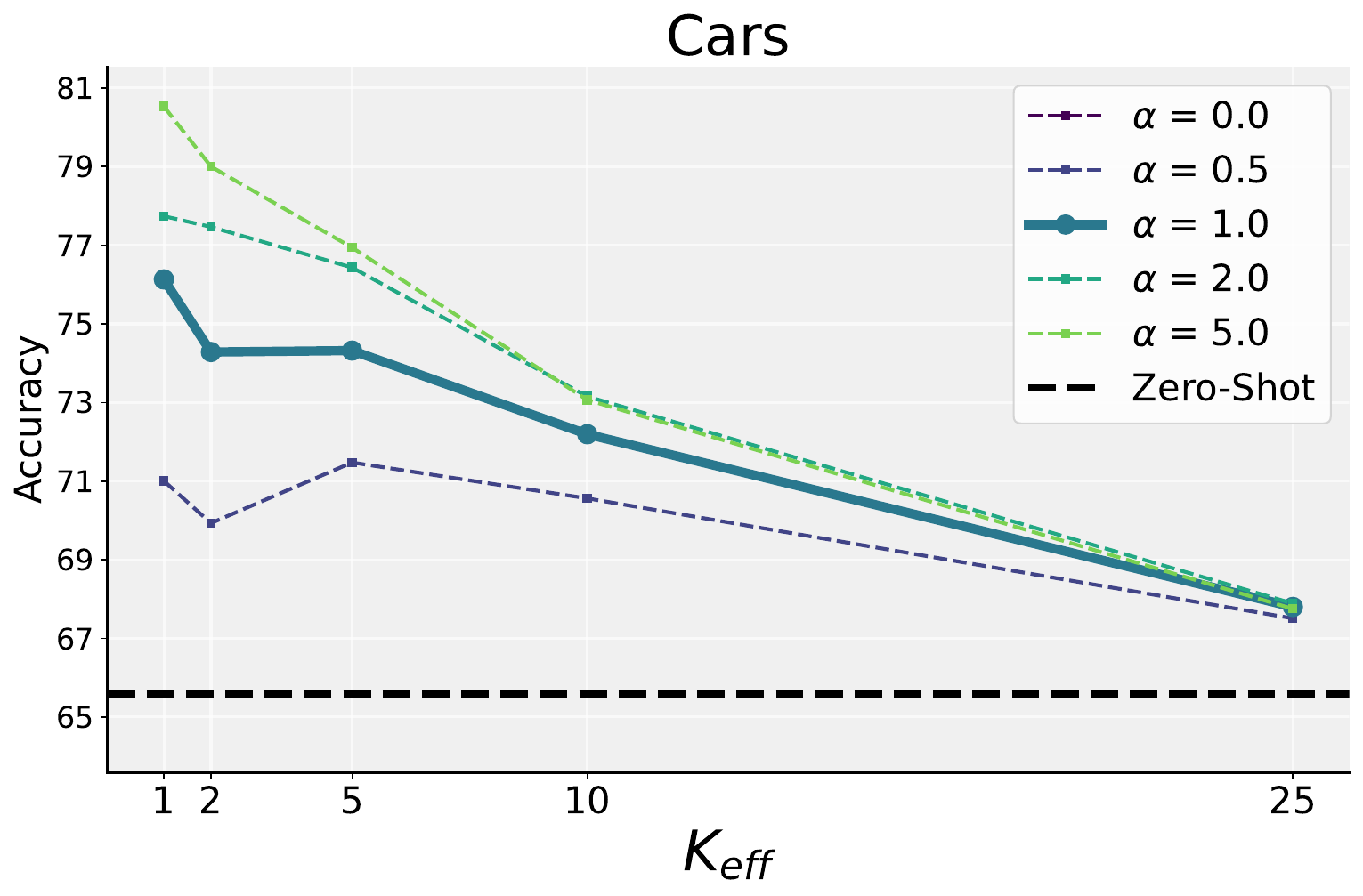}
    \caption{Effect of $\alpha$ with a batch size of 64 and 1 to 25 effective classes on ImageNet, DTD, and Cars dataset.}
    \label{fig:ablation_small}
\end{subfigure}

\vskip\baselineskip 

\begin{subfigure}[t]{0.97\textwidth}
    \centering
    \includegraphics[width=.33\textwidth]{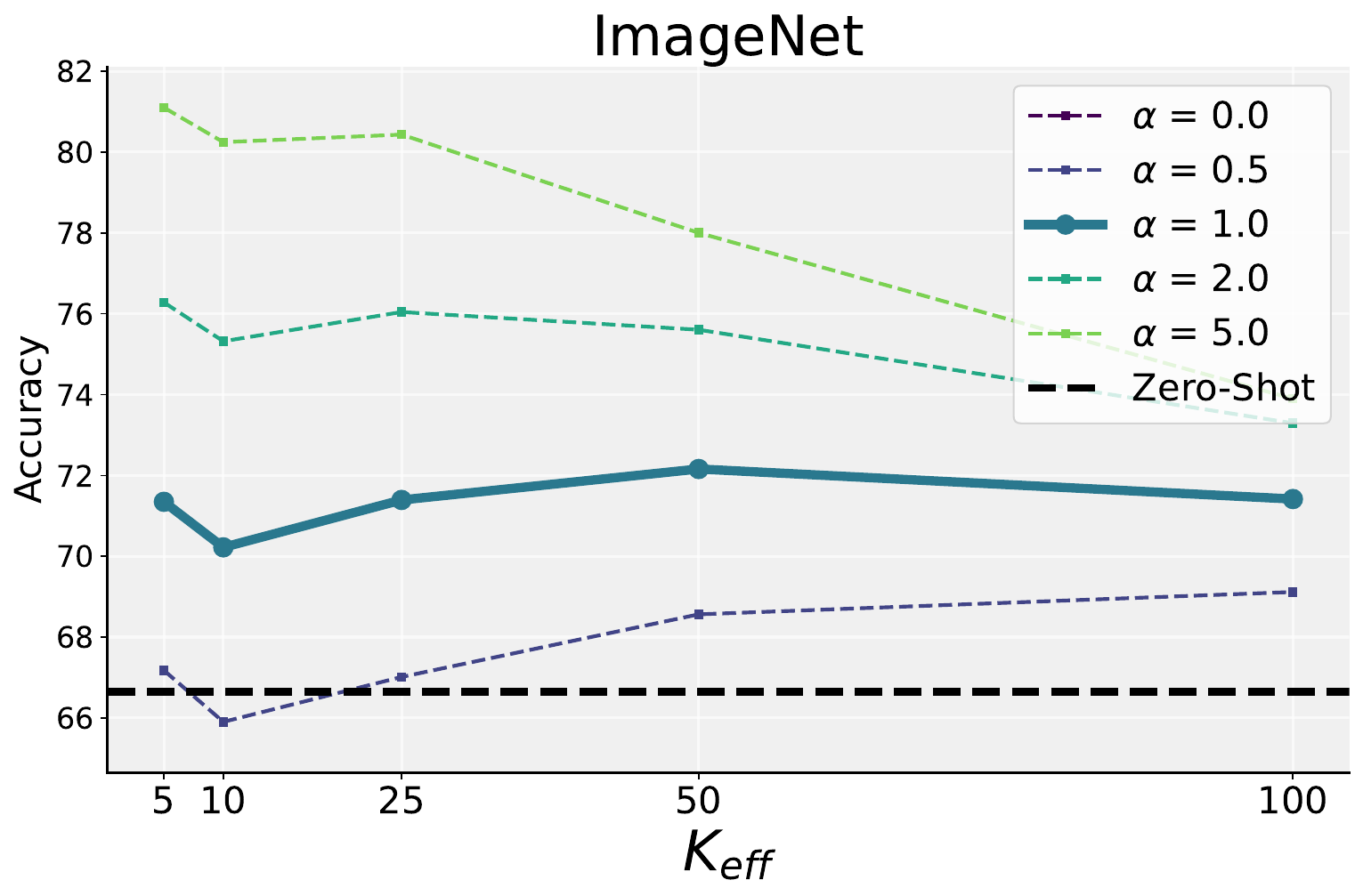}\hfill
    \includegraphics[width=.33\textwidth]{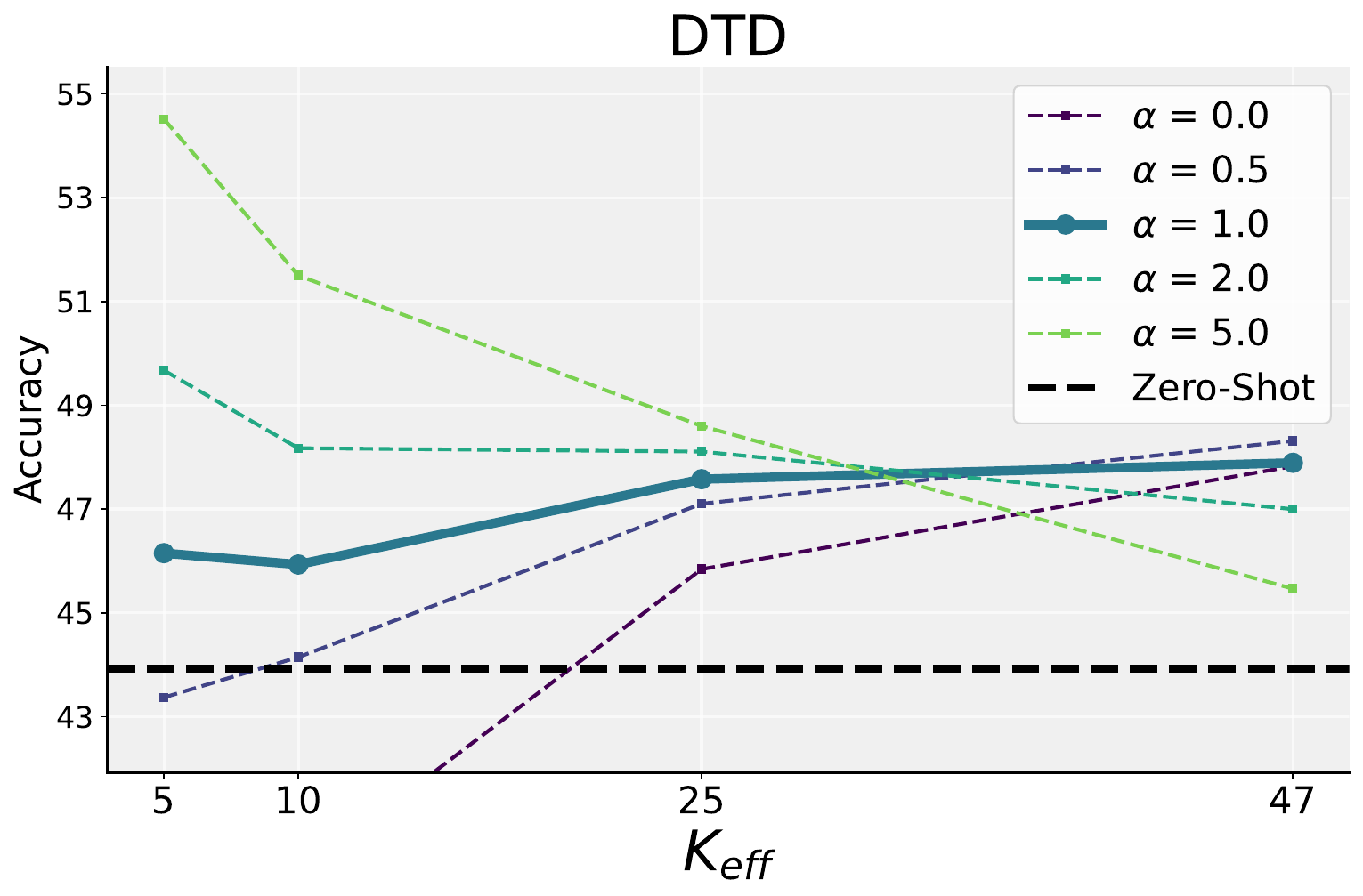}\hfill
    \includegraphics[width=.33\textwidth]{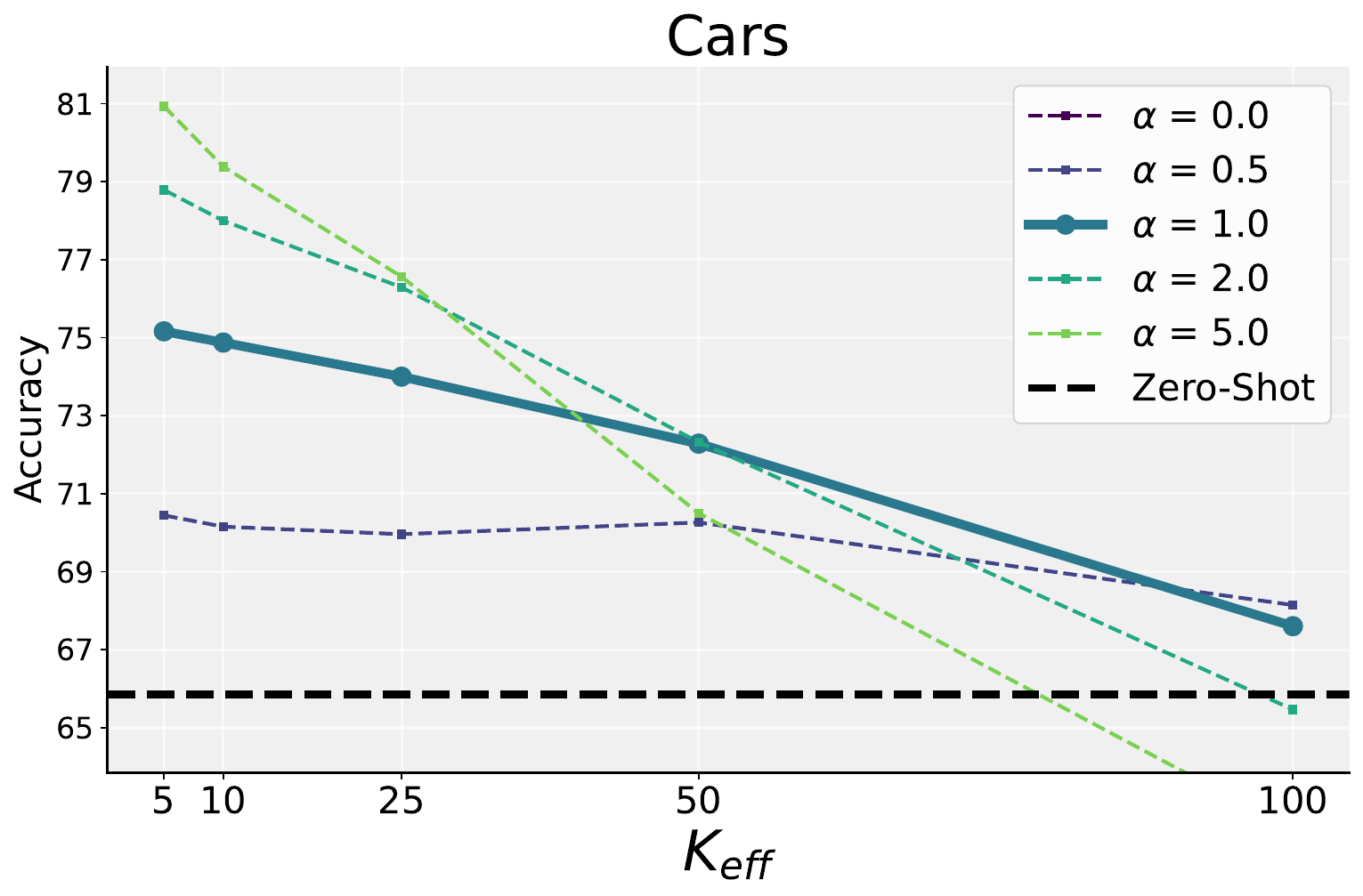}
    \caption{With a batch size of 1,000 and 5 to 100 effective classes on ImageNet, DTD, and Cars dataset.}
    \label{fig:ablation_big}
\end{subfigure}

\caption{Ablation study on the impact of the anchor weighting $\alpha$ across various numbers of effective classes ($K_{\text{eff}}$). The line corresponding to $\alpha=1$ (used in all our experiments) is highlighted with a wider stroke. Each reported performance is averaged over 1,000 tasks.}
\label{fig:ablation_anchor}
\end{figure*}

\paragraph{Runtime.} Table \ref{tab:runtime} highlights the efficiency of Stat${\cal A}$, capable of processing 50,000 samples in just a few seconds. Note that CLIP inference encompasses both image features computation and text prompts encoding. Stat${\cal A}$ acts as a robust additional processing, seamlessly applied atop the model after inference to enhance predictions. Its computational cost is minimal, adding negligible overhead compared to the initial inference.
\begin{table}[h]
    \caption{Runtime on a single Nvidia RTX 3090 GPU with 24 GB of memory for the ImageNet dataset.}
    \label{tab:runtime}
    \centering
    \resizebox{0.85\linewidth}{!}{%
    \begin{tabular}{cccc}
        \toprule 
        Batch size & CLIP inference & Stat${\cal A}$ & Total \\
        \midrule 
        1,000 & 6 sec.. & 0.1 sec. & 6.1 sec.\\
        50,000 & 5 min. & 15 sec. & $\sim$ 5 min.\\
        \bottomrule
    \end{tabular}}
\end{table}
\paragraph{Gaussian anchor term weighting factor.} We study the effect of the anchor term’s weighting factor in Figure \ref{fig:ablation_anchor}. Our observations show that tuning $\alpha$ to relatively large values can be highly beneficial when the ratio between the batch size and the number of effective classes is significantly greater than 1. For instance, Figure \ref{fig:ablation_small} demonstrates a +10\% improvement on the ImageNet dataset when $K_{\text{eff}} \leq 5$ with a batch size of 64, and a +13\% improvement when $5 \leq K_{\text{eff}}\leq 25$. In contrast, smaller $\alpha$ values can be more beneficial when the $K_{\text{eff}}$ increases, as depicted for DTD and Cars in Figure \ref{fig:ablation_big}. However, over-reliance on tuning $\alpha$ values would contradict our goal of maintaining a strong and versatile method across scenarios. While addressing this question lies beyond the scope of this paper, it presents an interesting avenue for future work.


\begin{table}[t]
    \centering
    \caption{Anchor strategies for various numbers of effective classes ($K_{\text{eff}}$) on the ImageNet dataset. Hard $\beta_k$ refers to Eq. \eqref{eq:hard_beta_k} and is used in all our experiments. CLIP (zero-shot) accuracy is $66.6$.}
    \label{tab:ablation_anchor_strategy}



    \begin{subtable}[t]{0.8\linewidth}
        \centering
        \caption{Batch size of 1,000.}
        \resizebox{\linewidth}{!}{%
        \begin{tabular}{lcccccc}
        \toprule
            \multicolumn{1}{c}{} & \multicolumn{5}{c}{$K_{eff}$} \\
            \cmidrule(lr){2-6}
            & 5 & 10 & 25 & 50 & 100  \\
            \midrule 
            Stat${\cal A}$ w/ soft $\beta_k$ & 70.3 & 69.2 & 70.7 & \textbf{72.7} & \textbf{72.8} \\
            Stat${\cal A}$ w/ hard $\beta_k$ &  \textbf{71.3} & \textbf{70.2} & \textbf{71.4} & 72.2 & 71.4 \\
        \bottomrule
        \end{tabular}}
        \label{tab:batch_size_1000}
    \end{subtable}


    \begin{subtable}[t]{0.8\linewidth}
        \centering
        \caption{Batch size of 5,000.}
        \resizebox{\linewidth}{!}{%
        \begin{tabular}{lccccc}
        \toprule
            \multicolumn{1}{c}{} & \multicolumn{5}{c}{$K_{eff}$} \\
            \cmidrule(lr){2-6}
            & 50 & 100 & 250 & 500 & 1,000 \\
            \midrule 
            Stat${\cal A}$ w/ soft $\beta_k$ & 69.7 & 69.2 & 70.8 & 69.4 & 66.1 \\
            Stat${\cal A}$ w/ hard $\beta_k$ & \textbf{70.8} & \textbf{70.4} & \textbf{71.1} & \textbf{69.5} & \textbf{66.5} \\
        \bottomrule
        \end{tabular}}
        \label{tab:batch_size_5000}
    \end{subtable}



\end{table}

\paragraph{$\beta_k$ computation.} We study the impact of our implementation choice of Section \ref{subsec:regularized} in Table \ref{tab:ablation_anchor_strategy}. Hard assignment proves to be more beneficial for large batches with very few classes. This can be attributed to the small residuals of the soft assignment, as predictions for a given class are never exactly zero (see softmax operator in Eq. \eqref{eq:z_update}). Soft $\beta_k$ brings greater improvement for smaller batches with relatively more classes. These insights suggest potential avenues for refining the anchor strategy, either by smoothing the small residuals of the prediction vectors or by better estimating the presence of a given class, in line with the observations made on the impact of the $\alpha$ weighting factor.

 \section{Conclusion}
\label{sec:conclusion}
In this work, we introduced a novel approach for test-time adaptation of VLMs. Our method, Stat${\cal A}$, leverages an effective anchor regularization term, allowing it to handle arbitrary number of effective classes. We hope this work will inspire further work for expanding the current research on VLMs adaptation, and for addressing more practical scenarios that go beyond simplistic data distribution assumptions.
\paragraph{Future work.} Our ablation studies suggest that incorporating a more adaptive anchor weighting could lead to substantial performance gains. Future investigation might also explore online refinement for our anchor term.

\section{Acknowledgments}
M.~Zanella is funded by the Walloon region under grant No.~2010235 (ARIAC by DIGITALWALLONIA4.AI). C.~Fuchs is funded by the MedReSyst
project, supported by FEDER and the Walloon Region. Part of the computational resources have been provided by the Consortium des Équipements de Calcul Intensif (CÉCI), funded by the Fonds de la Recherche Scientifique de Belgique (F.R.S.-FNRS) under Grant No. 2.5020.11 and by the Walloon Region.

{
    \small
    \bibliographystyle{ieeenat_fullname}
    \bibliography{main}

\begin{thebibliography}{46}
\providecommand{\natexlab}[1]{#1}
\providecommand{\url}[1]{\texttt{#1}}
\expandafter\ifx\csname urlstyle\endcsname\relax
  \providecommand{\doi}[1]{doi: #1}\else
  \providecommand{\doi}{doi: \begingroup \urlstyle{rm}\Url}\fi

\bibitem[Ando and Zhang(2007)]{scholkopf_learning_2007}
Rie~Kubota Ando and Tong Zhang.
\newblock Learning on {Graph} with {Laplacian} {Regularization}.
\newblock In \emph{Advances in {Neural} {Information} {Processing} {Systems} 19}, pages 25--32. The MIT Press, 2007.

\bibitem[Bossard et~al.(2014)Bossard, Guillaumin, and Van~Gool]{food}
Lukas Bossard, Matthieu Guillaumin, and Luc Van~Gool.
\newblock Food-101--mining discriminative components with random forests.
\newblock In \emph{Computer Vision--ECCV 2014: 13th European Conference, Zurich, Switzerland, September 6-12, 2014, Proceedings, Part VI 13}, pages 446--461. Springer, 2014.

\bibitem[Boudiaf et~al.(2020)Boudiaf, Ziko, Rony, Dolz, Piantanida, and Ben~Ayed]{boudiaf2020information}
Malik Boudiaf, Imtiaz Ziko, J{\'e}r{\^o}me Rony, Jos{\'e} Dolz, Pablo Piantanida, and Ismail Ben~Ayed.
\newblock Information maximization for few-shot learning.
\newblock \emph{Advances in Neural Information Processing Systems}, 33:\penalty0 2445--2457, 2020.

\bibitem[Boudiaf et~al.(2022)Boudiaf, Mueller, Ben~Ayed, and Bertinetto]{boudiaf2022parameter}
Malik Boudiaf, Romain Mueller, Ismail Ben~Ayed, and Luca Bertinetto.
\newblock Parameter-free online test-time adaptation.
\newblock In \emph{Proceedings of the IEEE/CVF Conference on Computer Vision and Pattern Recognition}, pages 8344--8353, 2022.

\bibitem[Boykov et~al.(2015)Boykov, Isack, Olsson, and Ben~Ayed]{Boykov-ICCV-05}
Y. Boykov, H.~N. Isack, C. Olsson, and I. Ben~Ayed.
\newblock Volumetric bias in segmentation and reconstruction: Secrets and solutions.
\newblock In \emph{IEEE/CVF International Conference on Computer Vision (ICCV)}, 2015.

\bibitem[Chen et~al.(2022)Chen, Wang, Darrell, and Ebrahimi]{chen2022contrastive}
Dian Chen, Dequan Wang, Trevor Darrell, and Sayna Ebrahimi.
\newblock Contrastive test-time adaptation.
\newblock In \emph{Proceedings of the IEEE/CVF Conference on Computer Vision and Pattern Recognition}, pages 295--305, 2022.

\bibitem[Cimpoi et~al.(2014)Cimpoi, Maji, Kokkinos, Mohamed, and Vedaldi]{dtd}
Mircea Cimpoi, Subhransu Maji, Iasonas Kokkinos, Sammy Mohamed, and Andrea Vedaldi.
\newblock Describing textures in the wild.
\newblock In \emph{Proceedings of the IEEE conference on computer vision and pattern recognition}, pages 3606--3613, 2014.

\bibitem[Deng et~al.(2009)Deng, Dong, Socher, Li, Li, and Fei-Fei]{imagenet}
Jia Deng, Wei Dong, Richard Socher, Li-Jia Li, Kai Li, and Li Fei-Fei.
\newblock Imagenet: A large-scale hierarchical image database.
\newblock In \emph{2009 IEEE Conference on Computer Vision and Pattern Recognition}, pages 248--255, 2009.

\bibitem[Farina et~al.(2024)Farina, Franchi, Iacca, Mancini, and Ricci]{farina2024frustratingly}
Matteo Farina, Gianni Franchi, Giovanni Iacca, Massimiliano Mancini, and Elisa Ricci.
\newblock Frustratingly easy test-time adaptation of vision-language models.
\newblock \emph{arXiv preprint arXiv:2405.18330}, 2024.

\bibitem[Fei-Fei et~al.(2004)Fei-Fei, Fergus, and Perona]{caltech101}
Li Fei-Fei, Rob Fergus, and Pietro Perona.
\newblock Learning generative visual models from few training examples: An incremental bayesian approach tested on 101 object categories.
\newblock In \emph{2004 conference on computer vision and pattern recognition workshop}, pages 178--178. IEEE, 2004.

\bibitem[Feng et~al.(2023)Feng, Yu, Liu, Khan, and Zuo]{Feng_2023_ICCV}
Chun-Mei Feng, Kai Yu, Yong Liu, Salman Khan, and Wangmeng Zuo.
\newblock Diverse data augmentation with diffusions for effective test-time prompt tuning.
\newblock In \emph{Proceedings of the IEEE/CVF International Conference on Computer Vision (ICCV)}, pages 2704--2714, 2023.

\bibitem[Gong et~al.(2022)Gong, Jeong, Kim, Kim, Shin, and Lee]{gong2022note}
Taesik Gong, Jongheon Jeong, Taewon Kim, Yewon Kim, Jinwoo Shin, and Sung-Ju Lee.
\newblock Note: Robust continual test-time adaptation against temporal correlation.
\newblock \emph{Advances in Neural Information Processing Systems}, 35:\penalty0 27253--27266, 2022.

\bibitem[Helber et~al.(2019)Helber, Bischke, Dengel, and Borth]{eurosat}
Patrick Helber, Benjamin Bischke, Andreas Dengel, and Damian Borth.
\newblock Eurosat: A novel dataset and deep learning benchmark for land use and land cover classification.
\newblock \emph{IEEE Journal of Selected Topics in Applied Earth Observations and Remote Sensing}, 12\penalty0 (7):\penalty0 2217--2226, 2019.

\bibitem[Karmanov et~al.(2024)Karmanov, Guan, Lu, El~Saddik, and Xing]{Karmanov_2024_CVPR}
Adilbek Karmanov, Dayan Guan, Shijian Lu, Abdulmotaleb El~Saddik, and Eric Xing.
\newblock Efficient test-time adaptation of vision-language models.
\newblock In \emph{Proceedings of the IEEE/CVF Conference on Computer Vision and Pattern Recognition (CVPR)}, pages 14162--14171, 2024.

\bibitem[Kearns et~al.(1998)Kearns, Mansour, and Ng]{Kearns-UAI-97}
Michael Kearns, Yishay Mansour, and Andrew~Y Ng.
\newblock An information-theoretic analysis of hard and soft assignment methods for clustering.
\newblock \emph{Learning in graphical models}, pages 495--520, 1998.

\bibitem[Khattak et~al.(2023)Khattak, Rasheed, Maaz, Khan, and Khan]{khattak2023maple}
Muhammad~Uzair Khattak, Hanoona Rasheed, Muhammad Maaz, Salman Khan, and Fahad~Shahbaz Khan.
\newblock Maple: Multi-modal prompt learning.
\newblock In \emph{Proceedings of the IEEE/CVF Conference on Computer Vision and Pattern Recognition}, pages 19113--19122, 2023.

\bibitem[Khoury et~al.(2024)Khoury, Zanella, G{\'e}rin, Godelaine, Macq, Mahmoudi, Vleeschouwer, and Ayed]{elkhoury2024enhancing}
Karim~El Khoury, Maxime Zanella, Beno{\^\i}t G{\'e}rin, Tiffanie Godelaine, Beno{\^\i}t Macq, Sa{\"i}d Mahmoudi, Christophe~De Vleeschouwer, and Ismail~Ben Ayed.
\newblock Enhancing remote sensing vision-language models for zero-shot scene classification.
\newblock \emph{arXiv preprint arXiv:2409.00698}, 2024.

\bibitem[Krause et~al.(2013)Krause, Stark, Deng, and Fei-Fei]{cars}
Jonathan Krause, Michael Stark, Jia Deng, and Li Fei-Fei.
\newblock 3d object representations for fine-grained categorization.
\newblock In \emph{Proceedings of the IEEE international conference on computer vision workshops}, pages 554--561, 2013.

\bibitem[Liu et~al.()Liu, Lee, Park, Kim, Yang, Hwang, and Yang]{liulearning}
Yanbin Liu, Juho Lee, Minseop Park, Saehoon Kim, Eunho Yang, Sung~Ju Hwang, and Yi Yang.
\newblock Learning to propagate labels: Transductive propagation network for few-shot learning.
\newblock In \emph{International Conference on Learning Representations}.

\bibitem[Ma et~al.(2024)Ma, Zhang, Guo, and Xu]{NEURIPS2023_cdd06402}
Xiaosong Ma, Jie Zhang, Song Guo, and Wenchao Xu.
\newblock Swapprompt: Test-time prompt adaptation for vision-language models.
\newblock \emph{Advances in Neural Information Processing Systems}, 36, 2024.

\bibitem[Maji et~al.(2013)Maji, Rahtu, Kannala, Blaschko, and Vedaldi]{aircraft}
Subhransu Maji, Esa Rahtu, Juho Kannala, Matthew Blaschko, and Andrea Vedaldi.
\newblock Fine-grained visual classification of aircraft.
\newblock \emph{arXiv preprint arXiv:1306.5151}, 2013.

\bibitem[Manli et~al.(2022)Manli, Weili, De-An, Zhiding, Tom, Anima, and Chaowei]{shu2022tpt}
Shu Manli, Nie Weili, Huang De-An, Yu Zhiding, Goldstein Tom, Anandkumar Anima, and Xiao Chaowei.
\newblock Test-time prompt tuning for zero-shot generalization in vision-language models.
\newblock In \emph{NeurIPS}, 2022.

\bibitem[Martin et~al.(2022)Martin, Boudiaf, Chouzenoux, Pesquet, and Ben~Ayed]{martin2022towards}
S{\'e}gol{\`e}ne Martin, Malik Boudiaf, Emilie Chouzenoux, Jean-Christophe Pesquet, and Ismail Ben~Ayed.
\newblock Towards practical few-shot query sets: transductive minimum description length inference.
\newblock \emph{Advances in Neural Information Processing Systems}, 35:\penalty0 34677--34688, 2022.

\bibitem[Martin et~al.(2024)Martin, Huang, Shakeri, Pesquet, and Ben~Ayed]{Martin_2024_CVPR}
S\'egol\`ene Martin, Yunshi Huang, Fereshteh Shakeri, Jean-Christophe Pesquet, and Ismail Ben~Ayed.
\newblock Transductive zero-shot and few-shot clip.
\newblock In \emph{Proceedings of the IEEE/CVF Conference on Computer Vision and Pattern Recognition (CVPR)}, pages 28816--28826, 2024.

\bibitem[Nilsback and Zisserman(2008)]{flower}
Maria-Elena Nilsback and Andrew Zisserman.
\newblock Automated flower classification over a large number of classes.
\newblock In \emph{2008 Sixth Indian conference on computer vision, graphics \& image processing}, pages 722--729. IEEE, 2008.

\bibitem[Osowiechi et~al.(2024)Osowiechi, Noori, Hakim, Yazdanpanah, Bahri, Cheraghalikhani, Dastani, Beizaee, Ayed, and Desrosiers]{osowiechi2024watt}
David Osowiechi, Mehrdad Noori, Gustavo Adolfo~Vargas Hakim, Moslem Yazdanpanah, Ali Bahri, Milad Cheraghalikhani, Sahar Dastani, Farzad Beizaee, Ismail~Ben Ayed, and Christian Desrosiers.
\newblock Watt: Weight average test-time adaption of clip.
\newblock \emph{arXiv preprint arXiv:2406.13875}, 2024.

\bibitem[Ouali et~al.(2023)Ouali, Bulat, Matinez, and Tzimiropoulos]{ouali2023black}
Yassine Ouali, Adrian Bulat, Brais Matinez, and Georgios Tzimiropoulos.
\newblock Black box few-shot adaptation for vision-language models.
\newblock In \emph{Proceedings of the IEEE/CVF International Conference on Computer Vision}, pages 15534--15546, 2023.

\bibitem[Parkhi et~al.(2012)Parkhi, Vedaldi, Zisserman, and Jawahar]{pets}
Omkar~M Parkhi, Andrea Vedaldi, Andrew Zisserman, and CV Jawahar.
\newblock Cats and dogs.
\newblock In \emph{2012 IEEE conference on computer vision and pattern recognition}, pages 3498--3505. IEEE, 2012.

\bibitem[Petersen et~al.(2008)Petersen, Pedersen, et~al.]{petersen2008matrix}
Kaare~Brandt Petersen, Michael~Syskind Pedersen, et~al.
\newblock The matrix cookbook.
\newblock \emph{Technical University of Denmark}, 7\penalty0 (15):\penalty0 510, 2008.

\bibitem[Radford et~al.(2021)Radford, Kim, Hallacy, Ramesh, Goh, Agarwal, Sastry, Askell, Mishkin, Clark, et~al.]{radford2021learning}
Alec Radford, Jong~Wook Kim, Chris Hallacy, Aditya Ramesh, Gabriel Goh, Sandhini Agarwal, Girish Sastry, Amanda Askell, Pamela Mishkin, Jack Clark, et~al.
\newblock Learning transferable visual models from natural language supervision.
\newblock In \emph{International conference on machine learning}, pages 8748--8763. PMLR, 2021.

\bibitem[Soomro et~al.(2012)Soomro, Zamir, and Shah]{ucf101}
Khurram Soomro, Amir~Roshan Zamir, and Mubarak Shah.
\newblock Ucf101: A dataset of 101 human actions classes from videos in the wild.
\newblock \emph{arXiv preprint arXiv:1212.0402}, 2012.

\bibitem[Stojni\'c et~al.(2024)Stojni\'c, Kalantidis, and Tolias]{Stojnic_2024_CVPR}
Vladan Stojni\'c, Yannis Kalantidis, and Giorgos Tolias.
\newblock Label propagation for zero-shot classification with vision-language models.
\newblock In \emph{Proceedings of the IEEE/CVF Conference on Computer Vision and Pattern Recognition (CVPR)}, 2024.

\bibitem[Wang et~al.(2021)Wang, Shelhamer, Liu, Olshausen, and Darrell]{wangtent}
Dequan Wang, Evan Shelhamer, Shaoteng Liu, Bruno Olshausen, and Trevor Darrell.
\newblock Tent: Fully test-time adaptation by entropy minimization.
\newblock In \emph{International Conference on Learning Representations}, 2021.

\bibitem[Wang et~al.(2024)Wang, Liang, Sheng, He, Wang, and Tan]{wang_hard_to_beat_2024}
Zhengbo Wang, Jian Liang, Lijun Sheng, Ran He, Zilei Wang, and Tieniu Tan.
\newblock A hard-to-beat baseline for training-free clip-based adaptation.
\newblock In \emph{The Twelfth International Conference on Learning Representations}, 2024.

\bibitem[Xiao et~al.(2010)Xiao, Hays, Ehinger, Oliva, and Torralba]{sun397}
Jianxiong Xiao, James Hays, Krista~A Ehinger, Aude Oliva, and Antonio Torralba.
\newblock Sun database: Large-scale scene recognition from abbey to zoo.
\newblock In \emph{2010 IEEE computer society conference on computer vision and pattern recognition}, pages 3485--3492. IEEE, 2010.

\bibitem[Yuan et~al.(2023)Yuan, Xie, and Li]{yuan2023robust}
Longhui Yuan, Binhui Xie, and Shuang Li.
\newblock Robust test-time adaptation in dynamic scenarios.
\newblock In \emph{Proceedings of the IEEE/CVF Conference on Computer Vision and Pattern Recognition}, pages 15922--15932, 2023.

\bibitem[Yuille and Rangarajan(2001)]{CCCP-2001}
A.~L. Yuille and A. Rangarajan.
\newblock The concave-convex procedure (cccp).
\newblock In \emph{Neural Information Processing Systems (NeurIPS)}, 2001.

\bibitem[Zanella and Ben~Ayed(2024{\natexlab{a}})]{zanella2024low}
Maxime Zanella and Ismail Ben~Ayed.
\newblock Low-rank few-shot adaptation of vision-language models.
\newblock In \emph{Proceedings of the IEEE/CVF Conference on Computer Vision and Pattern Recognition}, pages 1593--1603, 2024{\natexlab{a}}.

\bibitem[Zanella and Ben~Ayed(2024{\natexlab{b}})]{zanella2024test}
Maxime Zanella and Ismail Ben~Ayed.
\newblock On the test-time zero-shot generalization of vision-language models: Do we really need prompt learning?
\newblock In \emph{Proceedings of the IEEE/CVF Conference on Computer Vision and Pattern Recognition}, pages 23783--23793, 2024{\natexlab{b}}.

\bibitem[Zanella et~al.(2024{\natexlab{a}})Zanella, G{\'e}rin, and Ayed]{zanella2024boosting}
Maxime Zanella, Beno{\^\i}t G{\'e}rin, and Ismail~Ben Ayed.
\newblock Boosting vision-language models with transduction.
\newblock \emph{Neural Information Processing Systems (NeurIPS)}, 2024{\natexlab{a}}.

\bibitem[Zanella et~al.(2024{\natexlab{b}})Zanella, Shakeri, Huang, Bahig, and Ayed]{zanella2024boostinghisto}
Maxime Zanella, Fereshteh Shakeri, Yunshi Huang, Houda Bahig, and Ismail~Ben Ayed.
\newblock Boosting vision-language models for histopathology classification: Predict all at once.
\newblock In \emph{International Workshop on Foundation Models for General Medical AI}, pages 153--162. Springer, 2024{\natexlab{b}}.

\bibitem[Zhang et~al.(2022)Zhang, Zhang, Fang, Gao, Li, Dai, Qiao, and Li]{zhang2022tip}
Renrui Zhang, Wei Zhang, Rongyao Fang, Peng Gao, Kunchang Li, Jifeng Dai, Yu Qiao, and Hongsheng Li.
\newblock Tip-adapter: Training-free adaption of clip for few-shot classification.
\newblock In \emph{European conference on computer vision}, pages 493--510. Springer, 2022.

\bibitem[Zhang et~al.(2024)Zhang, Zhu, Tang, Ma, Zhou, and Zhang]{Zhang_2024_CVPR}
Yabin Zhang, Wenjie Zhu, Hui Tang, Zhiyuan Ma, Kaiyang Zhou, and Lei Zhang.
\newblock Dual memory networks: A versatile adaptation approach for vision-language models.
\newblock In \emph{Proceedings of the IEEE/CVF Conference on Computer Vision and Pattern Recognition (CVPR)}, pages 28718--28728, 2024.

\bibitem[Zhou et~al.(2003)Zhou, Bousquet, Lal, Weston, and Sch\"{o}lkopf]{NIPS2003_87682805}
Dengyong Zhou, Olivier Bousquet, Thomas Lal, Jason Weston, and Bernhard Sch\"{o}lkopf.
\newblock Learning with local and global consistency.
\newblock In \emph{Advances in Neural Information Processing Systems}. MIT Press, 2003.

\bibitem[Zhou et~al.(2022)Zhou, Yang, Loy, and Liu]{coop}
Kaiyang Zhou, Jingkang Yang, Chen~Change Loy, and Ziwei Liu.
\newblock Learning to prompt for vision-language models.
\newblock \emph{International Journal of Computer Vision (IJCV)}, 2022.

\bibitem[Ziko et~al.(2020)Ziko, Dolz, Granger, and Ayed]{ziko2020laplacian}
Imtiaz Ziko, Jose Dolz, Eric Granger, and Ismail~Ben Ayed.
\newblock Laplacian regularized few-shot learning.
\newblock In \emph{International conference on machine learning}, pages 11660--11670. PMLR, 2020.

\end{thebibliography}
}

\clearpage
\setcounter{page}{1}
\onecolumn

\appendix
\begin{center}
    {\Large \textbf{Realistic Test-Time Adaptation of Vision-Language Models}}\\[10pt]
    {\Large Supplementary Material}\\[5pt]
\end{center}
\section{Additional experimental details}
\label{sec:details_exp}
\paragraph{Datasets.} We follow the setting of previous work~\cite{coop}. We assess our method on ImageNet~\cite{imagenet} and ten other datasets for fine-grained classification of scenes (SUN397~\cite{sun397}), aircraft types (Aircraft~\cite{aircraft}), satellite imagery (EuroSAT~\cite{eurosat}), automobiles (StanfordCars~\cite{cars}), food items (Food101~\cite{food}), pet breeds (OxfordPets~\cite{pets}), flowers (Flower102~\cite{flower}), general objects (Caltech101~\cite{caltech101}), textures (DTD~\cite{dtd}) and human actions (UCF101~\cite{ucf101}). These diverse datasets provide a comprehensive visual classification benchmark. Additional information on the statistics of each dataset is provided in Table \ref{tab:appendix_datasets}.

\paragraph{Hyper-parameters} Generalization to unseen cases is crucial for TTA methods. Therefore, optimizing hyper-parameters for each scenario would require access to labels and prior knowledge of the specific scenario encountered during testing, which goes against the core purpose of the TTA approach. For instance, we found that TDA largely relies on dataset-specific hyper-parameters without clear guidance on how to tune them for a new dataset. Similarly, DMN conducts an hyper-parameter search in order to find the optimal balance between the logits obtained from the text prompts and the logits from their model's memory, using knowledge from ground truth labels. To ensure fairness in comparison, we use hyper-parameters optimized for ImageNet for all reported experiments.

\paragraph{Comparative methods.} We use the same handcrafted prompts for all methods, which are listed in Table \ref{tab:app_all_prompts}. Due to the more versatile scenarios studied in this paper, we find that our centroid initialization based on text embeddings much more robust, especially when the number of effective classes is reduced. Therefore, we implement it for TransCLIP instead of their original initialization based on the top-confident samples for each class. For ZLaP and Dirichlet we follow the hyper-parameters of their official implementation. For TDA and DMN, following our discussion, we use the hyper-parameters optimized for ImageNet. For TDA, this means the positive logits mixing coefficients is set to $2$, while the negative logits mixing coefficient is set to $0.117$. For DMN, since we only consider zero-shot scenarios, we only need to set the coefficient relative to the dynamic memory, which is set to $1$. For TENT, we use a learning rate of $0.001$ and 10 steps of gradient descent per batch.

\section{Kullback-Leibler divergence between two multivariate Gaussian distributions}
\label{sec:kullback}
Let $\mathcal{N}(\boldsymbol{\mu}_p, \boldsymbol{\Sigma}_p)$ and $\mathcal{N}(\boldsymbol{\mu}_q, \boldsymbol{\Sigma}_q)$ two multivariate Gaussian distributions with respective probability density functions $p$ and $q$. Then, we have
\begin{align}
    \mbox{KL}(p||q) &= \int_x p(x) \log \frac{p(x)}{q(x)} \mathop{}\!\mathrm{d}x  \\
    &= \mathbb{E}_p [\log(p) - \log(q) ]\\
    &= \mathbb{E}_p [\frac{1}{2} \log \frac{|\boldsymbol{\Sigma}_q|}{|\boldsymbol{\Sigma}_p|} - \frac{1}{2}(\mathbf{x} - \boldsymbol{\mu}_p) ^\top \boldsymbol{\Sigma}_p^{-1}(\mathbf{x} - \boldsymbol{\mu}_p) + \frac{1}{2}(\mathbf{x} - \boldsymbol{\mu}_q) ^\top \boldsymbol{\Sigma}_q^{-1}(\mathbf{x} - \boldsymbol{\mu}_q)] \\
    &= \frac{1}{2} \mathbb{E}_p [\log \frac{|\boldsymbol{\Sigma}_q|}{|\boldsymbol{\Sigma}_p|}] - \frac{1}{2}\mathbb{E}_p[(\mathbf{x} - \boldsymbol{\mu}_p) ^\top \boldsymbol{\Sigma}_p^{-1}(\mathbf{x} - \boldsymbol{\mu}_p)] + \frac{1}{2}\mathbb{E}_p[(\mathbf{x} - \boldsymbol{\mu}_q) ^\top \boldsymbol{\Sigma}_q^{-1}(\mathbf{x} - \boldsymbol{\mu}_q)] \\
    &= \frac{1}{2} \log \frac{|\boldsymbol{\Sigma}_q|}{|\boldsymbol{\Sigma}_p|} - \frac{1}{2}\mathbb{E}_p[(\mathbf{x} - \boldsymbol{\mu}_p) ^\top \boldsymbol{\Sigma}_p^{-1}(\mathbf{x} - \boldsymbol{\mu}_p)] + \frac{1}{2}\mathbb{E}_p[(\mathbf{x} - \boldsymbol{\mu}_q) ^\top \boldsymbol{\Sigma}_q^{-1}(\mathbf{x} - \boldsymbol{\mu}_q)].
\end{align}
We can rewrite the second term as
\begin{equation}
    (\mathbf{x} - \boldsymbol{\mu}_p) ^\top \boldsymbol{\Sigma}_p^{-1}(\mathbf{x} - \boldsymbol{\mu}_p) = \Tr \left( (\mathbf{x} - \boldsymbol{\mu}_p) ^\top \boldsymbol{\Sigma}_p^{-1}(\mathbf{x} - \boldsymbol{\mu}_p) \right) = \Tr \left( (\mathbf{x} - \boldsymbol{\mu}_p) (\mathbf{x} - \boldsymbol{\mu}_p)^\top  \boldsymbol{\Sigma}_p^{-1}  \right) 
\end{equation}
by using
\begin{equation}
    \Tr(ABC) = \Tr(BCA) = \Tr(CAB).
\end{equation}
For the third term, since we assume $\mathbf{x}$ follows a Gaussian distribution $\mathcal{N}(\boldsymbol{\mu}_p, \boldsymbol{\Sigma}_p)$, we have (see Matrix cookbook \cite{petersen2008matrix} Eq. 380 of Section 8.2)
\begin{equation}
    \mathbb{E}[(\mathbf{x}-\boldsymbol{\mu}_q)^\top \boldsymbol{\Sigma}_q^{-1} (\mathbf{x}-\boldsymbol{\mu}_q)] = (\boldsymbol{\mu}_p - \boldsymbol{\mu}_q)^\top \boldsymbol{\Sigma}_q^{-1} (\boldsymbol{\mu}_p - \boldsymbol{\mu}_q) + \Tr(\boldsymbol{\Sigma}_q^{-1}\boldsymbol{\Sigma}_p)
\end{equation}
And therefore
\begin{align}
    \mbox{KL}(p||q) &= \frac{1}{2} \log \frac{|\boldsymbol{\Sigma}_q|}{|\boldsymbol{\Sigma}_p|} - \frac{1}{2}\mathbb{E}_p[\Tr \left( ( \mathbf{x} - \boldsymbol{\mu}_p)  (\mathbf{x} - \boldsymbol{\mu}_p)^\top \boldsymbol{\Sigma}_p^{-1} \right)] + \frac{1}{2}((\boldsymbol{\mu}_p - \boldsymbol{\mu}_q)^\top \boldsymbol{\Sigma}_q^{-1}(\boldsymbol{\mu}_p - \boldsymbol{\mu}_q) + \Tr\{\boldsymbol{\Sigma}_q^{-1}\boldsymbol{\Sigma}_p\})\\
    &= \frac{1}{2} \log \frac{|\boldsymbol{\Sigma}_q|}{|\boldsymbol{\Sigma}_p|} - \frac{1}{2}(\Tr \left( \mathbb{E}_p [ (\mathbf{x} - \boldsymbol{\mu}_p) (\mathbf{x} - \boldsymbol{\mu}_p)^\top] \boldsymbol{\Sigma}_p^{-1} \right) + \frac{1}{2}((\boldsymbol{\mu}_p - \boldsymbol{\mu}_q)^\top \boldsymbol{\Sigma}_q^{-1}(\boldsymbol{\mu}_p - \boldsymbol{\mu}_q) + \Tr(\boldsymbol{\Sigma}_q^{-1}\boldsymbol{\Sigma}_p))
\end{align}
by using the fact that trace and expectation can be interchanged. Moreover, 
\begin{equation}
    \mathbb{E}_p[(\mathbf{x} - \boldsymbol{\mu}_p)(\mathbf{x} - \boldsymbol{\mu}_p)^\top] = \boldsymbol{\Sigma}_p
\end{equation}
which simplifies further the second term of the sum and gives
\begin{align}
    \mbox{KL}(p||q) &= \frac{1}{2} \log \frac{|\boldsymbol{\Sigma}_q|}{|\boldsymbol{\Sigma}_p|} - \frac{1}{2}(\Tr( \boldsymbol{I}_k ) + \frac{1}{2}((\boldsymbol{\mu}_p - \boldsymbol{\mu}_q)^\top \boldsymbol{\Sigma}_q^{-1}(\boldsymbol{\mu}_p - \boldsymbol{\mu}_q) + \Tr(\boldsymbol{\Sigma}_q^{-1}\boldsymbol{\Sigma}_p)) \\
    &= \frac{1}{2} \log \frac{|\boldsymbol{\Sigma}_q|}{|\boldsymbol{\Sigma}_p|} - \frac{d}{2} + \frac{1}{2}((\boldsymbol{\mu}_p - \boldsymbol{\mu}_q)^\top \boldsymbol{\Sigma}_q^{-1}(\boldsymbol{\mu}_p - \boldsymbol{\mu}_q) + \Tr(\boldsymbol{\Sigma}_q^{-1}\boldsymbol{\Sigma}_p))\\
    &= \frac{1}{2} \left(\log \frac{|\boldsymbol{\Sigma}_q|}{|\boldsymbol{\Sigma}_p|} - d + (\boldsymbol{\mu}_p - \boldsymbol{\mu}_q)^\top \boldsymbol{\Sigma}_q^{-1}(\boldsymbol{\mu}_p - \boldsymbol{\mu}_q) + \Tr(\boldsymbol{\Sigma}_q^{-1}\boldsymbol{\Sigma}_p) \right)
\end{align}
with $d$ the number of dimensions.

\section{Detailed derivations of the regularized updates of the parameters}
We write $p_k$ and $q_k$ the respective probability density functions of the distributions $\mathcal{N}(\boldsymbol{\mu}_k, \boldsymbol{\Sigma}_k)$  and $\mathcal{N}(\boldsymbol{\mu'}_k, \boldsymbol{\Sigma'}_k)$. With the results of Appendix Section \ref{sec:kullback}, we have
\begin{align}
\mathcal{A}(\boldsymbol{\mu}, \boldsymbol{\Sigma}) = \sum\limits_{k} \mbox{KL}(q_k||p_k) =\frac{1}{2}   \sum\limits_{k} (\boldsymbol{\mu'}_k - \boldsymbol{\mu}_k)^\top \boldsymbol{\Sigma}_k^{-1}(\boldsymbol{\mu'}_k - \boldsymbol{\mu}_k) \notag  + \Tr (\boldsymbol{\Sigma}_k^{-1}\boldsymbol{\Sigma'}_k) +\log \frac{|\boldsymbol{\Sigma}_k|}{| \boldsymbol{\Sigma'}_k|} - d .
\end{align} 
\subsection{With respect to $\boldsymbol{\mu}_k$}

\begin{align}
    \frac{\partial {\cal L}}{\boldsymbol{\mu}_k}
    &=\frac{\partial }{\boldsymbol{\mu}_k} \left( -  \sum_{i\in \cal Q}  {\mathbf z}_{i}^\top \log ({\mathbf p}_{i})
 -\sum_{i\in \cal Q} \sum_{j \in \cal Q} w_{ij} \mathbf{z}_{i}^\top \mathbf{z}_{j} + \sum_{i \in \cal Q} \mbox{KL} (\mathbf{z}_{i} || \hat{\mathbf{y}}_{i}) + \alpha \mathcal{A}(\boldsymbol{\mu}, \boldsymbol{\Sigma}) \right) \\
    &= \frac{\partial }{\boldsymbol{\mu}_k}\left(-  \sum_{i\in \cal Q}  {\mathbf z}_{i}^\top \log ({\mathbf p}_{i})
 -\sum_{i\in \cal Q} \sum_{j \in \cal Q} w_{ij} \mathbf{z}_{i}^\top \mathbf{z}_{j} + \sum_{i \in \cal Q} \mbox{KL} (\mathbf{z}_{i} || \hat{\mathbf{y}}_{i}) + \alpha \sum_{l=1}^{K} \mbox{KL} (\mathbf{q}_{l} || \mathbf{p}_{l}) \right) \\
& = \frac{\partial }{\boldsymbol{\mu}_k} \left(-  \sum_{i\in \cal Q}  z_{i,k} \left(-\frac{1}{2} \log |\boldsymbol{\Sigma}_k| -{\frac {1}{2}}({\mathbf f}_i - \boldsymbol{\mu}_k)^\top \boldsymbol{\Sigma}_k^{-1}({\mathbf f}_i - \boldsymbol{\mu}_k) \right) + \frac{\alpha}{2}  (\boldsymbol{\mu'}_k - \boldsymbol{\mu}_k)^\top \boldsymbol{\Sigma}_k^{-1}(\boldsymbol{\mu'}_k - \boldsymbol{\mu}_k) \right) \\
& =  -\sum_{i\in \cal Q} z_{i,k} \left(\boldsymbol{\Sigma}_k^{-1} ({\mathbf f}_i - \boldsymbol{\mu}_k) \right) +  \alpha \boldsymbol{\Sigma}_k^{-1} (\boldsymbol{\mu'}_k - \boldsymbol{\mu}_k)
\end{align}
Observe that the term $ \alpha \boldsymbol{\Sigma}_k^{-1} (\boldsymbol{\mu'}_k - \boldsymbol{\mu}_k)$ directly comes from the derivative of our statistical anchor $\mathcal{A}(\boldsymbol{\mu}, \boldsymbol{\Sigma})$ with regard to $\boldsymbol{\mu}_k$. By setting the derivative to 0
\begin{align}
    -\sum_{i\in \cal Q}  z_{i,k} ({\mathbf f}_i - \boldsymbol{\mu}_k) -  \alpha (\boldsymbol{\mu'}_k - \boldsymbol{\mu}_k) &= 0 \\
    \sum_{i\in \cal Q}  z_{i,k} {\mathbf f}_i +  \alpha   \boldsymbol{\mu'}_k &= \sum_{i\in \cal Q}  z_{i,k} \boldsymbol{\mu}_k + \alpha  \boldsymbol{\mu}_k  \\
    \sum_{i\in \cal Q}  z_{i,k} {\mathbf f}_i +  \alpha  \boldsymbol{\mu'}_k &= \left( \sum_{i\in \cal Q}  z_{i,k} + \alpha \right)  \boldsymbol{\mu}_k 
\end{align}
We then obtain the centroid update
\begin{align}
    \boldsymbol{\mu}_k = \frac{\sum_{i\in \cal Q}  z_{i,k} {\mathbf f}_i +  \alpha   \boldsymbol{\mu'}_k}{\sum_{i\in \cal Q}  z_{i,k} + \alpha} .
\end{align}
If we write 
\begin{equation}
    \beta_k = \frac{\sum_{i \in {\cal Q}} z_{i,k} }{\sum_{i \in {\cal Q}} z_{i,k} + \alpha }
\end{equation}
and
\begin{align}
    \boldsymbol{v}_k = \frac{\sum_{i \in {\cal Q}} z_{i,k}  {\mathbf f}_i}{ \sum_{i \in \cal Q} z_{i,k}}
\end{align}
we get the new centroid update
\begin{align}
 \boldsymbol{\mu}_k = \beta_k \boldsymbol{v}_k + (1 - \beta_k) \boldsymbol{\mu'}_k
\end{align}

\subsection{With respect to $\boldsymbol{\Sigma}_k^{-1}$}
\begin{align}
    \frac{\partial {\cal L}}{\boldsymbol{\Sigma}_k^{-1}} &=\frac{\partial }{\boldsymbol{\Sigma}_k^{-1}} \left( -  \sum_{i\in \cal Q}  {\mathbf z}_{i}^\top \log ({\mathbf p}_{i})
 -\sum_{i\in \cal Q} \sum_{j \in \cal Q} w_{ij} \mathbf{z}_{i}^\top \mathbf{z}_{j} + \sum_{i \in \cal Q} \mbox{KL} (\mathbf{z}_{i} || \hat{\mathbf{y}}_{i}) + \alpha \mathcal{A}(\boldsymbol{\mu}, \boldsymbol{\Sigma}) \right) \\ &= \frac{\partial }{\boldsymbol{\Sigma}_k^{-1}}\left(-  \sum_{i\in \cal Q}  {\mathbf z}_{i}^\top \log ({\mathbf p}_{i})
 -\sum_{i\in \cal Q} \sum_{j \in \cal Q} w_{ij} \mathbf{z}_{i}^\top \mathbf{z}_{j} + \sum_{i \in \cal Q} \mbox{KL} (\mathbf{z}_{i} || \hat{\mathbf{y}}_{i}) + \alpha \sum_{l = 1}^{K}  \mbox{KL} (\mathbf{q}_{l} || \mathbf{p}_{l}) \right) \\
 & = \frac{\partial }{\boldsymbol{\Sigma}_k^{-1}} \left(  -\sum_{i\in \cal Q}  {\mathbf z}_{i}^\top \left({-\frac{1}{2} \log |\boldsymbol{\Sigma}_k| - \frac {1}{2}}({\mathbf f}_i - \boldsymbol{\mu}_k)^\top \boldsymbol{\Sigma}_k^{-1}({\mathbf f}_i - \boldsymbol{\mu}_k) \right) +   \frac{\alpha}{2} (\log \frac{|\boldsymbol{\Sigma}_k|}{|\boldsymbol{\Sigma'}_k|} \notag \right.\\
 &  +  (\boldsymbol{\mu'}_k - \boldsymbol{\mu}_k)^\top \boldsymbol{\Sigma}_k^{-1}(\boldsymbol{\mu'}_k - \boldsymbol{\mu}_k) + \Tr(\boldsymbol{\Sigma}_k^{-1}\boldsymbol{\Sigma'}_k) \left.\vphantom{-\sum_{i\in \cal Q}})\right) 
\end{align}
Note that the term $\log \frac{|\boldsymbol{\Sigma}_k|}{|\boldsymbol{\Sigma'}_k|} \notag +  (\boldsymbol{\mu'}_k - \boldsymbol{\mu}_k)^\top \boldsymbol{\Sigma}_k^{-1}(\boldsymbol{\mu'}_k - \boldsymbol{\mu}_k) + \Tr(\boldsymbol{\Sigma}_k^{-1}\boldsymbol{\Sigma'}_k)$ directly comes from the derivative of our statistical anchor $\mathcal{A}(\boldsymbol{\mu}, \boldsymbol{\Sigma})$ with regard to $\boldsymbol{\Sigma}^{-1}_k$. Using the formulas (from Matrix cookbook \cite{petersen2008matrix})
\begin{align}
    \frac{\partial}{X} (\log |X|) &= (X^{-1})^\top \\
    \frac{\partial}{X^{-1}} (\log |X|) &= -X^\top
\end{align}
and
\begin{align}
    \frac{\partial}{X}(\Tr(AXB)) &= A^\top B^\top \\
    \frac{\partial}{X}(\Tr(AX^{-1}B )) &= -(X^{-1}BAX^{-1})^\top 
\end{align}
as well as the fact that covariances are symmetric ($\boldsymbol{\Sigma}^\top = \boldsymbol{\Sigma}$), setting the derivative to 0 yields
\begin{align}
    - \sum_{i\in \cal Q}  z_{i,k} \left(\boldsymbol{\Sigma}_k - ({\mathbf f}_i - \boldsymbol{\mu}_k)({\mathbf f}_i - \boldsymbol{\mu}_k)^\top \right) +   \alpha ( -\boldsymbol{\Sigma}_k^\top + (\boldsymbol{\mu'}_k - \boldsymbol{\mu}_k) (\boldsymbol{\mu'}_k - \boldsymbol{\mu}_k)^\top + \boldsymbol{\Sigma'}_k ^\top)  = 0 \\
    - \sum_{i\in \cal Q}  z_{i,k} \left(\boldsymbol{\Sigma}_k - ({\mathbf f}_i - \boldsymbol{\mu}_k)({\mathbf f}_i - \boldsymbol{\mu}_k)^\top \right) +   \alpha ( -\boldsymbol{\Sigma}_k + (\boldsymbol{\mu'}_k - \boldsymbol{\mu}_k) (\boldsymbol{\mu'}_k - \boldsymbol{\mu}_k)^\top + \boldsymbol{\Sigma'}_k )  = 0 \\
\end{align}
\begin{align}
    \left(\sum_{i\in \cal Q}   z_{i,k} + \alpha\right)\boldsymbol{\Sigma}_k = \sum_{i\in \cal Q}  z_{i,k} ({\mathbf f}_i - \boldsymbol{\mu}_k)({\mathbf f}_i - \boldsymbol{\mu}_k)^\top + \alpha (\boldsymbol{\Sigma'}_k + (\boldsymbol{\mu'}_k - \boldsymbol{\mu}_k) (\boldsymbol{\mu'}_k - \boldsymbol{\mu}_k)^\top).
\end{align}
We get 
\begin{align}
        \boldsymbol{\Sigma}_k = \frac{\sum_{i\in \cal Q}  z_{i,k} ({\mathbf f}_i - \boldsymbol{\mu}_k)({\mathbf f}_i - \boldsymbol{\mu}_k)^\top + \alpha (\boldsymbol{\Sigma'}_k + (\boldsymbol{\mu'}_k - \boldsymbol{\mu}_k) (\boldsymbol{\mu'}_k - \boldsymbol{\mu}_k)^\top)}{\sum_{i\in \cal Q}   z_{i,k} + \alpha}.
\end{align}
By writing the old $\boldsymbol{\Sigma}_k$-update
\begin{align}
    \boldsymbol{T}_k = \frac{\sum_{i\in \cal Q}  z_{i,k} ({\mathbf f}_i - \boldsymbol{\mu}_k)({\mathbf f}_i - \boldsymbol{\mu}_k)^\top}{\sum_{i\in \cal Q}  z_{i,k} },
\end{align}
we obtain the new covariance update 
\begin{align}
        \boldsymbol{\Sigma}_k = \beta_k \boldsymbol{T}_k + (1-\beta_k)(\boldsymbol{\Sigma'}_k + (\boldsymbol{\mu'}_k - \boldsymbol{\mu}_k) (\boldsymbol{\mu'}_k - \boldsymbol{\mu}_k)^\top).
\end{align}

\section{Detailed derivations of the complete formulation}
We refer to the derivations and the convergence proof in the TransCLIP paper \cite{zanella2024boosting}. The optimization follows a Block Majorize-Minimize (BMM) procedure over three blocks of variables: ${\mathbf z}$, $\boldsymbol{\mu}$, and $\boldsymbol{\Sigma}$. For the Majorize-Minimize (MM) with respect to the $\mathbf{z}$-block (while $\boldsymbol{\mu}$ and $\boldsymbol{\Sigma}$ are fixed), both the GMM- and KL-based terms are convex w.r.t ${\mathbf z}_i$. Consequently, we can proceed using similar arguments. For PSD matrix $\boldsymbol{W}$, the Laplacian regularization term in Eq. \eqref{original-transclip-zero-shot-objective} is concave. To address this, we can replace the quadratic Laplacian term by a linear bound. By introducing simplex constraints $\mathbf{z}_i \in \Delta_K$ ($\lambda_i$ the corresponding Lagrange multiplier) $i \in {\cal Q}$,
\begin{align}
    \frac{\partial {\cal L}}{\mathbf{z}_i}
    &=\frac{\partial }{\mathbf{z}_i} \left( -  \sum_{i\in \cal Q}  {\mathbf z}_{i}^\top \log ({\mathbf p}_{i})
 -\sum_{i\in \cal Q} \sum_{j \in \cal Q} w_{ij} \mathbf{z}_{i}^\top \mathbf{z}_{j} + \sum_{i \in \cal Q} \mbox{KL} (\mathbf{z}_{i} || \hat{\mathbf{y}}_{i}) + \alpha \mathcal{A}(\boldsymbol{\mu}, \boldsymbol{\Sigma}) + \sum_{i \in \cal Q} \lambda_i (z_i^\top \mathbbm{1}_K -1) \right) \\
 &= -\log ({\mathbf p}_{i}) - \sum_{j \in \cal Q} w_{ij} \mathbf{z}_{j} - \log(\hat{\mathbf{y}}_{i}) + \log(\mathbf{z}_{i}) + (1+\lambda_i)\mathbbm{1}_K .
\end{align}
Using the constraint
\begin{align}
    \mathbbm{1}_K^\top \mathbf{z}_i = 1 ,
\end{align}
we solve the Karush-Kuhn-Tucker (KKT) conditions independently for each $\mathbf{z}_i$ and finally obtain
\begin{align}
    \label{z-updates}
    \mathbf{z}_{i}^{(l+1)} = \frac{\hat{\mathbf{y}}_{i} \odot \exp (\log (\mathbf{p}_{i}) + \sum_{j \in {\cal Q}} w_{ij} \mathbf{z}_{j}^{(l)})}{(\hat{\mathbf{y}}_{i} \odot \exp (\log (\mathbf{p}_{i}) + \sum_{j\in {\cal Q}} w_{ij}\mathbf{z}_{j}^{(l)}))^\top\mathbbm{1}_K}.
\end{align}
Notice that the obtained $\mathbf{z}$-updates are decoupled, yielding computationally efficient transduction for large-scale datasets (see runtime in Table \ref{tab:runtime}).

\section{Text prompts}
We use the same text prompts for all our experiments. They are given in Table \ref{tab:app_all_prompts}.

\section{Implementation details for online test-time adaptation}
\label{supplemental:implem}
\begin{figure}[h]
\centering
\includegraphics[width=0.5\textwidth]{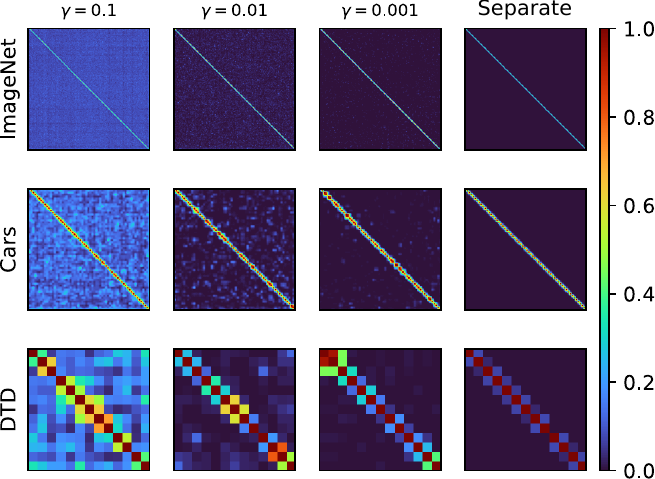}

\caption{Correlation matrix of per-batch $\ell_2$ normalized vectors of class proportions for batch size $128$. $x$ and $y$ axis of each plot is the batch index corresponding to the order in which the batches are processed. This illustrates the inter-batch correlation increasing as the Dirichlet parameter $\gamma$ decreases.}\label{fig:online_data_streams_correlation}
\end{figure}

For generating non i.i.d. data streams, we follow the setup of recent works \cite{yuan2023robust} and adopt a framework based on Dirichlet distributions. Namely, we distribute each class over a fixed number of slots according to proportions drawn following a Dirichlet distribution parametrized by a single scalar parameter $\gamma$. Therefore, for large values of $\gamma$ each class is evenly distributed among slots (i.i.d data stream) while for small values each class is distributed in a single slot (highly correlated data stream). Then, samples are randomly shuffled within each slot. For every dataset and a given batch size, the number of slots is $\min \{K, \floor*{\dfrac{|Q|}{\text{batch size}}} \}$. This is illustrated in Figure \ref{fig:online_data_streams_correlation}.
\begin{table}[]
\caption{Prompt templates used in all experiments.} 
\label{tab:app_all_prompts}
\centering
\resizebox{0.5\linewidth}{!}{
 \begin{tabular}{ccc}
    \toprule
    Dataset     &  Prompt template \\
    \midrule
   ImageNet     &   "\texttt{a photo of a [].}" \\
   SUN397     &   "\texttt{a photo of a [].}" \\
    Aircraft    &  "\texttt{a photo of a [], a type of aircraft.}",\\
    EuroSAT    & "\texttt{a centered satellite photo of [].}", \\
    Cars    &  "\texttt{a photo of a [].}", \\
    Food101    &  "\texttt{a photo of [], a type of food.}",\\
    Pets    &  "\texttt{a photo of [], a type of pet.}", \\
    Flower102    &  "\texttt{a photo of a [], a type of flower.}",\\
    Caltech101    &  "\texttt{a photo of a [].}", \\
    DTD    &  "\texttt{[] texture.}", \\
    UCF101    &  "\texttt{a photo of a person doing [].}",\\
    \bottomrule
    \end{tabular}}
\end{table}
\begin{table}[]
    \centering
        \caption{Additional information on the datasets.}
    \label{tab:appendix_datasets}
    \resizebox{0.6\linewidth}{!}{
    \begin{tabular}{ccccc}
    \toprule
        Dataset name & Other given name &  \# classes & \# test samples & task description \\
        \midrule
     SUN397 & Sun397 & 397 & 19,850 & scenes classification \\
          Aircraft & FGVCAircraft & 100 & 3,333 &  aircraft classification \\
          EuroSAT & EuroSAT & 10 & 8,100 & satellite images classification \\
        Cars & StanfordCars & 196 & 8,041 & cars classification \\
        Food101 & Food101 & 101 & 30,300 & food classification \\
        Pets & OxfordPets & 37 & 3,669 & pets classification \\
        Flowers102 & OxfordFlowers & 102 & 2,463 & flowers classification \\
        Caltech101 & Caltech101 & 101 & 2,465 & objects classification \\
        DTD & DTD & 47 & 1,692 & textures classification \\
        UCF101 & UCF101 & 101 & 3,783 & actions classification \\
        ImageNet & ImageNet & 1000 & 50,000 & objects classification \\
        \bottomrule
        
    \end{tabular}}
\end{table}

\section{Additional results}
We present results on four additional CLIP encoders: two convolutional neural networks (ResNet-50 and ResNet-101) and two transformer-based architectures (ViT-B/32 and ViT-L/14), aiming to demonstrate that the findings in the main paper generalize well to other model choices. For batch test-time adaptation (see Tables \ref{tab:appendix_small_batch}, \ref{tab:appendix_big_batch}, and \ref{tab:appendix_all}), we observe consistent improvements across various architectures and model sizes. Similarly, for online test-time adaptation (see Table \ref{tab:appendix_online}), the results show that the observed improvements remain consistent regardless of the architecture or model size.


\definecolor{AverageColor}{RGB}{255, 237, 206}
\definecolor{AverageDarkerColor}{RGB}{236, 227, 211}

\newcolumntype{n}{>{\columncolor{AverageColor}} p{1.7cm}}

\aboverulesep = 0.2mm 
\belowrulesep = 0.2mm 

\begin{table*}
\centering
\caption{Comparison of various CLIP encoders for the batch test-time adaptation setting with a batch size of 64. Each reported performance is averaged over 1,000 tasks.}
\label{tab:appendix_small_batch}
\begin{subtable}{\linewidth}
\centering
\caption{ResNet-50.}
\resizebox{\textwidth}{!}{%
\setlength\dashlinedash{0.2pt}
\setlength\dashlinegap{1.5pt}
\setlength\arrayrulewidth{0.3pt}
\renewcommand{\arraystretch}{1.}
    \begin{tabular}{p{1.7cm}l|nccccccccccc}
      \toprule
$K_{\text{eff}}$ & Method & \textsc{\fontsize{11}{10}\selectfont\textcolor{black}{Average}}& \rotatebox[origin=c]{45}{ImageNet} & \rotatebox[origin=c]{45}{SUN397} & \rotatebox[origin=c]{45}{Aircraft} & \rotatebox[origin=c]{45}{EuroSAT} & \rotatebox[origin=c]{45}{StanfordCars} & \rotatebox[origin=c]{45}{Food101} & \rotatebox[origin=c]{45}{Pets} & \rotatebox[origin=c]{45}{Flower102} & \rotatebox[origin=c]{45}{Caltech101} & \rotatebox[origin=c]{45}{DTD} & \rotatebox[origin=c]{45}{UCF101}  \\
\midrule
 & CLIP & $\textcolor{black}{58.7}$ & $\text{58.2}$ & $\text{58.9}$ & $\text{17.0}$ & $\text{36.2}$ & $\text{55.8}$ & $\text{77.4}$ & $\text{85.7}$ & $\text{66.1}$ & $\text{85.7}$ & $\text{42.8}$ & $\text{61.8}$  \\
\midrule
\multirow{1}{*}{\makecell{\texttt{Very Low}}} & \cellcolor{LightGray} Stat${\cal A}$ & \cellcolor{AverageDarkerColor} $  \textcolor{black}{\textbf{65.8}}_{\textcolor{darkerGreen}{\raisebox{2pt}{\normalsize+\textbf{7.1}}}}$ & \cellcolor{LightGray} $\text{68.2}_{\textcolor{darkerGreen}{\raisebox{2pt}{\normalsize+10.0}}}$ & \cellcolor{LightGray} $\text{63.7}_{\textcolor{darkerGreen}{\raisebox{2pt}{\normalsize+4.8}}}$ & \cellcolor{LightGray} $\text{21.1}_{\textcolor{darkerGreen}{\raisebox{2pt}{\normalsize+4.1}}}$ & \cellcolor{LightGray} $\text{43.3}_{\textcolor{darkerGreen}{\raisebox{2pt}{\normalsize+7.1}}}$ & \cellcolor{LightGray} $\text{71.3}_{\textcolor{darkerGreen}{\raisebox{2pt}{\normalsize+15.5}}}$ & \cellcolor{LightGray} $\text{87.1}_{\textcolor{darkerGreen}{\raisebox{2pt}{\normalsize+9.7}}}$ & \cellcolor{LightGray} $\text{93.1}_{\textcolor{darkerGreen}{\raisebox{2pt}{\normalsize+7.4}}}$ & \cellcolor{LightGray} $\text{74.1}_{\textcolor{darkerGreen}{\raisebox{2pt}{\normalsize+8.0}}}$ & \cellcolor{LightGray} $\text{90.1}_{\textcolor{darkerGreen}{\raisebox{2pt}{\normalsize+4.4}}}$ & \cellcolor{LightGray} $\text{45.3}_{\textcolor{darkerGreen}{\raisebox{2pt}{\normalsize+2.5}}}$ & \cellcolor{LightGray} $\text{66.2}_{\textcolor{darkerGreen}{\raisebox{2pt}{\normalsize+4.4}}}$  \\
\midrule
\multirow{1}{*}{\makecell{\texttt{Low}}} & \cellcolor{LightGray} Stat${\cal A}$ & \cellcolor{AverageDarkerColor} $  \textcolor{black}{\textbf{62.3}}_{\textcolor{darkerGreen}{\raisebox{2pt}{\normalsize+\textbf{3.7}}}}$ & \cellcolor{LightGray} $\text{65.0}_{\textcolor{darkerGreen}{\raisebox{2pt}{\normalsize+6.8}}}$ & \cellcolor{LightGray} $\text{62.8}_{\textcolor{darkerGreen}{\raisebox{2pt}{\normalsize+3.9}}}$ & \cellcolor{LightGray} $\text{17.8}_{\textcolor{darkerGreen}{\raisebox{2pt}{\normalsize+0.8}}}$ & \cellcolor{LightGray} $\text{31.7}_{\textcolor{red}{\raisebox{2pt}{\normalsize-4.5}}}$ & \cellcolor{LightGray} $\text{67.1}_{\textcolor{darkerGreen}{\raisebox{2pt}{\normalsize+11.3}}}$ & \cellcolor{LightGray} $\text{83.6}_{\textcolor{darkerGreen}{\raisebox{2pt}{\normalsize+6.2}}}$ & \cellcolor{LightGray} $\text{88.2}_{\textcolor{darkerGreen}{\raisebox{2pt}{\normalsize+2.5}}}$ & \cellcolor{LightGray} $\text{71.7}_{\textcolor{darkerGreen}{\raisebox{2pt}{\normalsize+5.6}}}$ & \cellcolor{LightGray} $\text{89.0}_{\textcolor{darkerGreen}{\raisebox{2pt}{\normalsize+3.3}}}$ & \cellcolor{LightGray} $\text{44.9}_{\textcolor{darkerGreen}{\raisebox{2pt}{\normalsize+2.1}}}$ & \cellcolor{LightGray} $\text{64.0}_{\textcolor{darkerGreen}{\raisebox{2pt}{\normalsize+2.2}}}$  \\
\midrule
\multirow{1}{*}{\makecell{\texttt{Medium}}} & \cellcolor{LightGray} Stat${\cal A}$ & \cellcolor{AverageDarkerColor} $  \textcolor{black}{\textbf{58.6}}_{\textcolor{red}{\raisebox{2pt}{\normalsize-\textbf{0.1}}}}$ & \cellcolor{LightGray} $\text{61.1}_{\textcolor{darkerGreen}{\raisebox{2pt}{\normalsize+2.9}}}$ & \cellcolor{LightGray} $\text{59.5}_{\textcolor{darkerGreen}{\raisebox{2pt}{\normalsize+0.6}}}$ & \cellcolor{LightGray} $\text{16.4}_{\textcolor{red}{\raisebox{2pt}{\normalsize-0.6}}}$ & \cellcolor{LightGray} $\text{27.3}_{\textcolor{red}{\raisebox{2pt}{\normalsize-8.9}}}$ & \cellcolor{LightGray} $\text{62.1}_{\textcolor{darkerGreen}{\raisebox{2pt}{\normalsize+6.3}}}$ & \cellcolor{LightGray} $\text{78.7}_{\textcolor{darkerGreen}{\raisebox{2pt}{\normalsize+1.3}}}$ & \cellcolor{LightGray} $\text{81.4}_{\textcolor{red}{\raisebox{2pt}{\normalsize-4.3}}}$ & \cellcolor{LightGray} $\text{64.1}_{\textcolor{red}{\raisebox{2pt}{\normalsize-2.0}}}$ & \cellcolor{LightGray} $\text{87.9}_{\textcolor{darkerGreen}{\raisebox{2pt}{\normalsize+2.2}}}$ & \cellcolor{LightGray} $\text{43.8}_{\textcolor{darkerGreen}{\raisebox{2pt}{\normalsize+1.0}}}$ & \cellcolor{LightGray} $\text{62.0}_{\textcolor{darkerGreen}{\raisebox{2pt}{\normalsize+0.2}}}$  \\
\midrule
\end{tabular}}
\end{subtable}
\\
\begin{subtable}{\linewidth}
\centering
\caption{ResNet-101.}
\resizebox{\textwidth}{!}{%
\setlength\dashlinedash{0.2pt}
\setlength\dashlinegap{1.5pt}
\setlength\arrayrulewidth{0.3pt}
\renewcommand{\arraystretch}{1.}
    \begin{tabular}{p{1.7cm}l|nccccccccccc}
      \toprule
$K_{\text{eff}}$ & Method & \textsc{\fontsize{11}{10}\selectfont\textcolor{black}{Average}}& \rotatebox[origin=c]{45}{ImageNet} & \rotatebox[origin=c]{45}{SUN397} & \rotatebox[origin=c]{45}{Aircraft} & \rotatebox[origin=c]{45}{EuroSAT} & \rotatebox[origin=c]{45}{StanfordCars} & \rotatebox[origin=c]{45}{Food101} & \rotatebox[origin=c]{45}{Pets} & \rotatebox[origin=c]{45}{Flower102} & \rotatebox[origin=c]{45}{Caltech101} & \rotatebox[origin=c]{45}{DTD} & \rotatebox[origin=c]{45}{UCF101}  \\
\midrule
 & CLIP & $\textcolor{black}{59.5}$ & $\text{61.3}$ & $\text{59.0}$ & $\text{17.9}$ & $\text{32.9}$ & $\text{63.2}$ & $\text{80.7}$ & $\text{86.9}$ & $\text{64.3}$ & $\text{89.9}$ & $\text{37.3}$ & $\text{61.1}$  \\
 
\midrule
\multirow{1}{*}{\makecell{\texttt{Very Low}}} & \cellcolor{LightGray} Stat${\cal A}$ & \cellcolor{AverageDarkerColor} $  \textcolor{black}{\textbf{66.2}}_{\textcolor{darkerGreen}{\raisebox{2pt}{\normalsize+\textbf{6.7}}}}$ & \cellcolor{LightGray} $\text{73.0}_{\textcolor{darkerGreen}{\raisebox{2pt}{\normalsize+11.7}}}$ & \cellcolor{LightGray} $\text{66.5}_{\textcolor{darkerGreen}{\raisebox{2pt}{\normalsize+7.5}}}$ & \cellcolor{LightGray} $\text{22.5}_{\textcolor{darkerGreen}{\raisebox{2pt}{\normalsize+4.6}}}$ & \cellcolor{LightGray} $\text{30.4}_{\textcolor{red}{\raisebox{2pt}{\normalsize-2.5}}}$ & \cellcolor{LightGray} $\text{76.2}_{\textcolor{darkerGreen}{\raisebox{2pt}{\normalsize+13.0}}}$ & \cellcolor{LightGray} $\text{89.5}_{\textcolor{darkerGreen}{\raisebox{2pt}{\normalsize+8.8}}}$ & \cellcolor{LightGray} $\text{95.2}_{\textcolor{darkerGreen}{\raisebox{2pt}{\normalsize+8.3}}}$ & \cellcolor{LightGray} $\text{74.6}_{\textcolor{darkerGreen}{\raisebox{2pt}{\normalsize+10.3}}}$ & \cellcolor{LightGray} $\text{91.9}_{\textcolor{darkerGreen}{\raisebox{2pt}{\normalsize+2.0}}}$ & \cellcolor{LightGray} $\text{42.9}_{\textcolor{darkerGreen}{\raisebox{2pt}{\normalsize+5.6}}}$ & \cellcolor{LightGray} $\text{65.1}_{\textcolor{darkerGreen}{\raisebox{2pt}{\normalsize+4.0}}}$  \\
\midrule
\multirow{1}{*}{\makecell{\texttt{Low}}} & \cellcolor{LightGray} Stat${\cal A}$ & \cellcolor{AverageDarkerColor} $  \textcolor{black}{\textbf{65.1}}_{\textcolor{darkerGreen}{\raisebox{2pt}{\normalsize+\textbf{5.6}}}}$ & \cellcolor{LightGray} $\text{71.2}_{\textcolor{darkerGreen}{\raisebox{2pt}{\normalsize+9.9}}}$ & \cellcolor{LightGray} $\text{65.9}_{\textcolor{darkerGreen}{\raisebox{2pt}{\normalsize+6.9}}}$ & \cellcolor{LightGray} $\text{20.0}_{\textcolor{darkerGreen}{\raisebox{2pt}{\normalsize+2.1}}}$ & \cellcolor{LightGray} $\text{29.6}_{\textcolor{red}{\raisebox{2pt}{\normalsize-3.3}}}$ & \cellcolor{LightGray} $\text{73.1}_{\textcolor{darkerGreen}{\raisebox{2pt}{\normalsize+9.9}}}$ & \cellcolor{LightGray} $\text{88.1}_{\textcolor{darkerGreen}{\raisebox{2pt}{\normalsize+7.4}}}$ & \cellcolor{LightGray} $\text{92.9}_{\textcolor{darkerGreen}{\raisebox{2pt}{\normalsize+6.0}}}$ & \cellcolor{LightGray} $\text{74.9}_{\textcolor{darkerGreen}{\raisebox{2pt}{\normalsize+10.6}}}$ & \cellcolor{LightGray} $\text{92.8}_{\textcolor{darkerGreen}{\raisebox{2pt}{\normalsize+2.9}}}$ & \cellcolor{LightGray} $\text{42.9}_{\textcolor{darkerGreen}{\raisebox{2pt}{\normalsize+5.6}}}$ & \cellcolor{LightGray} $\text{64.4}_{\textcolor{darkerGreen}{\raisebox{2pt}{\normalsize+3.3}}}$  \\
\midrule
\multirow{1}{*}{\makecell{\texttt{Medium}}} & \cellcolor{LightGray} Stat${\cal A}$ & \cellcolor{AverageDarkerColor} $  \textcolor{black}{\textbf{62.5}}_{\textcolor{darkerGreen}{\raisebox{2pt}{\normalsize+\textbf{3.0}}}}$ & \cellcolor{LightGray} $\text{67.0}_{\textcolor{darkerGreen}{\raisebox{2pt}{\normalsize+5.7}}}$ & \cellcolor{LightGray} $\text{62.7}_{\textcolor{darkerGreen}{\raisebox{2pt}{\normalsize+3.7}}}$ & \cellcolor{LightGray} $\text{18.6}_{\textcolor{darkerGreen}{\raisebox{2pt}{\normalsize+0.7}}}$ & \cellcolor{LightGray} $\text{28.7}_{\textcolor{red}{\raisebox{2pt}{\normalsize-4.2}}}$ & \cellcolor{LightGray} $\text{69.6}_{\textcolor{darkerGreen}{\raisebox{2pt}{\normalsize+6.4}}}$ & \cellcolor{LightGray} $\text{84.9}_{\textcolor{darkerGreen}{\raisebox{2pt}{\normalsize+4.2}}}$ & \cellcolor{LightGray} $\text{88.8}_{\textcolor{darkerGreen}{\raisebox{2pt}{\normalsize+1.9}}}$ & \cellcolor{LightGray} $\text{70.1}_{\textcolor{darkerGreen}{\raisebox{2pt}{\normalsize+5.8}}}$ & \cellcolor{LightGray} $\text{92.1}_{\textcolor{darkerGreen}{\raisebox{2pt}{\normalsize+2.2}}}$ & \cellcolor{LightGray} $\text{40.9}_{\textcolor{darkerGreen}{\raisebox{2pt}{\normalsize+3.6}}}$ & \cellcolor{LightGray} $\text{63.6}_{\textcolor{darkerGreen}{\raisebox{2pt}{\normalsize+2.5}}}$  \\
\midrule
\end{tabular}}
\end{subtable}
\\
\begin{subtable}{\linewidth}
\centering
\caption{ViT-B/32.}
\resizebox{\textwidth}{!}{%
\setlength\dashlinedash{0.2pt}
\setlength\dashlinegap{1.5pt}
\setlength\arrayrulewidth{0.3pt}
\renewcommand{\arraystretch}{1.}
    \begin{tabular}{p{1.7cm}l|nccccccccccc}
      \toprule
$K_{\text{eff}}$ & Method & \textsc{\fontsize{11}{10}\selectfont\textcolor{black}{Average}}& \rotatebox[origin=c]{45}{ImageNet} & \rotatebox[origin=c]{45}{SUN397} & \rotatebox[origin=c]{45}{Aircraft} & \rotatebox[origin=c]{45}{EuroSAT} & \rotatebox[origin=c]{45}{StanfordCars} & \rotatebox[origin=c]{45}{Food101} & \rotatebox[origin=c]{45}{Pets} & \rotatebox[origin=c]{45}{Flower102} & \rotatebox[origin=c]{45}{Caltech101} & \rotatebox[origin=c]{45}{DTD} & \rotatebox[origin=c]{45}{UCF101}  \\
\midrule
 & CLIP & $\textcolor{black}{61.9}$ & $\text{62.0}$ & $\text{62.1}$ & $\text{19.1}$ & $\text{45.4}$ & $\text{60.2}$ & $\text{80.4}$ & $\text{87.3}$ & $\text{66.6}$ & $\text{91.4}$ & $\text{42.7}$ & $\text{63.5}$  \\
 
\midrule
\multirow{1}{*}{\makecell{\texttt{Very Low}}} & \cellcolor{LightGray} Stat${\cal A}$ & \cellcolor{AverageDarkerColor} $  \textcolor{black}{\textbf{67.3}}_{\textcolor{darkerGreen}{\raisebox{2pt}{\normalsize+\textbf{5.4}}}}$ & \cellcolor{LightGray} $\text{68.1}_{\textcolor{darkerGreen}{\raisebox{2pt}{\normalsize+6.1}}}$ & \cellcolor{LightGray} $\text{65.6}_{\textcolor{darkerGreen}{\raisebox{2pt}{\normalsize+3.5}}}$ & \cellcolor{LightGray} $\text{23.0}_{\textcolor{darkerGreen}{\raisebox{2pt}{\normalsize+3.9}}}$ & \cellcolor{LightGray} $\text{53.2}_{\textcolor{darkerGreen}{\raisebox{2pt}{\normalsize+7.8}}}$ & \cellcolor{LightGray} $\text{71.9}_{\textcolor{darkerGreen}{\raisebox{2pt}{\normalsize+11.7}}}$ & \cellcolor{LightGray} $\text{85.8}_{\textcolor{darkerGreen}{\raisebox{2pt}{\normalsize+5.4}}}$ & \cellcolor{LightGray} $\text{94.3}_{\textcolor{darkerGreen}{\raisebox{2pt}{\normalsize+7.0}}}$ & \cellcolor{LightGray} $\text{74.9}_{\textcolor{darkerGreen}{\raisebox{2pt}{\normalsize+8.3}}}$ & \cellcolor{LightGray} $\text{93.4}_{\textcolor{darkerGreen}{\raisebox{2pt}{\normalsize+2.0}}}$ & \cellcolor{LightGray} $\text{45.2}_{\textcolor{darkerGreen}{\raisebox{2pt}{\normalsize+2.5}}}$ & \cellcolor{LightGray} $\text{64.5}_{\textcolor{darkerGreen}{\raisebox{2pt}{\normalsize+1.0}}}$  \\
\midrule
\multirow{1}{*}{\makecell{\texttt{Low}}} & \cellcolor{LightGray} Stat${\cal A}$ & \cellcolor{AverageDarkerColor} $  \textcolor{black}{\textbf{66.4}}_{\textcolor{darkerGreen}{\raisebox{2pt}{\normalsize+\textbf{4.5}}}}$ & \cellcolor{LightGray} $\text{67.2}_{\textcolor{darkerGreen}{\raisebox{2pt}{\normalsize+5.2}}}$ & \cellcolor{LightGray} $\text{65.7}_{\textcolor{darkerGreen}{\raisebox{2pt}{\normalsize+3.6}}}$ & \cellcolor{LightGray} $\text{21.9}_{\textcolor{darkerGreen}{\raisebox{2pt}{\normalsize+2.8}}}$ & \cellcolor{LightGray} $\text{50.1}_{\textcolor{darkerGreen}{\raisebox{2pt}{\normalsize+4.7}}}$ & \cellcolor{LightGray} $\text{69.3}_{\textcolor{darkerGreen}{\raisebox{2pt}{\normalsize+9.1}}}$ & \cellcolor{LightGray} $\text{84.5}_{\textcolor{darkerGreen}{\raisebox{2pt}{\normalsize+4.1}}}$ & \cellcolor{LightGray} $\text{92.5}_{\textcolor{darkerGreen}{\raisebox{2pt}{\normalsize+5.2}}}$ & \cellcolor{LightGray} $\text{75.3}_{\textcolor{darkerGreen}{\raisebox{2pt}{\normalsize+8.7}}}$ & \cellcolor{LightGray} $\text{93.2}_{\textcolor{darkerGreen}{\raisebox{2pt}{\normalsize+1.8}}}$ & \cellcolor{LightGray} $\text{46.1}_{\textcolor{darkerGreen}{\raisebox{2pt}{\normalsize+3.4}}}$ & \cellcolor{LightGray} $\text{64.6}_{\textcolor{darkerGreen}{\raisebox{2pt}{\normalsize+1.1}}}$  \\
\midrule
\multirow{1}{*}{\makecell{\texttt{Medium}}} & \cellcolor{LightGray} Stat${\cal A}$ & \cellcolor{AverageDarkerColor} $  \textcolor{black}{\textbf{64.3}}_{\textcolor{darkerGreen}{\raisebox{2pt}{\normalsize+\textbf{2.4}}}}$ & \cellcolor{LightGray} $\text{65.5}_{\textcolor{darkerGreen}{\raisebox{2pt}{\normalsize+3.5}}}$ & \cellcolor{LightGray} $\text{64.8}_{\textcolor{darkerGreen}{\raisebox{2pt}{\normalsize+2.7}}}$ & \cellcolor{LightGray} $\text{20.0}_{\textcolor{darkerGreen}{\raisebox{2pt}{\normalsize+0.9}}}$ & \cellcolor{LightGray} $\text{45.4}_{\textcolor{gray}{\raisebox{2pt}{\normalsize0.0}}}$ & \cellcolor{LightGray} $\text{65.0}_{\textcolor{darkerGreen}{\raisebox{2pt}{\normalsize+4.8}}}$ & \cellcolor{LightGray} $\text{82.6}_{\textcolor{darkerGreen}{\raisebox{2pt}{\normalsize+2.2}}}$ & \cellcolor{LightGray} $\text{89.4}_{\textcolor{darkerGreen}{\raisebox{2pt}{\normalsize+2.1}}}$ & \cellcolor{LightGray} $\text{70.6}_{\textcolor{darkerGreen}{\raisebox{2pt}{\normalsize+4.0}}}$ & \cellcolor{LightGray} $\text{92.9}_{\textcolor{darkerGreen}{\raisebox{2pt}{\normalsize+1.5}}}$ & \cellcolor{LightGray} $\text{46.7}_{\textcolor{darkerGreen}{\raisebox{2pt}{\normalsize+4.0}}}$ & \cellcolor{LightGray} $\text{64.4}_{\textcolor{darkerGreen}{\raisebox{2pt}{\normalsize+0.9}}}$  \\
\midrule
\end{tabular}}
\end{subtable}
\\
\begin{subtable}{\linewidth}
\centering
\caption{ViT-L/14.}
\resizebox{\textwidth}{!}{%
\setlength\dashlinedash{0.2pt}
\setlength\dashlinegap{1.5pt}
\setlength\arrayrulewidth{0.3pt}
\renewcommand{\arraystretch}{1.}
    \begin{tabular}{p{1.7cm}l|nccccccccccc}
      \toprule
$K_{\text{eff}}$ & Method & \textsc{\fontsize{11}{10}\selectfont\textcolor{black}{Average}}& \rotatebox[origin=c]{45}{ImageNet} & \rotatebox[origin=c]{45}{SUN397} & \rotatebox[origin=c]{45}{Aircraft} & \rotatebox[origin=c]{45}{EuroSAT} & \rotatebox[origin=c]{45}{StanfordCars} & \rotatebox[origin=c]{45}{Food101} & \rotatebox[origin=c]{45}{Pets} & \rotatebox[origin=c]{45}{Flower102} & \rotatebox[origin=c]{45}{Caltech101} & \rotatebox[origin=c]{45}{DTD} & \rotatebox[origin=c]{45}{UCF101}  \\
\midrule
 & CLIP & $\textcolor{black}{72.6}$ & $\text{73.5}$ & $\text{67.7}$ & $\text{32.5}$ & $\text{60.3}$ & $\text{76.9}$ & $\text{90.9}$ & $\text{93.5}$ & $\text{79.5}$ & $\text{95.2}$ & $\text{53.5}$ & $\text{74.9}$  \\
\midrule
\multirow{1}{*}{\makecell{\texttt{Very Low}}} & \cellcolor{LightGray} Stat${\cal A}$ & \cellcolor{AverageDarkerColor} $  \textcolor{black}{\textbf{77.3}}_{\textcolor{darkerGreen}{\raisebox{2pt}{\normalsize+\textbf{4.7}}}}$ & \cellcolor{LightGray} $\text{78.9}_{\textcolor{darkerGreen}{\raisebox{2pt}{\normalsize+5.4}}}$ & \cellcolor{LightGray} $\text{71.3}_{\textcolor{darkerGreen}{\raisebox{2pt}{\normalsize+3.6}}}$ & \cellcolor{LightGray} $\text{40.4}_{\textcolor{darkerGreen}{\raisebox{2pt}{\normalsize+7.9}}}$ & \cellcolor{LightGray} $\text{71.4}_{\textcolor{darkerGreen}{\raisebox{2pt}{\normalsize+11.1}}}$ & \cellcolor{LightGray} $\text{84.4}_{\textcolor{darkerGreen}{\raisebox{2pt}{\normalsize+7.5}}}$ & \cellcolor{LightGray} $\text{94.2}_{\textcolor{darkerGreen}{\raisebox{2pt}{\normalsize+3.3}}}$ & \cellcolor{LightGray} $\text{97.1}_{\textcolor{darkerGreen}{\raisebox{2pt}{\normalsize+3.6}}}$ & \cellcolor{LightGray} $\text{82.9}_{\textcolor{darkerGreen}{\raisebox{2pt}{\normalsize+3.4}}}$ & \cellcolor{LightGray} $\text{97.0}_{\textcolor{darkerGreen}{\raisebox{2pt}{\normalsize+1.8}}}$ & \cellcolor{LightGray} $\text{55.3}_{\textcolor{darkerGreen}{\raisebox{2pt}{\normalsize+1.8}}}$ & \cellcolor{LightGray} $\text{77.1}_{\textcolor{darkerGreen}{\raisebox{2pt}{\normalsize+2.2}}}$  \\
\midrule
\multirow{1}{*}{\makecell{\texttt{Low}}} & \cellcolor{LightGray} Stat${\cal A}$ & \cellcolor{AverageDarkerColor} $  \textcolor{black}{\textbf{76.1}}_{\textcolor{darkerGreen}{\raisebox{2pt}{\normalsize+\textbf{3.5}}}}$ & \cellcolor{LightGray} $\text{78.2}_{\textcolor{darkerGreen}{\raisebox{2pt}{\normalsize+4.7}}}$ & \cellcolor{LightGray} $\text{71.6}_{\textcolor{darkerGreen}{\raisebox{2pt}{\normalsize+3.9}}}$ & \cellcolor{LightGray} $\text{38.4}_{\textcolor{darkerGreen}{\raisebox{2pt}{\normalsize+5.9}}}$ & \cellcolor{LightGray} $\text{65.6}_{\textcolor{darkerGreen}{\raisebox{2pt}{\normalsize+5.3}}}$ & \cellcolor{LightGray} $\text{82.4}_{\textcolor{darkerGreen}{\raisebox{2pt}{\normalsize+5.5}}}$ & \cellcolor{LightGray} $\text{93.1}_{\textcolor{darkerGreen}{\raisebox{2pt}{\normalsize+2.2}}}$ & \cellcolor{LightGray} $\text{96.3}_{\textcolor{darkerGreen}{\raisebox{2pt}{\normalsize+2.8}}}$ & \cellcolor{LightGray} $\text{82.8}_{\textcolor{darkerGreen}{\raisebox{2pt}{\normalsize+3.3}}}$ & \cellcolor{LightGray} $\text{96.1}_{\textcolor{darkerGreen}{\raisebox{2pt}{\normalsize+0.9}}}$ & \cellcolor{LightGray} $\text{55.4}_{\textcolor{darkerGreen}{\raisebox{2pt}{\normalsize+1.9}}}$ & \cellcolor{LightGray} $\text{76.8}_{\textcolor{darkerGreen}{\raisebox{2pt}{\normalsize+1.9}}}$  \\
\midrule
\multirow{1}{*}{\makecell{\texttt{Medium}}} & \cellcolor{LightGray} Stat${\cal A}$ & \cellcolor{AverageDarkerColor} $  \textcolor{black}{\textbf{74.5}}_{\textcolor{darkerGreen}{\raisebox{2pt}{\normalsize+\textbf{2.0}}}}$ & \cellcolor{LightGray} $\text{76.6}_{\textcolor{darkerGreen}{\raisebox{2pt}{\normalsize+3.1}}}$ & \cellcolor{LightGray} $\text{70.0}_{\textcolor{darkerGreen}{\raisebox{2pt}{\normalsize+2.3}}}$ & \cellcolor{LightGray} $\text{36.4}_{\textcolor{darkerGreen}{\raisebox{2pt}{\normalsize+3.9}}}$ & \cellcolor{LightGray} $\text{62.6}_{\textcolor{darkerGreen}{\raisebox{2pt}{\normalsize+2.3}}}$ & \cellcolor{LightGray} $\text{80.6}_{\textcolor{darkerGreen}{\raisebox{2pt}{\normalsize+3.7}}}$ & \cellcolor{LightGray} $\text{92.1}_{\textcolor{darkerGreen}{\raisebox{2pt}{\normalsize+1.2}}}$ & \cellcolor{LightGray} $\text{93.9}_{\textcolor{darkerGreen}{\raisebox{2pt}{\normalsize+0.4}}}$ & \cellcolor{LightGray} $\text{80.8}_{\textcolor{darkerGreen}{\raisebox{2pt}{\normalsize+1.3}}}$ & \cellcolor{LightGray} $\text{95.6}_{\textcolor{darkerGreen}{\raisebox{2pt}{\normalsize+0.4}}}$ & \cellcolor{LightGray} $\text{54.6}_{\textcolor{darkerGreen}{\raisebox{2pt}{\normalsize+1.1}}}$ & \cellcolor{LightGray} $\text{77.1}_{\textcolor{darkerGreen}{\raisebox{2pt}{\normalsize+2.2}}}$  \\
\midrule
\end{tabular}}
\end{subtable}
\end{table*}


\definecolor{AverageColor}{RGB}{255, 237, 206}
\definecolor{AverageDarkerColor}{RGB}{236, 227, 211}

\newcolumntype{n}{>{\columncolor{AverageColor}} p{1.7cm}}

\aboverulesep = 0.2mm 
\belowrulesep = 0.2mm 

\begin{table*}
\centering
\caption{Comparison of various CLIP encoders for the batch test-time adaptation setting with a batch size of 1,000. Each reported performance is averaged over 1,000 tasks.}
\label{tab:appendix_big_batch}
\begin{subtable}{\linewidth}
\centering
\caption{ResNet-50.}
\resizebox{\textwidth}{!}{%
\setlength\dashlinedash{0.2pt}
\setlength\dashlinegap{1.5pt}
\setlength\arrayrulewidth{0.3pt}
\renewcommand{\arraystretch}{1.}
    \begin{tabular}{p{1.7cm}l|nccccccccccc}
      \toprule
$K_{\text{eff}}$ & Method & \textsc{\fontsize{11}{10}\selectfont\textcolor{black}{Average}}& \rotatebox[origin=c]{45}{ImageNet} & \rotatebox[origin=c]{45}{SUN397} & \rotatebox[origin=c]{45}{Aircraft} & \rotatebox[origin=c]{45}{EuroSAT} & \rotatebox[origin=c]{45}{StanfordCars} & \rotatebox[origin=c]{45}{Food101} & \rotatebox[origin=c]{45}{Pets} & \rotatebox[origin=c]{45}{Flower102} & \rotatebox[origin=c]{45}{Caltech101} & \rotatebox[origin=c]{45}{DTD} & \rotatebox[origin=c]{45}{UCF101}  \\
\midrule
 & CLIP & $\textcolor{black}{58.7}$ & $\text{58.2}$ & $\text{58.9}$ & $\text{17.0}$ & $\text{36.2}$ & $\text{55.8}$ & $\text{77.4}$ & $\text{85.7}$ & $\text{66.1}$ & $\text{85.7}$ & $\text{42.8}$ & $\text{61.8}$  \\
\midrule
\multirow{1}{*}{\makecell{\texttt{Medium}}} & \cellcolor{LightGray} Stat${\cal A}$ & \cellcolor{AverageDarkerColor} $  \textcolor{black}{\textbf{64.1}}_{\textcolor{darkerGreen}{\raisebox{2pt}{\normalsize+\textbf{5.4}}}}$ & \cellcolor{LightGray} $\text{65.2}_{\textcolor{darkerGreen}{\raisebox{2pt}{\normalsize+7.0}}}$ & \cellcolor{LightGray} $\text{61.5}_{\textcolor{darkerGreen}{\raisebox{2pt}{\normalsize+2.6}}}$ & \cellcolor{LightGray} $\text{18.6}_{\textcolor{darkerGreen}{\raisebox{2pt}{\normalsize+1.6}}}$ & \cellcolor{LightGray} $\text{51.2}_{\textcolor{darkerGreen}{\raisebox{2pt}{\normalsize+15.0}}}$ & \cellcolor{LightGray} $\text{67.2}_{\textcolor{darkerGreen}{\raisebox{2pt}{\normalsize+11.4}}}$ & \cellcolor{LightGray} $\text{80.9}_{\textcolor{darkerGreen}{\raisebox{2pt}{\normalsize+3.5}}}$ & \cellcolor{LightGray} $\text{89.1}_{\textcolor{darkerGreen}{\raisebox{2pt}{\normalsize+3.4}}}$ & \cellcolor{LightGray} $\text{70.7}_{\textcolor{darkerGreen}{\raisebox{2pt}{\normalsize+4.6}}}$ & \cellcolor{LightGray} $\text{88.5}_{\textcolor{darkerGreen}{\raisebox{2pt}{\normalsize+2.8}}}$ & \cellcolor{LightGray} $\text{46.8}_{\textcolor{darkerGreen}{\raisebox{2pt}{\normalsize+4.0}}}$ & \cellcolor{LightGray} $\text{65.3}_{\textcolor{darkerGreen}{\raisebox{2pt}{\normalsize+3.5}}}$  \\
\midrule
\multirow{1}{*}{\makecell{\texttt{High} }} & \cellcolor{LightGray} Stat${\cal A}$ & \cellcolor{AverageDarkerColor} $  \textcolor{black}{\textbf{63.5}}_{\textcolor{darkerGreen}{\raisebox{2pt}{\normalsize+\textbf{4.8}}}}$ & \cellcolor{LightGray} $\text{65.4}_{\textcolor{darkerGreen}{\raisebox{2pt}{\normalsize+7.2}}}$ & \cellcolor{LightGray} $\text{63.1}_{\textcolor{darkerGreen}{\raisebox{2pt}{\normalsize+4.2}}}$ & \cellcolor{LightGray} $\text{16.5}_{\textcolor{red}{\raisebox{2pt}{\normalsize-0.5}}}$ & \cellcolor{LightGray} $\text{51.7}_{\textcolor{darkerGreen}{\raisebox{2pt}{\normalsize+15.5}}}$ & \cellcolor{LightGray} $\text{65.4}_{\textcolor{darkerGreen}{\raisebox{2pt}{\normalsize+9.6}}}$ & \cellcolor{LightGray} $\text{81.0}_{\textcolor{darkerGreen}{\raisebox{2pt}{\normalsize+3.6}}}$ & \cellcolor{LightGray} $\text{84.4}_{\textcolor{red}{\raisebox{2pt}{\normalsize-1.3}}}$ & \cellcolor{LightGray} $\text{70.0}_{\textcolor{darkerGreen}{\raisebox{2pt}{\normalsize+3.9}}}$ & \cellcolor{LightGray} $\text{88.3}_{\textcolor{darkerGreen}{\raisebox{2pt}{\normalsize+2.6}}}$ & \cellcolor{LightGray} $\text{47.2}_{\textcolor{darkerGreen}{\raisebox{2pt}{\normalsize+4.4}}}$ & \cellcolor{LightGray} $\text{66.0}_{\textcolor{darkerGreen}{\raisebox{2pt}{\normalsize+4.2}}}$  \\
\midrule
\multirow{1}{*}{\makecell{\texttt{Very High}}} & \cellcolor{LightGray} Stat${\cal A}$ & \cellcolor{AverageDarkerColor} $  \textcolor{black}{\textbf{61.8}}_{\textcolor{darkerGreen}{\raisebox{2pt}{\normalsize+\textbf{3.1}}}}$ & \cellcolor{LightGray} $\text{63.5}_{\textcolor{darkerGreen}{\raisebox{2pt}{\normalsize+5.3}}}$ & \cellcolor{LightGray} $\text{62.4}_{\textcolor{darkerGreen}{\raisebox{2pt}{\normalsize+3.5}}}$ & \cellcolor{LightGray} $\text{14.8}_{\textcolor{red}{\raisebox{2pt}{\normalsize-2.2}}}$ & \cellcolor{LightGray} $\text{51.7}_{\textcolor{darkerGreen}{\raisebox{2pt}{\normalsize+15.5}}}$ & \cellcolor{LightGray} $\text{60.8}_{\textcolor{darkerGreen}{\raisebox{2pt}{\normalsize+5.0}}}$ & \cellcolor{LightGray} $\text{77.8}_{\textcolor{darkerGreen}{\raisebox{2pt}{\normalsize+0.4}}}$ & \cellcolor{LightGray} $\text{83.5}_{\textcolor{red}{\raisebox{2pt}{\normalsize-2.2}}}$ & \cellcolor{LightGray} $\text{66.2}_{\textcolor{darkerGreen}{\raisebox{2pt}{\normalsize+0.1}}}$ & \cellcolor{LightGray} $\text{87.9}_{\textcolor{darkerGreen}{\raisebox{2pt}{\normalsize+2.2}}}$ & \cellcolor{LightGray} $\text{46.6}_{\textcolor{darkerGreen}{\raisebox{2pt}{\normalsize+3.8}}}$ & \cellcolor{LightGray} $\text{64.5}_{\textcolor{darkerGreen}{\raisebox{2pt}{\normalsize+2.7}}}$  \\
\midrule
\end{tabular}}
\end{subtable}
\\
\begin{subtable}{\linewidth}
\centering
\caption{ResNet-101.}
\resizebox{\textwidth}{!}{%
\setlength\dashlinedash{0.2pt}
\setlength\dashlinegap{1.5pt}
\setlength\arrayrulewidth{0.3pt}
\renewcommand{\arraystretch}{1.}
    \begin{tabular}{p{1.7cm}l|nccccccccccc}
      \toprule
$K_{\text{eff}}$ & Method & \textsc{\fontsize{11}{10}\selectfont\textcolor{black}{Average}}& \rotatebox[origin=c]{45}{ImageNet} & \rotatebox[origin=c]{45}{SUN397} & \rotatebox[origin=c]{45}{Aircraft} & \rotatebox[origin=c]{45}{EuroSAT} & \rotatebox[origin=c]{45}{StanfordCars} & \rotatebox[origin=c]{45}{Food101} & \rotatebox[origin=c]{45}{Pets} & \rotatebox[origin=c]{45}{Flower102} & \rotatebox[origin=c]{45}{Caltech101} & \rotatebox[origin=c]{45}{DTD} & \rotatebox[origin=c]{45}{UCF101}  \\
\midrule
 & CLIP & $\textcolor{black}{59.5}$ & $\text{61.3}$ & $\text{59.0}$ & $\text{17.9}$ & $\text{32.9}$ & $\text{63.2}$ & $\text{80.7}$ & $\text{86.9}$ & $\text{64.3}$ & $\text{89.9}$ & $\text{37.3}$ & $\text{61.1}$  \\
\midrule
\multirow{1}{*}{\makecell{\texttt{Medium}}} & \cellcolor{LightGray} Stat${\cal A}$ & \cellcolor{AverageDarkerColor} $  \textcolor{black}{\textbf{65.0}}_{\textcolor{darkerGreen}{\raisebox{2pt}{\normalsize+\textbf{5.5}}}}$ & \cellcolor{LightGray} $\text{70.5}_{\textcolor{darkerGreen}{\raisebox{2pt}{\normalsize+9.2}}}$ & \cellcolor{LightGray} $\text{65.3}_{\textcolor{darkerGreen}{\raisebox{2pt}{\normalsize+6.3}}}$ & \cellcolor{LightGray} $\text{20.5}_{\textcolor{darkerGreen}{\raisebox{2pt}{\normalsize+2.6}}}$ & \cellcolor{LightGray} $\text{33.6}_{\textcolor{darkerGreen}{\raisebox{2pt}{\normalsize+0.7}}}$ & \cellcolor{LightGray} $\text{73.9}_{\textcolor{darkerGreen}{\raisebox{2pt}{\normalsize+10.7}}}$ & \cellcolor{LightGray} $\text{85.4}_{\textcolor{darkerGreen}{\raisebox{2pt}{\normalsize+4.7}}}$ & \cellcolor{LightGray} $\text{91.1}_{\textcolor{darkerGreen}{\raisebox{2pt}{\normalsize+4.2}}}$ & \cellcolor{LightGray} $\text{73.1}_{\textcolor{darkerGreen}{\raisebox{2pt}{\normalsize+8.8}}}$ & \cellcolor{LightGray} $\text{92.2}_{\textcolor{darkerGreen}{\raisebox{2pt}{\normalsize+2.3}}}$ & \cellcolor{LightGray} $\text{43.2}_{\textcolor{darkerGreen}{\raisebox{2pt}{\normalsize+5.9}}}$ & \cellcolor{LightGray} $\text{66.5}_{\textcolor{darkerGreen}{\raisebox{2pt}{\normalsize+5.4}}}$  \\
\midrule
\multirow{1}{*}{\makecell{\texttt{High} }} & \cellcolor{LightGray} Stat${\cal A}$ & \cellcolor{AverageDarkerColor} $  \textcolor{black}{\textbf{64.3}}_{\textcolor{darkerGreen}{\raisebox{2pt}{\normalsize+\textbf{4.8}}}}$ & \cellcolor{LightGray} $\text{71.4}_{\textcolor{darkerGreen}{\raisebox{2pt}{\normalsize+10.1}}}$ & \cellcolor{LightGray} $\text{66.2}_{\textcolor{darkerGreen}{\raisebox{2pt}{\normalsize+7.2}}}$ & \cellcolor{LightGray} $\text{18.6}_{\textcolor{darkerGreen}{\raisebox{2pt}{\normalsize+0.7}}}$ & \cellcolor{LightGray} $\text{32.8}_{\textcolor{red}{\raisebox{2pt}{\normalsize-0.1}}}$ & \cellcolor{LightGray} $\text{72.2}_{\textcolor{darkerGreen}{\raisebox{2pt}{\normalsize+9.0}}}$ & \cellcolor{LightGray} $\text{85.1}_{\textcolor{darkerGreen}{\raisebox{2pt}{\normalsize+4.4}}}$ & \cellcolor{LightGray} $\text{87.9}_{\textcolor{darkerGreen}{\raisebox{2pt}{\normalsize+1.0}}}$ & \cellcolor{LightGray} $\text{71.9}_{\textcolor{darkerGreen}{\raisebox{2pt}{\normalsize+7.6}}}$ & \cellcolor{LightGray} $\text{92.2}_{\textcolor{darkerGreen}{\raisebox{2pt}{\normalsize+2.3}}}$ & \cellcolor{LightGray} $\text{42.5}_{\textcolor{darkerGreen}{\raisebox{2pt}{\normalsize+5.2}}}$ & \cellcolor{LightGray} $\text{66.5}_{\textcolor{darkerGreen}{\raisebox{2pt}{\normalsize+5.4}}}$  \\
\midrule
\multirow{1}{*}{\makecell{\texttt{Very High}}} & \cellcolor{LightGray} Stat${\cal A}$ & \cellcolor{AverageDarkerColor} $  \textcolor{black}{\textbf{62.6}}_{\textcolor{darkerGreen}{\raisebox{2pt}{\normalsize+\textbf{3.1}}}}$ & \cellcolor{LightGray} $\text{70.1}_{\textcolor{darkerGreen}{\raisebox{2pt}{\normalsize+8.8}}}$ & \cellcolor{LightGray} $\text{65.4}_{\textcolor{darkerGreen}{\raisebox{2pt}{\normalsize+6.4}}}$ & \cellcolor{LightGray} $\text{16.9}_{\textcolor{red}{\raisebox{2pt}{\normalsize-1.0}}}$ & \cellcolor{LightGray} $\text{32.9}_{\textcolor{gray}{\raisebox{2pt}{\normalsize0.0}}}$ & \cellcolor{LightGray} $\text{68.2}_{\textcolor{darkerGreen}{\raisebox{2pt}{\normalsize+5.0}}}$ & \cellcolor{LightGray} $\text{82.4}_{\textcolor{darkerGreen}{\raisebox{2pt}{\normalsize+1.7}}}$ & \cellcolor{LightGray} $\text{87.2}_{\textcolor{darkerGreen}{\raisebox{2pt}{\normalsize+0.3}}}$ & \cellcolor{LightGray} $\text{68.7}_{\textcolor{darkerGreen}{\raisebox{2pt}{\normalsize+4.4}}}$ & \cellcolor{LightGray} $\text{91.3}_{\textcolor{darkerGreen}{\raisebox{2pt}{\normalsize+1.4}}}$ & \cellcolor{LightGray} $\text{41.9}_{\textcolor{darkerGreen}{\raisebox{2pt}{\normalsize+4.6}}}$ & \cellcolor{LightGray} $\text{63.8}_{\textcolor{darkerGreen}{\raisebox{2pt}{\normalsize+2.7}}}$  \\
\midrule
\end{tabular}}
\end{subtable}
\\
\begin{subtable}{\linewidth}
\centering
\caption{ViT-B/32.}
\resizebox{\textwidth}{!}{%
\setlength\dashlinedash{0.2pt}
\setlength\dashlinegap{1.5pt}
\setlength\arrayrulewidth{0.3pt}
\renewcommand{\arraystretch}{1.}
    \begin{tabular}{p{1.7cm}l|nccccccccccc}
      \toprule
$K_{\text{eff}}$ & Method & \textsc{\fontsize{11}{10}\selectfont\textcolor{black}{Average}}& \rotatebox[origin=c]{45}{ImageNet} & \rotatebox[origin=c]{45}{SUN397} & \rotatebox[origin=c]{45}{Aircraft} & \rotatebox[origin=c]{45}{EuroSAT} & \rotatebox[origin=c]{45}{StanfordCars} & \rotatebox[origin=c]{45}{Food101} & \rotatebox[origin=c]{45}{Pets} & \rotatebox[origin=c]{45}{Flower102} & \rotatebox[origin=c]{45}{Caltech101} & \rotatebox[origin=c]{45}{DTD} & \rotatebox[origin=c]{45}{UCF101}  \\
\midrule
 & CLIP & $\textcolor{black}{61.9}$ & $\text{62.0}$ & $\text{62.1}$ & $\text{19.1}$ & $\text{45.4}$ & $\text{60.2}$ & $\text{80.4}$ & $\text{87.3}$ & $\text{66.6}$ & $\text{91.4}$ & $\text{42.7}$ & $\text{63.5}$  \\
\midrule
\multirow{1}{*}{\makecell{\texttt{Medium}}} & \cellcolor{LightGray} Stat${\cal A}$ & \cellcolor{AverageDarkerColor} $  \textcolor{black}{\textbf{65.9}}_{\textcolor{darkerGreen}{\raisebox{2pt}{\normalsize+\textbf{4.0}}}}$ & \cellcolor{LightGray} $\text{65.9}_{\textcolor{darkerGreen}{\raisebox{2pt}{\normalsize+3.9}}}$ & \cellcolor{LightGray} $\text{63.3}_{\textcolor{darkerGreen}{\raisebox{2pt}{\normalsize+1.2}}}$ & \cellcolor{LightGray} $\text{21.9}_{\textcolor{darkerGreen}{\raisebox{2pt}{\normalsize+2.8}}}$ & \cellcolor{LightGray} $\text{51.3}_{\textcolor{darkerGreen}{\raisebox{2pt}{\normalsize+5.9}}}$ & \cellcolor{LightGray} $\text{69.3}_{\textcolor{darkerGreen}{\raisebox{2pt}{\normalsize+9.1}}}$ & \cellcolor{LightGray} $\text{82.2}_{\textcolor{darkerGreen}{\raisebox{2pt}{\normalsize+1.8}}}$ & \cellcolor{LightGray} $\text{90.3}_{\textcolor{darkerGreen}{\raisebox{2pt}{\normalsize+3.0}}}$ & \cellcolor{LightGray} $\text{74.1}_{\textcolor{darkerGreen}{\raisebox{2pt}{\normalsize+7.5}}}$ & \cellcolor{LightGray} $\text{92.6}_{\textcolor{darkerGreen}{\raisebox{2pt}{\normalsize+1.2}}}$ & \cellcolor{LightGray} $\text{47.4}_{\textcolor{darkerGreen}{\raisebox{2pt}{\normalsize+4.7}}}$ & \cellcolor{LightGray} $\text{66.1}_{\textcolor{darkerGreen}{\raisebox{2pt}{\normalsize+2.6}}}$  \\
\midrule
\multirow{1}{*}{\makecell{\texttt{High} }} & \cellcolor{LightGray} Stat${\cal A}$ & \cellcolor{AverageDarkerColor} $  \textcolor{black}{\textbf{66.0}}_{\textcolor{darkerGreen}{\raisebox{2pt}{\normalsize+\textbf{4.1}}}}$ & \cellcolor{LightGray} $\text{67.0}_{\textcolor{darkerGreen}{\raisebox{2pt}{\normalsize+5.0}}}$ & \cellcolor{LightGray} $\text{65.0}_{\textcolor{darkerGreen}{\raisebox{2pt}{\normalsize+2.9}}}$ & \cellcolor{LightGray} $\text{20.2}_{\textcolor{darkerGreen}{\raisebox{2pt}{\normalsize+1.1}}}$ & \cellcolor{LightGray} $\text{51.1}_{\textcolor{darkerGreen}{\raisebox{2pt}{\normalsize+5.7}}}$ & \cellcolor{LightGray} $\text{68.5}_{\textcolor{darkerGreen}{\raisebox{2pt}{\normalsize+8.3}}}$ & \cellcolor{LightGray} $\text{82.7}_{\textcolor{darkerGreen}{\raisebox{2pt}{\normalsize+2.3}}}$ & \cellcolor{LightGray} $\text{88.5}_{\textcolor{darkerGreen}{\raisebox{2pt}{\normalsize+1.2}}}$ & \cellcolor{LightGray} $\text{73.7}_{\textcolor{darkerGreen}{\raisebox{2pt}{\normalsize+7.1}}}$ & \cellcolor{LightGray} $\text{92.5}_{\textcolor{darkerGreen}{\raisebox{2pt}{\normalsize+1.1}}}$ & \cellcolor{LightGray} $\text{49.5}_{\textcolor{darkerGreen}{\raisebox{2pt}{\normalsize+6.8}}}$ & \cellcolor{LightGray} $\text{66.9}_{\textcolor{darkerGreen}{\raisebox{2pt}{\normalsize+3.4}}}$  \\
\midrule
\multirow{1}{*}{\makecell{\texttt{Very High}}} & \cellcolor{LightGray} Stat${\cal A}$ & \cellcolor{AverageDarkerColor} $  \textcolor{black}{\textbf{65.1}}_{\textcolor{darkerGreen}{\raisebox{2pt}{\normalsize+\textbf{3.2}}}}$ & \cellcolor{LightGray} $\text{66.6}_{\textcolor{darkerGreen}{\raisebox{2pt}{\normalsize+4.6}}}$ & \cellcolor{LightGray} $\text{66.0}_{\textcolor{darkerGreen}{\raisebox{2pt}{\normalsize+3.9}}}$ & \cellcolor{LightGray} $\text{18.8}_{\textcolor{red}{\raisebox{2pt}{\normalsize-0.3}}}$ & \cellcolor{LightGray} $\text{51.0}_{\textcolor{darkerGreen}{\raisebox{2pt}{\normalsize+5.6}}}$ & \cellcolor{LightGray} $\text{65.1}_{\textcolor{darkerGreen}{\raisebox{2pt}{\normalsize+4.9}}}$ & \cellcolor{LightGray} $\text{81.5}_{\textcolor{darkerGreen}{\raisebox{2pt}{\normalsize+1.1}}}$ & \cellcolor{LightGray} $\text{88.0}_{\textcolor{darkerGreen}{\raisebox{2pt}{\normalsize+0.7}}}$ & \cellcolor{LightGray} $\text{70.6}_{\textcolor{darkerGreen}{\raisebox{2pt}{\normalsize+4.0}}}$ & \cellcolor{LightGray} $\text{91.9}_{\textcolor{darkerGreen}{\raisebox{2pt}{\normalsize+0.5}}}$ & \cellcolor{LightGray} $\text{49.5}_{\textcolor{darkerGreen}{\raisebox{2pt}{\normalsize+6.8}}}$ & \cellcolor{LightGray} $\text{66.5}_{\textcolor{darkerGreen}{\raisebox{2pt}{\normalsize+3.0}}}$  \\
\midrule
\end{tabular}}
\end{subtable}
\\
\begin{subtable}{\linewidth}
\centering
\caption{ViT-L/14.}
\resizebox{\textwidth}{!}{%
\setlength\dashlinedash{0.2pt}
\setlength\dashlinegap{1.5pt}
\setlength\arrayrulewidth{0.3pt}
\renewcommand{\arraystretch}{1.}
    \begin{tabular}{p{1.7cm}l|nccccccccccc}
      \toprule
$K_{\text{eff}}$ & Method & \textsc{\fontsize{11}{10}\selectfont\textcolor{black}{Average}}& \rotatebox[origin=c]{45}{ImageNet} & \rotatebox[origin=c]{45}{SUN397} & \rotatebox[origin=c]{45}{Aircraft} & \rotatebox[origin=c]{45}{EuroSAT} & \rotatebox[origin=c]{45}{StanfordCars} & \rotatebox[origin=c]{45}{Food101} & \rotatebox[origin=c]{45}{Pets} & \rotatebox[origin=c]{45}{Flower102} & \rotatebox[origin=c]{45}{Caltech101} & \rotatebox[origin=c]{45}{DTD} & \rotatebox[origin=c]{45}{UCF101}  \\
\midrule
 & CLIP & $\textcolor{black}{72.6}$ & $\text{73.5}$ & $\text{67.7}$ & $\text{32.5}$ & $\text{60.3}$ & $\text{76.9}$ & $\text{90.9}$ & $\text{93.5}$ & $\text{79.5}$ & $\text{95.2}$ & $\text{53.5}$ & $\text{74.9}$  \\
\midrule
\multirow{1}{*}{\makecell{\texttt{Medium}}} & \cellcolor{LightGray} Stat${\cal A}$ & \cellcolor{AverageDarkerColor} $  \textcolor{black}{\textbf{76.0}}_{\textcolor{darkerGreen}{\raisebox{2pt}{\normalsize+\textbf{3.4}}}}$ & \cellcolor{LightGray} $\text{76.2}_{\textcolor{darkerGreen}{\raisebox{2pt}{\normalsize+2.7}}}$ & \cellcolor{LightGray} $\text{69.4}_{\textcolor{darkerGreen}{\raisebox{2pt}{\normalsize+1.7}}}$ & \cellcolor{LightGray} $\text{39.1}_{\textcolor{darkerGreen}{\raisebox{2pt}{\normalsize+6.6}}}$ & \cellcolor{LightGray} $\text{71.0}_{\textcolor{darkerGreen}{\raisebox{2pt}{\normalsize+10.7}}}$ & \cellcolor{LightGray} $\text{81.9}_{\textcolor{darkerGreen}{\raisebox{2pt}{\normalsize+5.0}}}$ & \cellcolor{LightGray} $\text{91.7}_{\textcolor{darkerGreen}{\raisebox{2pt}{\normalsize+0.8}}}$ & \cellcolor{LightGray} $\text{94.8}_{\textcolor{darkerGreen}{\raisebox{2pt}{\normalsize+1.3}}}$ & \cellcolor{LightGray} $\text{81.9}_{\textcolor{darkerGreen}{\raisebox{2pt}{\normalsize+2.4}}}$ & \cellcolor{LightGray} $\text{95.6}_{\textcolor{darkerGreen}{\raisebox{2pt}{\normalsize+0.4}}}$ & \cellcolor{LightGray} $\text{56.9}_{\textcolor{darkerGreen}{\raisebox{2pt}{\normalsize+3.4}}}$ & \cellcolor{LightGray} $\text{77.6}_{\textcolor{darkerGreen}{\raisebox{2pt}{\normalsize+2.7}}}$  \\
\midrule
\multirow{1}{*}{\makecell{\texttt{High} }} & \cellcolor{LightGray} Stat${\cal A}$ & \cellcolor{AverageDarkerColor} $  \textcolor{black}{\textbf{76.3}}_{\textcolor{darkerGreen}{\raisebox{2pt}{\normalsize+\textbf{3.7}}}}$ & \cellcolor{LightGray} $\text{77.2}_{\textcolor{darkerGreen}{\raisebox{2pt}{\normalsize+3.7}}}$ & \cellcolor{LightGray} $\text{70.9}_{\textcolor{darkerGreen}{\raisebox{2pt}{\normalsize+3.2}}}$ & \cellcolor{LightGray} $\text{36.8}_{\textcolor{darkerGreen}{\raisebox{2pt}{\normalsize+4.3}}}$ & \cellcolor{LightGray} $\text{71.2}_{\textcolor{darkerGreen}{\raisebox{2pt}{\normalsize+10.9}}}$ & \cellcolor{LightGray} $\text{82.0}_{\textcolor{darkerGreen}{\raisebox{2pt}{\normalsize+5.1}}}$ & \cellcolor{LightGray} $\text{92.3}_{\textcolor{darkerGreen}{\raisebox{2pt}{\normalsize+1.4}}}$ & \cellcolor{LightGray} $\text{94.3}_{\textcolor{darkerGreen}{\raisebox{2pt}{\normalsize+0.8}}}$ & \cellcolor{LightGray} $\text{81.9}_{\textcolor{darkerGreen}{\raisebox{2pt}{\normalsize+2.4}}}$ & \cellcolor{LightGray} $\text{95.3}_{\textcolor{darkerGreen}{\raisebox{2pt}{\normalsize+0.1}}}$ & \cellcolor{LightGray} $\text{58.7}_{\textcolor{darkerGreen}{\raisebox{2pt}{\normalsize+5.2}}}$ & \cellcolor{LightGray} $\text{78.8}_{\textcolor{darkerGreen}{\raisebox{2pt}{\normalsize+3.9}}}$  \\
\midrule
\multirow{1}{*}{\makecell{\texttt{Very High}}} & \cellcolor{LightGray} Stat${\cal A}$ & \cellcolor{AverageDarkerColor} $  \textcolor{black}{\textbf{75.7}}_{\textcolor{darkerGreen}{\raisebox{2pt}{\normalsize+\textbf{3.1}}}}$ & \cellcolor{LightGray} $\text{77.3}_{\textcolor{darkerGreen}{\raisebox{2pt}{\normalsize+3.8}}}$ & \cellcolor{LightGray} $\text{71.6}_{\textcolor{darkerGreen}{\raisebox{2pt}{\normalsize+3.9}}}$ & \cellcolor{LightGray} $\text{33.7}_{\textcolor{darkerGreen}{\raisebox{2pt}{\normalsize+1.2}}}$ & \cellcolor{LightGray} $\text{71.2}_{\textcolor{darkerGreen}{\raisebox{2pt}{\normalsize+10.9}}}$ & \cellcolor{LightGray} $\text{79.5}_{\textcolor{darkerGreen}{\raisebox{2pt}{\normalsize+2.6}}}$ & \cellcolor{LightGray} $\text{91.7}_{\textcolor{darkerGreen}{\raisebox{2pt}{\normalsize+0.8}}}$ & \cellcolor{LightGray} $\text{94.1}_{\textcolor{darkerGreen}{\raisebox{2pt}{\normalsize+0.6}}}$ & \cellcolor{LightGray} $\text{80.7}_{\textcolor{darkerGreen}{\raisebox{2pt}{\normalsize+1.2}}}$ & \cellcolor{LightGray} $\text{94.9}_{\textcolor{red}{\raisebox{2pt}{\normalsize-0.3}}}$ & \cellcolor{LightGray} $\text{59.0}_{\textcolor{darkerGreen}{\raisebox{2pt}{\normalsize+5.5}}}$ & \cellcolor{LightGray} $\text{78.7}_{\textcolor{darkerGreen}{\raisebox{2pt}{\normalsize+3.8}}}$  \\
\midrule
\end{tabular}}
\end{subtable}
\end{table*}




\definecolor{AverageColor}{RGB}{255, 237, 206}
\definecolor{AverageDarkerColor}{RGB}{236, 227, 211}

\newcolumntype{n}{>{\columncolor{AverageColor}} p{1.7cm}}

\aboverulesep = 0.2mm 
\belowrulesep = 0.2mm 

\begin{table*}
\centering
\caption{Comparison of various CLIP encoders for the batch test-time adaptation setting on whole datasets. Each reported performance is averaged over 1,000 tasks.}
\label{tab:appendix_all}
\begin{subtable}{\linewidth}
\centering
\caption{ResNet-50.}
\resizebox{\textwidth}{!}{%
\setlength\dashlinedash{0.2pt}
\setlength\dashlinegap{1.5pt}
\setlength\arrayrulewidth{0.3pt}
\renewcommand{\arraystretch}{1.}
    \begin{tabular}{p{1.7cm}l|nccccccccccc}
      \toprule
$K_{\text{eff}}$ & Method & \textsc{\fontsize{11}{10}\selectfont\textcolor{black}{Average}}& \rotatebox[origin=c]{45}{ImageNet} & \rotatebox[origin=c]{45}{SUN397} & \rotatebox[origin=c]{45}{Aircraft} & \rotatebox[origin=c]{45}{EuroSAT} & \rotatebox[origin=c]{45}{StanfordCars} & \rotatebox[origin=c]{45}{Food101} & \rotatebox[origin=c]{45}{Pets} & \rotatebox[origin=c]{45}{Flower102} & \rotatebox[origin=c]{45}{Caltech101} & \rotatebox[origin=c]{45}{DTD} & \rotatebox[origin=c]{45}{UCF101}  \\
\midrule
 & CLIP & $\textcolor{black}{58.7}$ & $\text{58.2}$ & $\text{58.9}$ & $\text{17.0}$ & $\text{36.2}$ & $\text{55.8}$ & $\text{77.4}$ & $\text{85.7}$ & $\text{66.1}$ & $\text{85.7}$ & $\text{42.8}$ & $\text{61.8}$  \\
 
\midrule

 \texttt{All} & \cellcolor{LightGray} Stat${\cal A}$ & \cellcolor{AverageDarkerColor} $\textcolor{black}{\textbf{62.4}}_{\textcolor{darkerGreen}{\raisebox{2pt}{\normalsize+\textbf{3.7}}}}$ & \cellcolor{LightGray} $\text{60.4}_{\textcolor{darkerGreen}{\raisebox{2pt}{\normalsize+2.2}}}$ & \cellcolor{LightGray} $\text{64.3}_{\textcolor{darkerGreen}{\raisebox{2pt}{\normalsize+5.4}}}$ & \cellcolor{LightGray} $\text{16.0}_{\textcolor{red}{\raisebox{2pt}{\normalsize-1.0}}}$ & \cellcolor{LightGray} $\text{50.5}_{\textcolor{darkerGreen}{\raisebox{2pt}{\normalsize+14.3}}}$ & \cellcolor{LightGray} $\text{58.2}_{\textcolor{darkerGreen}{\raisebox{2pt}{\normalsize+2.4}}}$ & \cellcolor{LightGray} $\text{77.9}_{\textcolor{darkerGreen}{\raisebox{2pt}{\normalsize+0.5}}}$ & \cellcolor{LightGray} $\text{87.7}_{\textcolor{darkerGreen}{\raisebox{2pt}{\normalsize+2.0}}}$ & \cellcolor{LightGray} $\text{67.7}_{\textcolor{darkerGreen}{\raisebox{2pt}{\normalsize+1.6}}}$ & \cellcolor{LightGray} $\text{87.3}_{\textcolor{darkerGreen}{\raisebox{2pt}{\normalsize+1.6}}}$ & \cellcolor{LightGray} $\text{48.5}_{\textcolor{darkerGreen}{\raisebox{2pt}{\normalsize+5.7}}}$ & \cellcolor{LightGray} $\text{67.5}_{\textcolor{darkerGreen}{\raisebox{2pt}{\normalsize+5.7}}}$  \\
\midrule
\end{tabular}}
\end{subtable}
\\
\begin{subtable}{\linewidth}
\centering
\caption{ResNet-101.}
\resizebox{\textwidth}{!}{%
\setlength\dashlinedash{0.2pt}
\setlength\dashlinegap{1.5pt}
\setlength\arrayrulewidth{0.3pt}
\renewcommand{\arraystretch}{1.}
    \begin{tabular}{p{1.7cm}l|nccccccccccc}
      \toprule
$K_{\text{eff}}$ & Method & \textsc{\fontsize{11}{10}\selectfont\textcolor{black}{Average}}& \rotatebox[origin=c]{45}{ImageNet} & \rotatebox[origin=c]{45}{SUN397} & \rotatebox[origin=c]{45}{Aircraft} & \rotatebox[origin=c]{45}{EuroSAT} & \rotatebox[origin=c]{45}{StanfordCars} & \rotatebox[origin=c]{45}{Food101} & \rotatebox[origin=c]{45}{Pets} & \rotatebox[origin=c]{45}{Flower102} & \rotatebox[origin=c]{45}{Caltech101} & \rotatebox[origin=c]{45}{DTD} & \rotatebox[origin=c]{45}{UCF101}  \\
\midrule
 & CLIP & $\textcolor{black}{59.5}$ & $\text{61.3}$ & $\text{59.0}$ & $\text{17.9}$ & $\text{32.9}$ & $\text{63.2}$ & $\text{80.7}$ & $\text{86.9}$ & $\text{64.3}$ & $\text{89.9}$ & $\text{37.3}$ & $\text{61.1}$  \\
 
\midrule

  \texttt{All} & \cellcolor{LightGray} Stat${\cal A}$ & \cellcolor{AverageDarkerColor} $\textcolor{black}{\textbf{63.6}}_{\textcolor{darkerGreen}{\raisebox{2pt}{\normalsize+\textbf{4.1}}}}$ & \cellcolor{LightGray} $\text{64.4}_{\textcolor{darkerGreen}{\raisebox{2pt}{\normalsize+3.1}}}$ & \cellcolor{LightGray} $\text{64.9}_{\textcolor{darkerGreen}{\raisebox{2pt}{\normalsize+5.9}}}$ & \cellcolor{LightGray} $\text{18.3}_{\textcolor{darkerGreen}{\raisebox{2pt}{\normalsize+0.4}}}$ & \cellcolor{LightGray} $\text{43.3}_{\textcolor{darkerGreen}{\raisebox{2pt}{\normalsize+10.4}}}$ & \cellcolor{LightGray} $\text{66.2}_{\textcolor{darkerGreen}{\raisebox{2pt}{\normalsize+3.0}}}$ & \cellcolor{LightGray} $\text{81.9}_{\textcolor{darkerGreen}{\raisebox{2pt}{\normalsize+1.2}}}$ & \cellcolor{LightGray} $\text{88.8}_{\textcolor{darkerGreen}{\raisebox{2pt}{\normalsize+1.9}}}$ & \cellcolor{LightGray} $\text{69.1}_{\textcolor{darkerGreen}{\raisebox{2pt}{\normalsize+4.8}}}$ & \cellcolor{LightGray} $\text{91.5}_{\textcolor{darkerGreen}{\raisebox{2pt}{\normalsize+1.6}}}$ & \cellcolor{LightGray} $\text{42.9}_{\textcolor{darkerGreen}{\raisebox{2pt}{\normalsize+5.6}}}$ & \cellcolor{LightGray} $\text{67.8}_{\textcolor{darkerGreen}{\raisebox{2pt}{\normalsize+6.7}}}$  \\
\midrule
\end{tabular}}
\end{subtable}
\\
\begin{subtable}{\linewidth}
\centering
\caption{ViT-B/32.}
\resizebox{\textwidth}{!}{%
\setlength\dashlinedash{0.2pt}
\setlength\dashlinegap{1.5pt}
\setlength\arrayrulewidth{0.3pt}
\renewcommand{\arraystretch}{1.}
    \begin{tabular}{p{1.7cm}l|nccccccccccc}
      \toprule
$K_{\text{eff}}$ & Method & \textsc{\fontsize{11}{10}\selectfont\textcolor{black}{Average}}& \rotatebox[origin=c]{45}{ImageNet} & \rotatebox[origin=c]{45}{SUN397} & \rotatebox[origin=c]{45}{Aircraft} & \rotatebox[origin=c]{45}{EuroSAT} & \rotatebox[origin=c]{45}{StanfordCars} & \rotatebox[origin=c]{45}{Food101} & \rotatebox[origin=c]{45}{Pets} & \rotatebox[origin=c]{45}{Flower102} & \rotatebox[origin=c]{45}{Caltech101} & \rotatebox[origin=c]{45}{DTD} & \rotatebox[origin=c]{45}{UCF101}  \\
\midrule
 & CLIP & $\textcolor{black}{61.9}$ & $\text{62.0}$ & $\text{62.1}$ & $\text{19.1}$ & $\text{45.4}$ & $\text{60.2}$ & $\text{80.4}$ & $\text{87.3}$ & $\text{66.6}$ & $\text{91.4}$ & $\text{42.7}$ & $\text{63.5}$  \\
 
\midrule

  \texttt{All} & \cellcolor{LightGray} Stat${\cal A}$ & \cellcolor{AverageDarkerColor} $\textcolor{black}{\textbf{65.5}}_{\textcolor{darkerGreen}{\raisebox{2pt}{\normalsize+\textbf{3.7}}}}$ & \cellcolor{LightGray} $\text{64.6}_{\textcolor{darkerGreen}{\raisebox{2pt}{\normalsize+2.6}}}$ & \cellcolor{LightGray} $\text{67.1}_{\textcolor{darkerGreen}{\raisebox{2pt}{\normalsize+5.0}}}$ & \cellcolor{LightGray} $\text{19.7}_{\textcolor{darkerGreen}{\raisebox{2pt}{\normalsize+0.6}}}$ & \cellcolor{LightGray} $\text{54.3}_{\textcolor{darkerGreen}{\raisebox{2pt}{\normalsize+8.9}}}$ & \cellcolor{LightGray} $\text{62.5}_{\textcolor{darkerGreen}{\raisebox{2pt}{\normalsize+2.3}}}$ & \cellcolor{LightGray} $\text{81.4}_{\textcolor{darkerGreen}{\raisebox{2pt}{\normalsize+1.0}}}$ & \cellcolor{LightGray} $\text{89.2}_{\textcolor{darkerGreen}{\raisebox{2pt}{\normalsize+1.9}}}$ & \cellcolor{LightGray} $\text{71.5}_{\textcolor{darkerGreen}{\raisebox{2pt}{\normalsize+4.9}}}$ & \cellcolor{LightGray} $\text{91.5}_{\textcolor{darkerGreen}{\raisebox{2pt}{\normalsize+0.1}}}$ & \cellcolor{LightGray} $\text{50.8}_{\textcolor{darkerGreen}{\raisebox{2pt}{\normalsize+8.1}}}$ & \cellcolor{LightGray} $\text{68.4}_{\textcolor{darkerGreen}{\raisebox{2pt}{\normalsize+4.9}}}$  \\
\midrule
\end{tabular}}
\end{subtable}
\\
\begin{subtable}{\linewidth}
\centering
\caption{ViT-L/14.}
\resizebox{\textwidth}{!}{%
\setlength\dashlinedash{0.2pt}
\setlength\dashlinegap{1.5pt}
\setlength\arrayrulewidth{0.3pt}
\renewcommand{\arraystretch}{1.}
    \begin{tabular}{p{1.7cm}l|nccccccccccc}
      \toprule
$K_{\text{eff}}$ & Method & \textsc{\fontsize{11}{10}\selectfont\textcolor{black}{Average}}& \rotatebox[origin=c]{45}{ImageNet} & \rotatebox[origin=c]{45}{SUN397} & \rotatebox[origin=c]{45}{Aircraft} & \rotatebox[origin=c]{45}{EuroSAT} & \rotatebox[origin=c]{45}{StanfordCars} & \rotatebox[origin=c]{45}{Food101} & \rotatebox[origin=c]{45}{Pets} & \rotatebox[origin=c]{45}{Flower102} & \rotatebox[origin=c]{45}{Caltech101} & \rotatebox[origin=c]{45}{DTD} & \rotatebox[origin=c]{45}{UCF101}  \\
\midrule
 & CLIP & $\textcolor{black}{72.6}$ & $\text{73.5}$ & $\text{67.7}$ & $\text{32.5}$ & $\text{60.3}$ & $\text{76.9}$ & $\text{90.9}$ & $\text{93.5}$ & $\text{79.5}$ & $\text{95.2}$ & $\text{53.5}$ & $\text{74.9}$  \\

\midrule

  \texttt{All} & \cellcolor{LightGray} Stat${\cal A}$ & \cellcolor{AverageDarkerColor} $\textcolor{black}{\textbf{76.7}}_{\textcolor{darkerGreen}{\raisebox{2pt}{\normalsize+\textbf{4.1}}}}$ & \cellcolor{LightGray} $\text{76.9}_{\textcolor{darkerGreen}{\raisebox{2pt}{\normalsize+3.4}}}$ & \cellcolor{LightGray} $\text{72.8}_{\textcolor{darkerGreen}{\raisebox{2pt}{\normalsize+5.1}}}$ & \cellcolor{LightGray} $\text{35.4}_{\textcolor{darkerGreen}{\raisebox{2pt}{\normalsize+2.9}}}$ & \cellcolor{LightGray} $\text{76.7}_{\textcolor{darkerGreen}{\raisebox{2pt}{\normalsize+16.4}}}$ & \cellcolor{LightGray} $\text{78.0}_{\textcolor{darkerGreen}{\raisebox{2pt}{\normalsize+1.1}}}$ & \cellcolor{LightGray} $\text{91.8}_{\textcolor{darkerGreen}{\raisebox{2pt}{\normalsize+0.9}}}$ & \cellcolor{LightGray} $\text{94.6}_{\textcolor{darkerGreen}{\raisebox{2pt}{\normalsize+1.1}}}$ & \cellcolor{LightGray} $\text{81.8}_{\textcolor{darkerGreen}{\raisebox{2pt}{\normalsize+2.3}}}$ & \cellcolor{LightGray} $\text{95.0}_{\textcolor{red}{\raisebox{2pt}{\normalsize-0.2}}}$ & \cellcolor{LightGray} $\text{59.7}_{\textcolor{darkerGreen}{\raisebox{2pt}{\normalsize+6.2}}}$ & \cellcolor{LightGray} $\text{81.3}_{\textcolor{darkerGreen}{\raisebox{2pt}{\normalsize+6.4}}}$  \\
\midrule
\end{tabular}}
\end{subtable}
\end{table*}



\definecolor{AverageColor}{RGB}{255, 237, 206}
\definecolor{AverageDarkerColor}{RGB}{236, 227, 211}

\newcolumntype{n}{>{\columncolor{AverageColor}} p{1.7cm}}

\aboverulesep = 0.2mm 
\belowrulesep = 0.2mm 

\begin{table*}
\centering
\caption{Comparison of various CLIP encoders for the online test-time adaptation setting with a batch size of 128. Each reported performance is averaged over 100 tasks.}
\label{tab:appendix_online}
\begin{subtable}{\linewidth}
\centering
\caption{ResNet-50.}
\resizebox{\textwidth}{!}{%
\setlength\dashlinedash{0.2pt}
\setlength\dashlinegap{1.5pt}
\setlength\arrayrulewidth{0.3pt}
\renewcommand{\arraystretch}{1.}
    \begin{tabular}{ll|ncccccccccccc}
      \toprule
Scenario & Method & \textsc{\fontsize{11}{10}\selectfont\textcolor{black}{Average}} & \rotatebox[origin=c]{45}{ImageNet} & \rotatebox[origin=c]{45}{SUN397} & \rotatebox[origin=c]{45}{Aircraft} & \rotatebox[origin=c]{45}{EuroSAT} & \rotatebox[origin=c]{45}{StanfordCars} & \rotatebox[origin=c]{45}{Food101} & \rotatebox[origin=c]{45}{Pets} & \rotatebox[origin=c]{45}{Flower102} & \rotatebox[origin=c]{45}{Caltech101} & \rotatebox[origin=c]{45}{DTD} & \rotatebox[origin=c]{45}{UCF101} \\
\midrule
  & CLIP& $\text{58.7}$ & $\text{58.2}$ & $\text{58.9}$ & $\text{17.0}$ & $\text{36.2}$ & $\text{55.8}$ & $\text{77.4}$ & $\text{85.7}$ & $\text{66.1}$ & $\text{85.7}$ & $\text{42.8}$ & $\text{61.8}$ \\
\midrule
\multirow{1}{*}{\makecell{\texttt{Low} }} & \cellcolor{LightGray} Stat${\cal A}$ & \cellcolor{AverageDarkerColor} $\text{58.4}_{\textcolor{red}{\raisebox{2pt}{\normalsize-\textbf{0.3}}}}$ & \cellcolor{LightGray} $\text{54.6}_{\textcolor{red}{\raisebox{2pt}{\normalsize-\textbf{3.6}}}}$ & \cellcolor{LightGray} $\text{56.6}_{\textcolor{red}{\raisebox{2pt}{\normalsize-\textbf{2.3}}}}$ & \cellcolor{LightGray} $\text{15.1}_{\textcolor{red}{\raisebox{2pt}{\normalsize-\textbf{1.9}}}}$ & \cellcolor{LightGray} $\text{39.7}_{\textcolor{darkerGreen}{\raisebox{2pt}{\normalsize+\textbf{3.5}}}}$ & \cellcolor{LightGray} $\text{57.6}_{\textcolor{darkerGreen}{\raisebox{2pt}{\normalsize+\textbf{1.8}}}}$ & \cellcolor{LightGray} $\text{79.4}_{\textcolor{darkerGreen}{\raisebox{2pt}{\normalsize+\textbf{2.0}}}}$ & \cellcolor{LightGray} $\text{85.1}_{\textcolor{red}{\raisebox{2pt}{\normalsize-\textbf{0.6}}}}$ & \cellcolor{LightGray} $\text{60.7}_{\textcolor{red}{\raisebox{2pt}{\normalsize-\textbf{5.4}}}}$ & \cellcolor{LightGray} $\text{87.8}_{\textcolor{darkerGreen}{\raisebox{2pt}{\normalsize+\textbf{2.1}}}}$ & \cellcolor{LightGray} $\text{44.4}_{\textcolor{darkerGreen}{\raisebox{2pt}{\normalsize+\textbf{1.6}}}}$ & \cellcolor{LightGray} $\text{61.7}_{\textcolor{red}{\raisebox{2pt}{\normalsize-\textbf{0.1}}}}$ \\
\midrule
\multirow{1}{*}{\makecell{\texttt{Medium} }} & \cellcolor{LightGray} Stat${\cal A}$ & \cellcolor{AverageDarkerColor} $\textbf{62.8}_{\textcolor{darkerGreen}{\raisebox{2pt}{\normalsize+\textbf{4.1}}}}$ & \cellcolor{LightGray} $\text{59.6}_{\textcolor{darkerGreen}{\raisebox{2pt}{\normalsize+\textbf{1.4}}}}$ & \cellcolor{LightGray} $\text{60.8}_{\textcolor{darkerGreen}{\raisebox{2pt}{\normalsize+\textbf{1.9}}}}$ & \cellcolor{LightGray} $\text{17.7}_{\textcolor{darkerGreen}{\raisebox{2pt}{\normalsize+\textbf{0.7}}}}$ & \cellcolor{LightGray} $\text{43.5}_{\textcolor{darkerGreen}{\raisebox{2pt}{\normalsize+\textbf{7.3}}}}$ & \cellcolor{LightGray} $\text{65.9}_{\textcolor{darkerGreen}{\raisebox{2pt}{\normalsize+\textbf{10.1}}}}$ & \cellcolor{LightGray} $\text{84.5}_{\textcolor{darkerGreen}{\raisebox{2pt}{\normalsize+\textbf{7.1}}}}$ & \cellcolor{LightGray} $\text{90.6}_{\textcolor{darkerGreen}{\raisebox{2pt}{\normalsize+\textbf{4.9}}}}$ & \cellcolor{LightGray} $\text{68.1}_{\textcolor{darkerGreen}{\raisebox{2pt}{\normalsize+\textbf{2.0}}}}$ & \cellcolor{LightGray} $\text{89.3}_{\textcolor{darkerGreen}{\raisebox{2pt}{\normalsize+\textbf{3.6}}}}$ & \cellcolor{LightGray} $\text{45.5}_{\textcolor{darkerGreen}{\raisebox{2pt}{\normalsize+\textbf{2.7}}}}$ & \cellcolor{LightGray} $\text{64.5}_{\textcolor{darkerGreen}{\raisebox{2pt}{\normalsize+\textbf{2.7}}}}$ \\
\midrule
\multirow{1}{*}{\makecell{\texttt{High}}} & \cellcolor{LightGray} Stat${\cal A}$ & \cellcolor{AverageDarkerColor} $\textbf{64.3}_{\textcolor{darkerGreen}{\raisebox{2pt}{\normalsize+\textbf{5.6}}}}$ & \cellcolor{LightGray} $\text{64.7}_{\textcolor{darkerGreen}{\raisebox{2pt}{\normalsize+\textbf{6.5}}}}$ & \cellcolor{LightGray} $\text{62.6}_{\textcolor{darkerGreen}{\raisebox{2pt}{\normalsize+\textbf{3.7}}}}$ & \cellcolor{LightGray} $\text{18.5}_{\textcolor{darkerGreen}{\raisebox{2pt}{\normalsize+\textbf{1.5}}}}$ & \cellcolor{LightGray} $\text{43.6}_{\textcolor{darkerGreen}{\raisebox{2pt}{\normalsize+\textbf{7.4}}}}$ & \cellcolor{LightGray} $\text{68.5}_{\textcolor{darkerGreen}{\raisebox{2pt}{\normalsize+\textbf{12.7}}}}$ & \cellcolor{LightGray} $\text{85.8}_{\textcolor{darkerGreen}{\raisebox{2pt}{\normalsize+\textbf{8.4}}}}$ & \cellcolor{LightGray} $\text{92.2}_{\textcolor{darkerGreen}{\raisebox{2pt}{\normalsize+\textbf{6.5}}}}$ & \cellcolor{LightGray} $\text{70.1}_{\textcolor{darkerGreen}{\raisebox{2pt}{\normalsize+\textbf{4.0}}}}$ & \cellcolor{LightGray} $\text{89.8}_{\textcolor{darkerGreen}{\raisebox{2pt}{\normalsize+\textbf{4.1}}}}$ & \cellcolor{LightGray} $\text{45.9}_{\textcolor{darkerGreen}{\raisebox{2pt}{\normalsize+\textbf{3.1}}}}$ & \cellcolor{LightGray} $\text{65.2}_{\textcolor{darkerGreen}{\raisebox{2pt}{\normalsize+\textbf{3.4}}}}$ \\
\midrule
\multirow{1}{*}{\makecell{\texttt{Separate}}} & \cellcolor{LightGray} Stat${\cal A}$ & \cellcolor{AverageDarkerColor} $\textbf{65.1}_{\textcolor{darkerGreen}{\raisebox{2pt}{\normalsize+\textbf{6.4}}}}$ & \cellcolor{LightGray} $\text{66.6}_{\textcolor{darkerGreen}{\raisebox{2pt}{\normalsize+\textbf{8.4}}}}$ & \cellcolor{LightGray} $\text{62.6}_{\textcolor{darkerGreen}{\raisebox{2pt}{\normalsize+\textbf{3.7}}}}$ & \cellcolor{LightGray} $\text{19.8}_{\textcolor{darkerGreen}{\raisebox{2pt}{\normalsize+\textbf{2.8}}}}$ & \cellcolor{LightGray} $\text{44.3}_{\textcolor{darkerGreen}{\raisebox{2pt}{\normalsize+\textbf{8.1}}}}$ & \cellcolor{LightGray} $\text{69.5}_{\textcolor{darkerGreen}{\raisebox{2pt}{\normalsize+\textbf{13.7}}}}$ & \cellcolor{LightGray} $\text{85.6}_{\textcolor{darkerGreen}{\raisebox{2pt}{\normalsize+\textbf{8.2}}}}$ & \cellcolor{LightGray} $\text{93.8}_{\textcolor{darkerGreen}{\raisebox{2pt}{\normalsize+\textbf{8.1}}}}$ & \cellcolor{LightGray} $\text{71.9}_{\textcolor{darkerGreen}{\raisebox{2pt}{\normalsize+\textbf{5.8}}}}$ & \cellcolor{LightGray} $\text{90.2}_{\textcolor{darkerGreen}{\raisebox{2pt}{\normalsize+\textbf{4.5}}}}$ & \cellcolor{LightGray} $\text{46.0}_{\textcolor{darkerGreen}{\raisebox{2pt}{\normalsize+\textbf{3.2}}}}$ & \cellcolor{LightGray} $\text{65.3}_{\textcolor{darkerGreen}{\raisebox{2pt}{\normalsize+\textbf{3.5}}}}$ \\
\midrule
\end{tabular}}
\end{subtable}
\\
\begin{subtable}{\linewidth}
\centering
\caption{ResNet-101.}
\resizebox{\textwidth}{!}{%
\setlength\dashlinedash{0.2pt}
\setlength\dashlinegap{1.5pt}
\setlength\arrayrulewidth{0.3pt}
\renewcommand{\arraystretch}{1.}
    \begin{tabular}{ll|ncccccccccccc}
      \toprule
Scenario & Method & \textsc{\fontsize{11}{10}\selectfont\textcolor{black}{Average}} & \rotatebox[origin=c]{45}{ImageNet} & \rotatebox[origin=c]{45}{SUN397} & \rotatebox[origin=c]{45}{Aircraft} & \rotatebox[origin=c]{45}{EuroSAT} & \rotatebox[origin=c]{45}{StanfordCars} & \rotatebox[origin=c]{45}{Food101} & \rotatebox[origin=c]{45}{Pets} & \rotatebox[origin=c]{45}{Flower102} & \rotatebox[origin=c]{45}{Caltech101} & \rotatebox[origin=c]{45}{DTD} & \rotatebox[origin=c]{45}{UCF101} \\
\midrule
  & CLIP& $\text{59.5}$ & $\text{61.3}$ & $\text{59.0}$ & $\text{17.9}$ & $\text{32.9}$ & $\text{63.2}$ & $\text{80.7}$ & $\text{86.9}$ & $\text{64.3}$ & $\text{89.9}$ & $\text{37.3}$ & $\text{61.1}$ \\
\midrule
\multirow{1}{*}{\makecell{\texttt{Low}}} & \cellcolor{LightGray} Stat${\cal A}$ & \cellcolor{AverageDarkerColor} $\textbf{61.3}_{\textcolor{darkerGreen}{\raisebox{2pt}{\normalsize+\textbf{1.8}}}}$ & \cellcolor{LightGray} $\text{60.5}_{\textcolor{red}{\raisebox{2pt}{\normalsize-\textbf{0.8}}}}$ & \cellcolor{LightGray} $\text{59.3}_{\textcolor{darkerGreen}{\raisebox{2pt}{\normalsize+\textbf{0.3}}}}$ & \cellcolor{LightGray} $\text{16.9}_{\textcolor{red}{\raisebox{2pt}{\normalsize-\textbf{1.0}}}}$ & \cellcolor{LightGray} $\text{32.7}_{\textcolor{red}{\raisebox{2pt}{\normalsize-\textbf{0.2}}}}$ & \cellcolor{LightGray} $\text{65.5}_{\textcolor{darkerGreen}{\raisebox{2pt}{\normalsize+\textbf{2.3}}}}$ & \cellcolor{LightGray} $\text{84.9}_{\textcolor{darkerGreen}{\raisebox{2pt}{\normalsize+\textbf{4.2}}}}$ & \cellcolor{LightGray} $\text{91.0}_{\textcolor{darkerGreen}{\raisebox{2pt}{\normalsize+\textbf{4.1}}}}$ & \cellcolor{LightGray} $\text{67.8}_{\textcolor{darkerGreen}{\raisebox{2pt}{\normalsize+\textbf{3.5}}}}$ & \cellcolor{LightGray} $\text{92.2}_{\textcolor{darkerGreen}{\raisebox{2pt}{\normalsize+\textbf{2.3}}}}$ & \cellcolor{LightGray} $\text{41.1}_{\textcolor{darkerGreen}{\raisebox{2pt}{\normalsize+\textbf{3.8}}}}$ & \cellcolor{LightGray} $\text{62.8}_{\textcolor{darkerGreen}{\raisebox{2pt}{\normalsize+\textbf{1.7}}}}$ \\
\midrule
\multirow{1}{*}{\makecell{\texttt{Medium} }} & \cellcolor{LightGray} Stat${\cal A}$ & \cellcolor{AverageDarkerColor} $\textbf{64.6}_{\textcolor{darkerGreen}{\raisebox{2pt}{\normalsize+\textbf{5.1}}}}$ & \cellcolor{LightGray} $\text{66.1}_{\textcolor{darkerGreen}{\raisebox{2pt}{\normalsize+\textbf{4.8}}}}$ & \cellcolor{LightGray} $\text{64.2}_{\textcolor{darkerGreen}{\raisebox{2pt}{\normalsize+\textbf{5.2}}}}$ & \cellcolor{LightGray} $\text{19.7}_{\textcolor{darkerGreen}{\raisebox{2pt}{\normalsize+\textbf{1.8}}}}$ & \cellcolor{LightGray} $\text{33.3}_{\textcolor{darkerGreen}{\raisebox{2pt}{\normalsize+\textbf{0.4}}}}$ & \cellcolor{LightGray} $\text{72.2}_{\textcolor{darkerGreen}{\raisebox{2pt}{\normalsize+\textbf{9.0}}}}$ & \cellcolor{LightGray} $\text{88.1}_{\textcolor{darkerGreen}{\raisebox{2pt}{\normalsize+\textbf{7.4}}}}$ & \cellcolor{LightGray} $\text{94.1}_{\textcolor{darkerGreen}{\raisebox{2pt}{\normalsize+\textbf{7.2}}}}$ & \cellcolor{LightGray} $\text{72.1}_{\textcolor{darkerGreen}{\raisebox{2pt}{\normalsize+\textbf{7.8}}}}$ & \cellcolor{LightGray} $\text{93.2}_{\textcolor{darkerGreen}{\raisebox{2pt}{\normalsize+\textbf{3.3}}}}$ & \cellcolor{LightGray} $\text{42.9}_{\textcolor{darkerGreen}{\raisebox{2pt}{\normalsize+\textbf{5.6}}}}$ & \cellcolor{LightGray} $\text{65.2}_{\textcolor{darkerGreen}{\raisebox{2pt}{\normalsize+\textbf{4.1}}}}$ \\
\midrule
\multirow{1}{*}{\makecell{\texttt{High} }} & \cellcolor{LightGray} Stat${\cal A}$ & \cellcolor{AverageDarkerColor} $\textbf{65.7}_{\textcolor{darkerGreen}{\raisebox{2pt}{\normalsize+\textbf{6.2}}}}$ & \cellcolor{LightGray} $\text{70.5}_{\textcolor{darkerGreen}{\raisebox{2pt}{\normalsize+\textbf{9.2}}}}$ & \cellcolor{LightGray} $\text{65.9}_{\textcolor{darkerGreen}{\raisebox{2pt}{\normalsize+\textbf{6.9}}}}$ & \cellcolor{LightGray} $\text{20.6}_{\textcolor{darkerGreen}{\raisebox{2pt}{\normalsize+\textbf{2.7}}}}$ & \cellcolor{LightGray} $\text{33.5}_{\textcolor{darkerGreen}{\raisebox{2pt}{\normalsize+\textbf{0.6}}}}$ & \cellcolor{LightGray} $\text{74.1}_{\textcolor{darkerGreen}{\raisebox{2pt}{\normalsize+\textbf{10.9}}}}$ & \cellcolor{LightGray} $\text{88.7}_{\textcolor{darkerGreen}{\raisebox{2pt}{\normalsize+\textbf{8.0}}}}$ & \cellcolor{LightGray} $\text{94.4}_{\textcolor{darkerGreen}{\raisebox{2pt}{\normalsize+\textbf{7.5}}}}$ & \cellcolor{LightGray} $\text{73.1}_{\textcolor{darkerGreen}{\raisebox{2pt}{\normalsize+\textbf{8.8}}}}$ & \cellcolor{LightGray} $\text{93.4}_{\textcolor{darkerGreen}{\raisebox{2pt}{\normalsize+\textbf{3.5}}}}$ & \cellcolor{LightGray} $\text{43.0}_{\textcolor{darkerGreen}{\raisebox{2pt}{\normalsize+\textbf{5.7}}}}$ & \cellcolor{LightGray} $\text{65.7}_{\textcolor{darkerGreen}{\raisebox{2pt}{\normalsize+\textbf{4.6}}}}$ \\
\midrule
\multirow{1}{*}{\makecell{\texttt{Separate}}} & \cellcolor{LightGray} Stat${\cal A}$ & \cellcolor{AverageDarkerColor} $\textbf{65.8}_{\textcolor{darkerGreen}{\raisebox{2pt}{\normalsize+\textbf{6.3}}}}$ & \cellcolor{LightGray} $\text{71.4}_{\textcolor{darkerGreen}{\raisebox{2pt}{\normalsize+\textbf{10.1}}}}$ & \cellcolor{LightGray} $\text{65.7}_{\textcolor{darkerGreen}{\raisebox{2pt}{\normalsize+\textbf{6.7}}}}$ & \cellcolor{LightGray} $\text{22.1}_{\textcolor{darkerGreen}{\raisebox{2pt}{\normalsize+\textbf{4.2}}}}$ & \cellcolor{LightGray} $\text{32.2}_{\textcolor{red}{\raisebox{2pt}{\normalsize-\textbf{0.7}}}}$ & \cellcolor{LightGray} $\text{74.9}_{\textcolor{darkerGreen}{\raisebox{2pt}{\normalsize+\textbf{11.7}}}}$ & \cellcolor{LightGray} $\text{88.5}_{\textcolor{darkerGreen}{\raisebox{2pt}{\normalsize+\textbf{7.8}}}}$ & \cellcolor{LightGray} $\text{94.2}_{\textcolor{darkerGreen}{\raisebox{2pt}{\normalsize+\textbf{7.3}}}}$ & \cellcolor{LightGray} $\text{73.9}_{\textcolor{darkerGreen}{\raisebox{2pt}{\normalsize+\textbf{9.6}}}}$ & \cellcolor{LightGray} $\text{93.4}_{\textcolor{darkerGreen}{\raisebox{2pt}{\normalsize+\textbf{3.5}}}}$ & \cellcolor{LightGray} $\text{41.9}_{\textcolor{darkerGreen}{\raisebox{2pt}{\normalsize+\textbf{4.6}}}}$ & \cellcolor{LightGray} $\text{65.7}_{\textcolor{darkerGreen}{\raisebox{2pt}{\normalsize+\textbf{4.6}}}}$ \\
\midrule
\end{tabular}}
\end{subtable}
\\
\begin{subtable}{\linewidth}
\centering
\caption{ViT-B/32.}
\resizebox{\textwidth}{!}{%
\setlength\dashlinedash{0.2pt}
\setlength\dashlinegap{1.5pt}
\setlength\arrayrulewidth{0.3pt}
\renewcommand{\arraystretch}{1.}
    \begin{tabular}{ll|ncccccccccccc}
      \toprule
Scenario & Method & \textsc{\fontsize{11}{10}\selectfont\textcolor{black}{Average}} & \rotatebox[origin=c]{45}{ImageNet} & \rotatebox[origin=c]{45}{SUN397} & \rotatebox[origin=c]{45}{Aircraft} & \rotatebox[origin=c]{45}{EuroSAT} & \rotatebox[origin=c]{45}{StanfordCars} & \rotatebox[origin=c]{45}{Food101} & \rotatebox[origin=c]{45}{Pets} & \rotatebox[origin=c]{45}{Flower102} & \rotatebox[origin=c]{45}{Caltech101} & \rotatebox[origin=c]{45}{DTD} & \rotatebox[origin=c]{45}{UCF101} \\
\midrule
  & CLIP& $\text{61.9}$ & $\text{62.0}$ & $\text{62.1}$ & $\text{19.1}$ & $\text{45.4}$ & $\text{60.2}$ & $\text{80.4}$ & $\text{87.3}$ & $\text{66.6}$ & $\text{91.4}$ & $\text{42.7}$ & $\text{63.5}$ \\
 
\midrule
\multirow{1}{*}{\makecell{\texttt{Low} }} & \cellcolor{LightGray} Stat${\cal A}$ & \cellcolor{AverageDarkerColor} $\textbf{63.9}_{\textcolor{darkerGreen}{\raisebox{2pt}{\normalsize+\textbf{2.0}}}}$ & \cellcolor{LightGray} $\text{61.4}_{\textcolor{red}{\raisebox{2pt}{\normalsize-\textbf{0.6}}}}$ & \cellcolor{LightGray} $\text{62.7}_{\textcolor{darkerGreen}{\raisebox{2pt}{\normalsize+\textbf{0.6}}}}$ & \cellcolor{LightGray} $\text{19.2}_{\textcolor{darkerGreen}{\raisebox{2pt}{\normalsize+\textbf{0.1}}}}$ & \cellcolor{LightGray} $\text{51.0}_{\textcolor{darkerGreen}{\raisebox{2pt}{\normalsize+\textbf{5.6}}}}$ & \cellcolor{LightGray} $\text{61.8}_{\textcolor{darkerGreen}{\raisebox{2pt}{\normalsize+\textbf{1.6}}}}$ & \cellcolor{LightGray} $\text{82.6}_{\textcolor{darkerGreen}{\raisebox{2pt}{\normalsize+\textbf{2.2}}}}$ & \cellcolor{LightGray} $\text{91.0}_{\textcolor{darkerGreen}{\raisebox{2pt}{\normalsize+\textbf{3.7}}}}$ & \cellcolor{LightGray} $\text{69.0}_{\textcolor{darkerGreen}{\raisebox{2pt}{\normalsize+\textbf{2.4}}}}$ & \cellcolor{LightGray} $\text{92.9}_{\textcolor{darkerGreen}{\raisebox{2pt}{\normalsize+\textbf{1.5}}}}$ & \cellcolor{LightGray} $\text{46.4}_{\textcolor{darkerGreen}{\raisebox{2pt}{\normalsize+\textbf{3.7}}}}$ & \cellcolor{LightGray} $\text{64.4}_{\textcolor{darkerGreen}{\raisebox{2pt}{\normalsize+\textbf{0.9}}}}$ \\
\midrule
\multirow{1}{*}{\makecell{\texttt{Medium} }} & \cellcolor{LightGray} Stat${\cal A}$ & \cellcolor{AverageDarkerColor} $\textbf{65.8}_{\textcolor{darkerGreen}{\raisebox{2pt}{\normalsize+\textbf{3.9}}}}$ & \cellcolor{LightGray} $\text{64.6}_{\textcolor{darkerGreen}{\raisebox{2pt}{\normalsize+\textbf{2.6}}}}$ & \cellcolor{LightGray} $\text{64.8}_{\textcolor{darkerGreen}{\raisebox{2pt}{\normalsize+\textbf{2.7}}}}$ & \cellcolor{LightGray} $\text{21.4}_{\textcolor{darkerGreen}{\raisebox{2pt}{\normalsize+\textbf{2.3}}}}$ & \cellcolor{LightGray} $\text{49.9}_{\textcolor{darkerGreen}{\raisebox{2pt}{\normalsize+\textbf{4.5}}}}$ & \cellcolor{LightGray} $\text{68.1}_{\textcolor{darkerGreen}{\raisebox{2pt}{\normalsize+\textbf{7.9}}}}$ & \cellcolor{LightGray} $\text{84.4}_{\textcolor{darkerGreen}{\raisebox{2pt}{\normalsize+\textbf{4.0}}}}$ & \cellcolor{LightGray} $\text{92.8}_{\textcolor{darkerGreen}{\raisebox{2pt}{\normalsize+\textbf{5.5}}}}$ & \cellcolor{LightGray} $\text{72.5}_{\textcolor{darkerGreen}{\raisebox{2pt}{\normalsize+\textbf{5.9}}}}$ & \cellcolor{LightGray} $\text{93.5}_{\textcolor{darkerGreen}{\raisebox{2pt}{\normalsize+\textbf{2.1}}}}$ & \cellcolor{LightGray} $\text{46.4}_{\textcolor{darkerGreen}{\raisebox{2pt}{\normalsize+\textbf{3.7}}}}$ & \cellcolor{LightGray} $\text{65.5}_{\textcolor{darkerGreen}{\raisebox{2pt}{\normalsize+\textbf{2.0}}}}$ \\
\midrule
\multirow{1}{*}{\makecell{\texttt{High} }} & \cellcolor{LightGray} Stat${\cal A}$ & \cellcolor{AverageDarkerColor} $\textbf{66.4}_{\textcolor{darkerGreen}{\raisebox{2pt}{\normalsize+\textbf{4.6}}}}$ & \cellcolor{LightGray} $\text{66.9}_{\textcolor{darkerGreen}{\raisebox{2pt}{\normalsize+\textbf{4.9}}}}$ & \cellcolor{LightGray} $\text{64.9}_{\textcolor{darkerGreen}{\raisebox{2pt}{\normalsize+\textbf{2.8}}}}$ & \cellcolor{LightGray} $\text{22.0}_{\textcolor{darkerGreen}{\raisebox{2pt}{\normalsize+\textbf{2.9}}}}$ & \cellcolor{LightGray} $\text{50.1}_{\textcolor{darkerGreen}{\raisebox{2pt}{\normalsize+\textbf{4.7}}}}$ & \cellcolor{LightGray} $\text{69.9}_{\textcolor{darkerGreen}{\raisebox{2pt}{\normalsize+\textbf{9.7}}}}$ & \cellcolor{LightGray} $\text{84.6}_{\textcolor{darkerGreen}{\raisebox{2pt}{\normalsize+\textbf{4.2}}}}$ & \cellcolor{LightGray} $\text{93.2}_{\textcolor{darkerGreen}{\raisebox{2pt}{\normalsize+\textbf{5.9}}}}$ & \cellcolor{LightGray} $\text{73.5}_{\textcolor{darkerGreen}{\raisebox{2pt}{\normalsize+\textbf{6.9}}}}$ & \cellcolor{LightGray} $\text{93.7}_{\textcolor{darkerGreen}{\raisebox{2pt}{\normalsize+\textbf{2.3}}}}$ & \cellcolor{LightGray} $\text{46.3}_{\textcolor{darkerGreen}{\raisebox{2pt}{\normalsize+\textbf{3.6}}}}$ & \cellcolor{LightGray} $\text{65.6}_{\textcolor{darkerGreen}{\raisebox{2pt}{\normalsize+\textbf{2.1}}}}$ \\
\midrule
\multirow{1}{*}{\makecell{\texttt{Separate} }} & \cellcolor{LightGray} Stat${\cal A}$ & \cellcolor{AverageDarkerColor} $\textbf{65.9}_{\textcolor{darkerGreen}{\raisebox{2pt}{\normalsize+\textbf{4.0}}}}$ & \cellcolor{LightGray} $\text{67.0}_{\textcolor{darkerGreen}{\raisebox{2pt}{\normalsize+\textbf{5.0}}}}$ & \cellcolor{LightGray} $\text{63.8}_{\textcolor{darkerGreen}{\raisebox{2pt}{\normalsize+\textbf{1.7}}}}$ & \cellcolor{LightGray} $\text{22.9}_{\textcolor{darkerGreen}{\raisebox{2pt}{\normalsize+\textbf{3.8}}}}$ & \cellcolor{LightGray} $\text{44.9}_{\textcolor{red}{\raisebox{2pt}{\normalsize-\textbf{0.5}}}}$ & \cellcolor{LightGray} $\text{70.4}_{\textcolor{darkerGreen}{\raisebox{2pt}{\normalsize+\textbf{10.2}}}}$ & \cellcolor{LightGray} $\text{84.1}_{\textcolor{darkerGreen}{\raisebox{2pt}{\normalsize+\textbf{3.7}}}}$ & \cellcolor{LightGray} $\text{92.8}_{\textcolor{darkerGreen}{\raisebox{2pt}{\normalsize+\textbf{5.5}}}}$ & \cellcolor{LightGray} $\text{74.6}_{\textcolor{darkerGreen}{\raisebox{2pt}{\normalsize+\textbf{8.0}}}}$ & \cellcolor{LightGray} $\text{94.0}_{\textcolor{darkerGreen}{\raisebox{2pt}{\normalsize+\textbf{2.6}}}}$ & \cellcolor{LightGray} $\text{45.1}_{\textcolor{darkerGreen}{\raisebox{2pt}{\normalsize+\textbf{2.4}}}}$ & \cellcolor{LightGray} $\text{65.0}_{\textcolor{darkerGreen}{\raisebox{2pt}{\normalsize+\textbf{1.5}}}}$ \\
\midrule
\end{tabular}}
\end{subtable}
\\
\begin{subtable}{\linewidth}
\centering
\caption{ViT-L/14.}
\resizebox{\textwidth}{!}{%
\setlength\dashlinedash{0.2pt}
\setlength\dashlinegap{1.5pt}
\setlength\arrayrulewidth{0.3pt}
\renewcommand{\arraystretch}{1.}
    \begin{tabular}{ll|ncccccccccccc}
      \toprule
Scenario & Method & \textsc{\fontsize{11}{10}\selectfont\textcolor{black}{Average}} & \rotatebox[origin=c]{45}{ImageNet} & \rotatebox[origin=c]{45}{SUN397} & \rotatebox[origin=c]{45}{Aircraft} & \rotatebox[origin=c]{45}{EuroSAT} & \rotatebox[origin=c]{45}{StanfordCars} & \rotatebox[origin=c]{45}{Food101} & \rotatebox[origin=c]{45}{Pets} & \rotatebox[origin=c]{45}{Flower102} & \rotatebox[origin=c]{45}{Caltech101} & \rotatebox[origin=c]{45}{DTD} & \rotatebox[origin=c]{45}{UCF101} \\
\midrule
  & CLIP& $\text{72.6}$ & $\text{73.5}$ & $\text{67.7}$ & $\text{32.5}$ & $\text{60.3}$ & $\text{76.9}$ & $\text{90.9}$ & $\text{93.5}$ & $\text{79.5}$ & $\text{95.2}$ & $\text{53.5}$ & $\text{74.9}$ \\
\midrule
\multirow{1}{*}{\makecell{\texttt{Low} }} & \cellcolor{LightGray} Stat${\cal A}$ & \cellcolor{AverageDarkerColor} $\textbf{74.3}_{\textcolor{darkerGreen}{\raisebox{2pt}{\normalsize+\textbf{1.7}}}}$ & \cellcolor{LightGray} $\text{73.3}_{\textcolor{red}{\raisebox{2pt}{\normalsize-\textbf{0.2}}}}$ & \cellcolor{LightGray} $\text{68.2}_{\textcolor{darkerGreen}{\raisebox{2pt}{\normalsize+\textbf{0.5}}}}$ & \cellcolor{LightGray} $\text{34.1}_{\textcolor{darkerGreen}{\raisebox{2pt}{\normalsize+\textbf{1.6}}}}$ & \cellcolor{LightGray} $\text{68.8}_{\textcolor{darkerGreen}{\raisebox{2pt}{\normalsize+\textbf{8.5}}}}$ & \cellcolor{LightGray} $\text{77.7}_{\textcolor{darkerGreen}{\raisebox{2pt}{\normalsize+\textbf{0.8}}}}$ & \cellcolor{LightGray} $\text{92.0}_{\textcolor{darkerGreen}{\raisebox{2pt}{\normalsize+\textbf{1.1}}}}$ & \cellcolor{LightGray} $\text{95.0}_{\textcolor{darkerGreen}{\raisebox{2pt}{\normalsize+\textbf{1.5}}}}$ & \cellcolor{LightGray} $\text{80.2}_{\textcolor{darkerGreen}{\raisebox{2pt}{\normalsize+\textbf{0.7}}}}$ & \cellcolor{LightGray} $\text{95.6}_{\textcolor{darkerGreen}{\raisebox{2pt}{\normalsize+\textbf{0.4}}}}$ & \cellcolor{LightGray} $\text{55.4}_{\textcolor{darkerGreen}{\raisebox{2pt}{\normalsize+\textbf{1.9}}}}$ & \cellcolor{LightGray} $\text{76.9}_{\textcolor{darkerGreen}{\raisebox{2pt}{\normalsize+\textbf{2.0}}}}$ \\
\midrule
\multirow{1}{*}{\makecell{\texttt{Medium} }} & \cellcolor{LightGray} Stat${\cal A}$ & \cellcolor{AverageDarkerColor} $\textbf{76.0}_{\textcolor{darkerGreen}{\raisebox{2pt}{\normalsize+\textbf{3.4}}}}$ & \cellcolor{LightGray} $\text{75.8}_{\textcolor{darkerGreen}{\raisebox{2pt}{\normalsize+\textbf{2.3}}}}$ & \cellcolor{LightGray} $\text{70.6}_{\textcolor{darkerGreen}{\raisebox{2pt}{\normalsize+\textbf{2.9}}}}$ & \cellcolor{LightGray} $\text{38.3}_{\textcolor{darkerGreen}{\raisebox{2pt}{\normalsize+\textbf{5.8}}}}$ & \cellcolor{LightGray} $\text{68.9}_{\textcolor{darkerGreen}{\raisebox{2pt}{\normalsize+\textbf{8.6}}}}$ & \cellcolor{LightGray} $\text{81.9}_{\textcolor{darkerGreen}{\raisebox{2pt}{\normalsize+\textbf{5.0}}}}$ & \cellcolor{LightGray} $\text{93.2}_{\textcolor{darkerGreen}{\raisebox{2pt}{\normalsize+\textbf{2.3}}}}$ & \cellcolor{LightGray} $\text{96.3}_{\textcolor{darkerGreen}{\raisebox{2pt}{\normalsize+\textbf{2.8}}}}$ & \cellcolor{LightGray} $\text{81.5}_{\textcolor{darkerGreen}{\raisebox{2pt}{\normalsize+\textbf{2.0}}}}$ & \cellcolor{LightGray} $\text{95.8}_{\textcolor{darkerGreen}{\raisebox{2pt}{\normalsize+\textbf{0.6}}}}$ & \cellcolor{LightGray} $\text{55.6}_{\textcolor{darkerGreen}{\raisebox{2pt}{\normalsize+\textbf{2.1}}}}$ & \cellcolor{LightGray} $\text{77.6}_{\textcolor{darkerGreen}{\raisebox{2pt}{\normalsize+\textbf{2.7}}}}$ \\
\midrule
\multirow{1}{*}{\makecell{\texttt{High} }} & \cellcolor{LightGray} Stat${\cal A}$ & \cellcolor{AverageDarkerColor} $\textbf{76.4}_{\textcolor{darkerGreen}{\raisebox{2pt}{\normalsize+\textbf{3.8}}}}$ & \cellcolor{LightGray} $\text{77.6}_{\textcolor{darkerGreen}{\raisebox{2pt}{\normalsize+\textbf{4.1}}}}$ & \cellcolor{LightGray} $\text{71.1}_{\textcolor{darkerGreen}{\raisebox{2pt}{\normalsize+\textbf{3.4}}}}$ & \cellcolor{LightGray} $\text{39.6}_{\textcolor{darkerGreen}{\raisebox{2pt}{\normalsize+\textbf{7.1}}}}$ & \cellcolor{LightGray} $\text{68.9}_{\textcolor{darkerGreen}{\raisebox{2pt}{\normalsize+\textbf{8.6}}}}$ & \cellcolor{LightGray} $\text{82.9}_{\textcolor{darkerGreen}{\raisebox{2pt}{\normalsize+\textbf{6.0}}}}$ & \cellcolor{LightGray} $\text{93.5}_{\textcolor{darkerGreen}{\raisebox{2pt}{\normalsize+\textbf{2.6}}}}$ & \cellcolor{LightGray} $\text{96.5}_{\textcolor{darkerGreen}{\raisebox{2pt}{\normalsize+\textbf{3.0}}}}$ & \cellcolor{LightGray} $\text{81.9}_{\textcolor{darkerGreen}{\raisebox{2pt}{\normalsize+\textbf{2.4}}}}$ & \cellcolor{LightGray} $\text{95.7}_{\textcolor{darkerGreen}{\raisebox{2pt}{\normalsize+\textbf{0.5}}}}$ & \cellcolor{LightGray} $\text{55.5}_{\textcolor{darkerGreen}{\raisebox{2pt}{\normalsize+\textbf{2.0}}}}$ & \cellcolor{LightGray} $\text{77.5}_{\textcolor{darkerGreen}{\raisebox{2pt}{\normalsize+\textbf{2.6}}}}$ \\
\midrule
\multirow{1}{*}{\makecell{\texttt{Separate} }} & \cellcolor{LightGray} Stat${\cal A}$ & \cellcolor{AverageDarkerColor} $\textbf{76.1}_{\textcolor{darkerGreen}{\raisebox{2pt}{\normalsize+\textbf{3.6}}}}$ & \cellcolor{LightGray} $\text{77.6}_{\textcolor{darkerGreen}{\raisebox{2pt}{\normalsize+\textbf{4.1}}}}$ & \cellcolor{LightGray} $\text{70.5}_{\textcolor{darkerGreen}{\raisebox{2pt}{\normalsize+\textbf{2.8}}}}$ & \cellcolor{LightGray} $\text{41.3}_{\textcolor{darkerGreen}{\raisebox{2pt}{\normalsize+\textbf{8.8}}}}$ & \cellcolor{LightGray} $\text{66.3}_{\textcolor{darkerGreen}{\raisebox{2pt}{\normalsize+\textbf{6.0}}}}$ & \cellcolor{LightGray} $\text{83.2}_{\textcolor{darkerGreen}{\raisebox{2pt}{\normalsize+\textbf{6.3}}}}$ & \cellcolor{LightGray} $\text{93.5}_{\textcolor{darkerGreen}{\raisebox{2pt}{\normalsize+\textbf{2.6}}}}$ & \cellcolor{LightGray} $\text{96.3}_{\textcolor{darkerGreen}{\raisebox{2pt}{\normalsize+\textbf{2.8}}}}$ & \cellcolor{LightGray} $\text{82.0}_{\textcolor{darkerGreen}{\raisebox{2pt}{\normalsize+\textbf{2.5}}}}$ & \cellcolor{LightGray} $\text{95.8}_{\textcolor{darkerGreen}{\raisebox{2pt}{\normalsize+\textbf{0.6}}}}$ & \cellcolor{LightGray} $\text{54.5}_{\textcolor{darkerGreen}{\raisebox{2pt}{\normalsize+\textbf{1.0}}}}$ & \cellcolor{LightGray} $\text{76.8}_{\textcolor{darkerGreen}{\raisebox{2pt}{\normalsize+\textbf{1.9}}}}$ \\
\midrule
\end{tabular}}
\end{subtable}
\end{table*}



\end{document}